\pdfoutput=1

\documentclass[sigconf,9pt]{acmart}

\usepackage{setspace}
\usepackage{enumitem}

\usepackage{colortbl}
\usepackage[ruled,linesnumbered]{algorithm2e}
\usepackage{tabularx}
\usepackage{multirow}

\usepackage{booktabs,subcaption,amsfonts,dcolumn}
\newcolumntype{d}[1]{D..{#1}}

\usepackage{mathtools}

\usepackage{bm}

\usepackage{overpic}

\begin{document}

\acmYear{2026}\copyrightyear{2026}
\setcopyright{cc}
\setcctype[4.0]{by-nc-nd}
\acmConference[MobiSys '26]{The 24th Annual International Conference on Mobile Systems, Applications and Services}{June 21--25, 2026}{Cambridge, United Kingdom}
\acmBooktitle{The 24th Annual International Conference on Mobile Systems, Applications and Services (MobiSys '26), June 21--25, 2026, Cambridge, United Kingdom}
\acmDOI{10.1145/3745756.3809252}
\acmISBN{979-8-4007-2027-7/26/06}

\title{Physical Self-Supervised Learning: IMU Sensing without Manual Labels}

\author{Yuyang Leng}
\affiliation{
  \institution{George Mason University}
  \city{Fairfax}
  \state{VA}
  \country{USA}
}
\email{yleng2@gmu.edu}

\author{Renyuan Liu}
\affiliation{
  \institution{George Mason University}
  \city{Fairfax}
  \state{VA}
  \country{USA}
}
\email{rliu23@gmu.edu}

\author{Shaohan Hu}
\affiliation{
  \institution{Global Technology Applied Research, JPMorgan Chase}
  \city{New York}
  \state{NY}
  \country{USA}
}
\email{shaohan.hu@jpmchase.com}

\author{Peijun Zhao}
\affiliation{
  \institution{Global Technology Applied Research, JPMorgan Chase}
  \city{New York}
  \state{NY}
  \country{USA}
}
\email{peijun.zhao@jpmchase.com}

\author{Chun-Fu (Richard) Chen}
\affiliation{
  \institution{Global Technology Applied Research, JPMorgan Chase}
  \city{New York}
  \state{NY}
  \country{USA}
}
\email{richard.cf.chen@jpmchase.com}

\author{Songqing Chen}
\affiliation{
  \institution{George Mason University}
  \city{Fairfax}
  \state{VA}
  \country{USA}
}
\email{sqchen@gmu.edu}

\author{Shuochao Yao}
\affiliation{
  \institution{George Mason University}
  \city{Fairfax}
  \state{VA}
  \country{USA}
}
\email{shuochao@gmu.edu}

\sloppy

\begin{abstract}
\vspace{-0.1cm}
Deep neural networks have become a promising approach for IMU-based sensing, but their scalability is fundamentally limited by costly labeled data and poor robustness to heterogeneous devices, placements, and users. Existing unsupervised and self-supervised methods reduce but do not remove this dependence, still requiring labeled data for domain adaptation and largely ignoring known physical structure.
We propose physical self-supervised learning, an autoencoder-style paradigm for label-free IMU sensing. We replace the conventional neural decoder with an auto-adaptive physics decoder—a learnable family of kinematic equations that enforces explicit physical structure while adapting across environments—and adopt a hybrid two-stage IMU encoder with reconstruction in a structured latent space to mitigate sensor noise. Our framework further introduces probabilistic frequency–spatial constraints to disentangle sensor and object motion, a multi-view kinematic tree to exploit sparse physical self-supervised signals, and an uncertainty-aware formulation to handle the inherent ambiguity of IMU inference.
Evaluated on inertial tracking and full-body motion capture over public datasets and realistic deployments, physical self-supervised learning reduces errors by up to 5× for tracking and 4× for motion capture in challenging generalization scenarios, consistently outperforming state-of-the-art supervised and self-supervised baselines without any labels. Our code is available at \href{https://github.com/YuyangLeng/physical-ssl-imu-label-free}{\textcolor{cyan}{https://github.com/YuyangLeng/physical-ssl-imu-label-free}}
\vspace{-0.4cm}
\end{abstract}

\renewcommand{\shortauthors}{Y. Leng, R. Liu, S. Hu, P. Zhao, C.-F. Chen, S. Chen, and S. Yao}

\maketitle

{
\newcommand{\modelname}{\textbf{NaLF}}

\vspace{-0.3cm}
\section{Introduction}
For over a decade, the mobile sensing community has leveraged deep neural networks (DNNs) to tackle a wide range of challenges~\cite{lane2015can,lane2015deepear,yao2017deepsense}. Rapid advances in deep learning have led to powerful data-driven paradigms capable of addressing increasingly complex mobile sensing problems efficiently on mobile devices~\cite{jiang2018towards,ouyang2021clusterfl,yao2018sensegan,liu2025device,liu2025daf,liu2024dynaspa,leng2023scaleflow,yao2017deepiot,yao2018fastdeepiot}.
Among the many sensing modalities, the Inertial Measurement Unit (IMU) is among the most ubiquitous: IMUs are embedded in virtually all modern smartphones, wearables, and IoT devices, and they underpin large-scale applications such as smart health~\cite{fan2021headfi,hao2013isleep}, autonomous vehicles~\cite{he2023vi,zhang2023robust}, drones~\cite{10.1145/3636534.3690699,gowda2016tracking}, and augmented reality~\cite{10.1145/3636534.3690676,ponton2023sparseposer}.

Despite this progress, two major challenges continue to limit data-driven (DNN-based) mobile sensing: the scarcity of labeled data and the heterogeneity of sensing environments.
Unlike well-established DNN application domains such as vision and language, where data collection and labeling can be decoupled thanks to the human-interpretable nature of the data, embedded and mobile sensing tasks often require labels to be obtained simultaneously with data collection. This typically depends on deploying additional sensor modalities or involving human annotators in controlled lab environments, making labeling extremely costly and fundamentally unscalable.
At the same time, the performance of data-driven models is highly sensitive to the sensing environment. Factors such as device manufacturing variations, user behavior, deployment locations, sensor orientations, and movement during the sensing period (e.g., a loosely worn smartwatch or a smartphone in a loose pocket or bag), along with other contextual variables, can all critically impact model performance~\cite{stisen2015smart}. These two issues reinforce one another: to combat environmental heterogeneity, we need more labeled data collected outside controlled lab settings, yet obtaining such labels requires heavy manual effort or extra instrumentation and is difficult to scale in practice.

This paper therefore asks the following question: \emph{Can we develop a learning paradigm for IMU sensing that requires no manually labeled data, yet achieves performance comparable to or even surpassing that of supervised counterparts across diverse sensing environments?}

Unsupervised and self-supervised learning paradigms are natural candidates when seeking to reduce labeling effort. Many of the most successful approaches in this space are autoencoder-style methods: from classical autoencoders~\cite{goodfellow2016deep}, to variational autoencoders (VAEs)~\cite{kingma2013auto}, to the recent masked autoencoders (MAEs)~\cite{he2022masked}. Each of these has inspired a family of models designed to reduce labeling requirements while automatically learning useful representations from raw data.
In the mobile sensing domain, there have also been significant efforts to adapt these techniques to sensing tasks~\cite{limu_loc,haresamudram2020masked,rahimi2021self,ouyang2022cosmo,kara2024freqmae}. Unfortunately, none of these efforts fully eliminates the need for labels: a non-trivial fraction of labeled data (typically on the order of $10\% \sim 20\%$) is still required for domain adaptation. As a result, the data-scaling problem in mobile sensing remains unresolved, since manual data collection and labeling are still needed whenever the sensing environment changes.

Our key observation is that existing approaches grant autoencoder-style models too much freedom. On the one hand, because both the encoder and decoder are parameterized by DNNs, these models excel at learning representations that preserve information and suppress noise, but they also tend to produce uninterpretable latent spaces. Consequently, additional labeled data are needed to map these latent representations to the physical quantities of interest and to align them across sensing environments. On the other hand, most IMU sensing tasks are governed by kinematic equations that are known a priori. In the absence of explicit guidance, current autoencoder-style methods expend much of their capacity fitting these known physical dynamics, rather than focusing on adapting to the underlying variability of real-world sensing environments. 
Our intuition is that physical laws already provide rich prior knowledge that can be embedded into a new learning paradigm—one that simultaneously guides the model to output physically meaningful IMU quantities and directs its capacity toward automatically adapting to complex, hard-to-model sensing environments.

To this end, we propose \emph{physical self-supervised  learning}, a novel autoencoder-style paradigm that enables label-free learning for IMU sensing tasks (e.g., inertial tracking and full-body motion capture). Our paradigm replaces the fully DNN-parameterized decoder with an auto-adaptive physics decoder: a parameterized set of kinematic equations that automatically adapts to data to model a family of kinematic systems. Unlike traditional physics-based modeling, which requires handcrafting precise physical models for each sensing setup—often laborious and brittle—and unlike purely data-driven decoders that yield opaque representations, our auto-adaptive physics decoder embeds explicit kinematic structure while learning to adapt to diverse sensing scenarios, including variations in sensor placement and movement. In addition, we replace a monolithic encoder with a hybrid, two-stage IMU encoder and perform reconstruction in a structured latent space, allowing the reconstruction loss to act in the latent domain and thereby mitigating the impact of sensor noise.

Beyond this core architectural design, our physical self-supervised  framework introduces three additional contributions.
First, we propose \emph{probabilistic frequency–spatial constraints} that naturally disentangle sensor motion from object motion.
Second, we extend the classical kinematic tree for articulated objects to a \emph{multi-view kinematic tree}, enabling the use of sparse physical self-supervised  signals from multiple sensor locations.
Third, we develop an uncertainty-aware formulation that explicitly models ambiguity and helps resolve the inherently ill-posed nature of IMU sensing tasks.

We evaluate our physical self-supervised  learning framework on two representative IMU sensing tasks:
(1) Inertial tracking: estimating the global translation and orientation of an object using a single IMU, and
(2) Motion capture: recovering full-body pose using multiple IMUs.
Across multiple public datasets and realistic deployment scenarios, our framework consistently outperforms state-of-the-art supervised and self-supervised fine-tuning baselines. For inertial tracking, it reduces error by 1.5×–2× compared to prior deep learning approaches, and under challenging generalization settings the improvement reaches up to 5×. For motion capture, our method achieves 1.3×–1.8× lower error, with 3×–4× gains in generalization scenarios.
Taken together, these results demonstrate that our label-free physical self-supervised  learning framework not only outperforms all supervised and self-supervised baselines, but also generalizes robustly across sensor deployments, users, motion patterns, and datasets.

The rest of this paper is organized as follows. Section~\ref{sec:related} reviews related work. Section~\ref{sec:technical} presents the technical details of our physical self-supervised  learning framework. The experimental setup and evaluation results are given in Sections~\ref{sec:exp_setup} and~\ref{sec:evaluation}. Finally, Section~\ref{sec:conclusion} concludes the paper.
\section{Related Work}~\label{sec:related}
% \textbf{Sensing Data Heterogeneity.}
\noindent
\textbf{Self-Supervised Learning.}
Due to the scarcity of labeled data and the diverse nature of sensing data, self-supervised learning has become a key area of research in the mobile sensing community. Techniques such as contrastive learning~\cite{ouyang2022cosmo,haresamudram2021contrastive,qian2022makes} and masked reconstruction~\cite{limu_loc,haresamudram2020masked,rahimi2021self} have been adapted and tailored specifically for sensing data. Additionally, multi-modality sensing has been explored to facilitate cross-domain information fusion~\cite{ouyang2022cosmo,kara2024freqmae}. While these approaches have made progress in mitigating the challenges posed by limited labeled data and sensing heterogeneity, they still rely on a moderate fraction of labeled data (approximately $10\%\sim20\%$) to achieve performance on par with supervised methods. Additionally, incorporating labeled data from diverse sensing domains remains essential for effectively addressing the issue of sensing heterogeneity.

\noindent
\textbf{Physics Augmented Learning.}
The integration of physical information to enhance the learning process has been widely explored across various communities. In the machine learning field, physics-informed neural networks have been developed to embed physical knowledge, such as differential equation~\cite{cai2021physics,krishnapriyan2021characterizing,cai2021physicsB} and physics symmetry~\cite{cohen2016group,cohen2018spherical,satorras2021n}, directly into the learning process. However, most existing approaches aim to augment and regularize the learning process using predefined physical rules, rather than achieving fully label-free training. Furthermore, these methods are often evaluated on simplified synthetic datasets, limiting their applicability to more complex, real-world sensing scenarios.

In mobile sensing research, recent efforts have also been made on utilizing physical information to enhance learning-based sensing tasks. Researchers have explored augmenting IMU data by synthesizing complementary data from various sources, such as physical knowledge~\cite{xu2023practically,luo2021phyaug}, video clips~\cite{kwon2020imutube}, images~\cite{yoon2022img2imu}, multi-sensor correlations~\cite{zhang2023physics}, and text-to-motion synthesis models~\cite{leng2023generating}. These data augmentation techniques, when combined with existing labeled datasets, have demonstrated improved performance in sensing tasks and enhanced cross-dataset generalization. However, most of the existing work is limited to human activity recognition classification tasks and still heavily depends on labeled datasets. 

\noindent
\textbf{IMU Sensing.} 
IMU sensing tasks, such as inertial tracking and motion capture, have been a focus of study in the  community for decades~\cite{barshan1995inertial,marins2001extended,sabatini2006quaternion,lee2009gyro,bachmann2001inertial,shen2018closing,zhou2014use,von2017sparse,shen2016smartwatch,tautges2011motion,liu2019real}. These techniques have been widely applied across diverse domains, including indoor localization~\cite{zheng2014travi}, gesture recognition~\cite{khanna2024hand}, autonomous systems~\cite{ahmed2016accurate}, fitness~\cite{khurana2018gymcam}, context-aware interfaces~\cite{ahuja2020direction}, and rehabilitation~\cite{mousavi2014review}. For these impactful applications, our goal is to enable label-free training and facilitate large-scale deployment on IoT devices.

\begin{figure*}[!h]
    \includegraphics[scale=0.5]{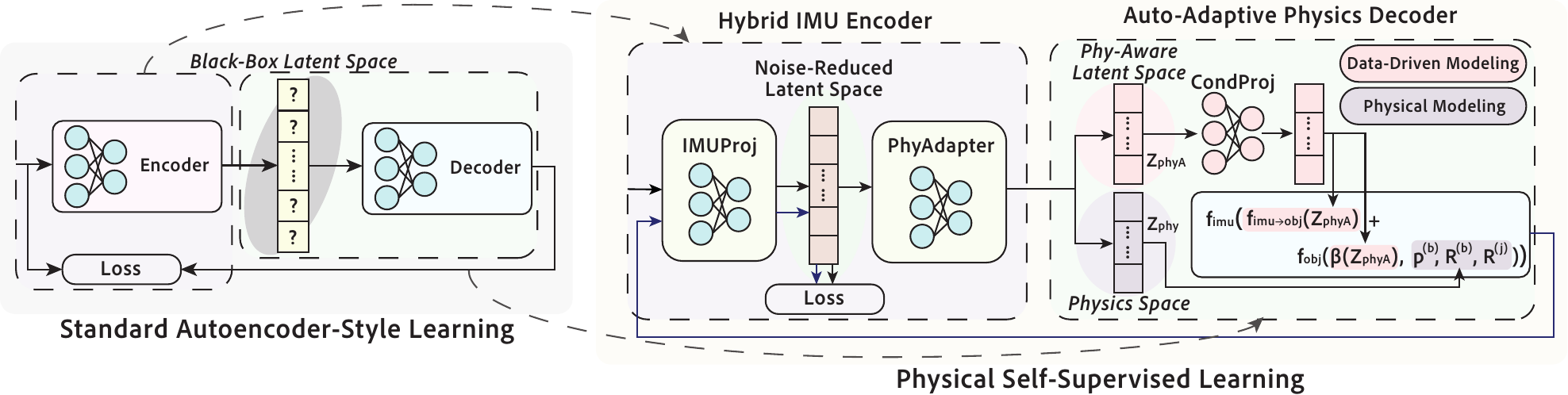}
\caption{Physical Self-Supervised Learning.}
\label{fig:phyLearn_overview}
\end{figure*}
\section{Physical Self-Supervised Learning}~\label{sec:technical}
This section presents our physical self-Supervised learning framework for IMU sensing tasks. We first introduce the overall paradigm in Section~\ref{sec:paradigm}. Section~\ref{sec:imu_motion} explains how we disentangle sensor motion from object motion. Section~\ref{sec:multiView_tree} describes how the multi-view kinematic tree leverages sparse physical self-supervision, and  Section~\ref{sec:uncertaitny} details our uncertainty-aware physical self-Supervised learning.

\subsection{Basic Physical Self-Supervised Paradigm}\label{sec:paradigm}
\subsubsection{Framework Overview}
Although our proposed physical self-Supervised learning framework still follows an auto-encoder-like structure, this section describes how its three key components—the encoder, decoder, and loss function—are redesigned in our new learning paradigm to enable sensing without manual labels. As illustrated in Figure~\ref{fig:phyLearn_overview}, the input is a temporal sequence of 3D IMU measurements from one or more sensors, which is first processed by the hybrid IMU encoder to infer both the target physical states (e.g., joint motions) and an environment-aware representation of the sensing setup. These outputs are then passed to the Auto-Adaptive Physics Decoder, where the environment-aware representation is further transformed by CondProj layers into learnable environment variables, such as bone lengths and sensor-related parameters. Together with the predicted physical states, these variables are used in parameterized physical equations to reconstruct the corresponding IMU measurements. To reduce the effect of sensor noise, the reconstructed sequence is fed into the encoder again, and the reconstruction objective is enforced in the intermediate latent space rather than directly in the raw data space. Through this training process, the model learns to produce physical states and environment variables that are both physically consistent and robust to real-world sensing noise. During inference, the auto-adaptive physical decoder is no longer needed, while the encoder directly predicts the target physical states and environment variables from the input IMU sequences.
\subsubsection{Auto-Adaptive Physics Decoder}
Unlike common autoencoder families (e.g., variational autoencoders and masked autoencoders), which typically rely on fully learnable DNN decoders to map the latent space back to the data space and often produce latent representations that are difficult to interpret, our auto-adaptive physics decoder instead uses explicit, parameterized physical equations that automatically adapt to the data to model a family of kinematic systems. The parameters of these equations are conditioned on the input IMU readings. This design removes the need to perfectly specify a single underlying physical model and allows physical self-supervised learning to focus on adapting to different sensing environments rather than merely fitting a fixed set of known dynamics.

For clarity, we first consider the IMU kinematics of a point mass, which we use to describe reference-frame transformations between the sensed object and the IMU, given by the following system of ODEs:
\begin{equation}
    \begin{cases}
        \frac{d}{dt}p^{(\text{imu})}(t) = v^{(\text{imu})}(t)\\
        \frac{d}{dt}v^{(\text{imu})}(t) = R^{(\text{imu})}(t)a(t) + g\\
        \frac{d}{dt}R^{(\text{imu})}(t) = R^{(\text{imu})}(t)\hat{\omega}(t)
    \end{cases}
    \label{equ:motion}
\end{equation}

\noindent
where $p^{(\text{imu})}(t)$ and $v^{(\text{imu})}(t)$ denote the position and velocity expressed in the global frame of reference (GFR); $R^{(\text{imu})}(t)$ is the orientation, represented as a rotation matrix that maps vectors from the local frame of reference (LFR) to the GFR; $a(t)$ is the accelerometer measurement; $g$ is the gravity vector; and $\hat{\omega}(t)$ is the skew-symmetric matrix corresponding to the gyroscope measurement.

To construct the auto-adaptive physics decoder, we derive a discrete-time formulation suitable for sampled IMU data by discretizing the continuous-time dynamics in \eqref{equ:motion}. Let $t_k = k\Delta t$ and denote $p^{(\text{imu})}_k = p^{(\text{imu})}(t_k)$, $v^{(\text{imu})}_k = v(^{(\text{imu})}t_k)$, $R^{(\text{imu})}_k = R^{(\text{imu})}(t_k)$, $a_k = a(t_k)$, and $\hat{\omega}_k = \hat{\omega}(t_k)$. Using a first-order (Euler) approximation, we obtain
\begin{equation}
    \begin{cases}
        p^{(\text{imu})}_{k+1} = p^{(\text{imu})}_k + v^{(\text{imu})}_k \Delta t, \\
        v^{(\text{imu})}_{k+1} = v^{(\text{imu})}_k + \big(R^{(\text{imu})}_k a_k + g\big)\Delta t, \\
        R^{(\text{imu})}_{k+1} = R^{(\text{imu})}_k \exp\!\big(\hat{\omega}_k \Delta t\big),
    \end{cases}
    \label{equ:motion_discrete}
\end{equation}

\noindent
where $\exp(\cdot)$ denotes the matrix exponential. For small $\Delta t$, we further approximate $\exp(\hat{\omega}_k \Delta t) \approx I + \hat{\omega}_k \Delta t$.

Therefore, the point-mass forward kinematics can be written as
\begin{equation}
    [a, \omega] = f_{\text{imu}}(p^{(\text{imu})}, R^{(\text{imu})})
\label{equ:imu_forward}
\end{equation}

\noindent
where $f_{\text{imu}}(\cdot)$ denotes the differentiable forward IMU model that maps point-mass kinematics to IMU readings. This model forms one component of our physics decoder, which is implemented based on the discrete-time dynamics in \eqref{equ:motion_discrete}.

In practice, however, many IMU-sensing tasks involve objects with substantially more complex kinematic structure, such as human arms or full-body motion, where each IMU measurement reflects the coupled dynamics of an articulated rigid-body system rather than a single point mass. We model such an articulated object as a kinematic tree~\cite{wittenburg2013dynamics}: a tree-structured assembly of rigid links connected by joints and rooted at a global base joint. As an illustrative example, we use the SMPL kinematic tree~\cite{loper2023smpl} for human pose and motion capture with IMUs. SMPL is a parametric 3D human body model in which body shape is represented by shape parameters $\vec{\beta} \in \mathbb{R}^{10}$, given as coefficients of a PCA body-shape basis. Body pose is defined by the relative 3D rotations of the 23 non-root joints in the kinematic tree. Let $\{R^{(j)}\}_{j=1}^{23}$ denote these joint rotations; in practice, we parameterize them using a 6D continuous representation rather than a 3D minimal one to improve continuity in the rotation space~\cite{zhou2019continuity}. The global translation and orientation of the root joint (i.e., the pelvis) are represented by $\{p^{(b)}, R^{(b)}\}$.

There exists a differentiable forward-kinematics mapping for the SMPL human pose model (and, more generally, other articulated objects) that yields the positions and orientations of all joints in the GFR:
\begin{equation}
    [p^{(j)}, R^{(j)}] = f_{\text{obj}}(\beta, p^{(b)}, R^{(b)}, \{R^{(j)}\})
    \label{equ:obj_forward}
\end{equation}

\noindent
where $z{_\text{phy}} = [p^{(b)}, R^{(b)}, \{R^{(j)}\}_{j=1}^{23}]$ denotes the desired kinematic quantities of the object of interest, explicitly produced by the “physical-space’’ branch of the hybrid IMU encoder. The object-specific parameters $\vec{\beta}(Z_{\text{phyA}})$ (e.g., bone lengths in the human SMPL model) are adapted by a learnable DNN, CondProj, whose outputs depend on the physics-aware latent feature $Z_{\text{phyA}}$, thereby making inference over object parameters explicitly conditioned on the current IMU inputs.

Unfortunately, purely physics-based modeling cannot resolve everything, because the IMU placement and its relative motion with respect to the object of interest are generally unknown and are nearly impossible to model explicitly (e.g., a phone’s movement in a pocket or bag, or a smartwatch’s motion on a wrist over time). Rather than hand-crafting explicit parameterized models for these factors, we adopt a data-driven approach that directly predicts the IMU placement and its relative motion with respect to a joint using a learnable DNN, CondProj, which takes the physics-aware latent feature $Z_{\text{PhyA}}$ as input:
\begin{equation}
    [p^{(\text{imu}\rightarrow \text{obj})}, R^{(\text{imu}\rightarrow \text{obj})}] = f_{\text{imu}\rightarrow \text{obj}}(Z_{\text{PhyA}})
    \label{equ:obj2imu_forward}
\end{equation}

\noindent
We detail how the IMU placement and its relative motion are parameterized with DNNs, and how this relative motion is disentangled from the object’s movement, in Section~\ref{sec:imu_motion}. Therefore, the IMU’s motion in the GFR is given by
\begin{equation}
\begin{aligned}
    [p^{(\text{imu})}, R^{(\text{imu})}] = & f_{\text{imu}\rightarrow \text{obj}}(Z_{\text{PhyA}}) + \\
    & \phantom{aa} f_{\text{obj}}(\beta(Z_{\text{PhyA}}), p^{(b)}, R^{(b)}, R^{(j)})
\end{aligned}
    \label{equ:imuGRF_forward}
\end{equation}

\noindent
which corresponds to composing the object’s motion~\eqref{equ:obj_forward} with the learned IMU’s motion relative to the object~\eqref{equ:obj2imu_forward}. As shown in Figure~\ref{fig:phyLearn_overview}, by further transforming the IMU’s motion into the LFR using the forward IMU model~\eqref{equ:imu_forward}, we thus obtain our auto-adaptive physics decoder.

\subsubsection{Hybrid IMU Encoder}
Although our auto-adaptive physics decoder enables interpretable IMU sensing without manual labels and uses data-driven modeling to avoid inaccurate and labor-intensive hand-crafted physical models, standard autoencoder-style training still relies on reconstruction error in the data space as its primary supervision signal. IMU measurements are notoriously noisy, which makes a purely data-space reconstruction loss much less effective.

Our key idea is therefore to decompose the encoder into two cascaded components: the first produces a noise-reduced latent space that preserves task-relevant information, and the second maps this latent representation into a physics-aware space tailored to the auto-adaptive physics decoder. By enforcing reconstruction in the denoised latent space rather than the raw IMU data space, we mitigate the impact of sensor noise. We refer to these two encoder components as IMUProj and PhyAdapter, respectively. 

To obtain a noise-reduced latent space, we first pre-train IMUProj using the standard masked autoencoder recipe, a strategy that has proven effective for denoising and representation learning on a variety of time-series data and tasks~\cite{liang2024foundation, benidis2022deep, limu_loc}. 
We adopt a miniature ViT architecture~\cite{zhou2021ibot}, reducing the depth from 12 to 6 transformer blocks (12M parameters), and make minor adaptations for IMU data, including masking along both the temporal and sensor dimensions and applying positional embeddings only in the temporal domain. We deliberately omit sensor-wise positional embeddings because sensor placement is not known a priori (e.g., the IMU may be in a phone in a pocket or a watch on a wrist). PhyAdapter is implemented as a shallow MLP that maps features from this noise-reduced latent space into the physics and physics-aware spaces. During physical self-supervised learning, PhyAdapter is trained jointly with our physics decoder, while IMUProj remains frozen.
\subsection{Disentangling IMU and Object Motion}~\label{sec:imu_motion}

During IMU sensing, we typically have no prior knowledge of where the sensors are attached, their initial position and orientation, or how they move relative to the object over time. In the previous section, we modeled this IMU–object relative motion using the learned forward mapping $f_{\text{imu}\rightarrow \text{obj}}(\cdot)$ obtained via a data-driven approach. However, data-driven learning is not a panacea: it remains challenging to disentangle the sensor’s relative motion from the underlying object motion during training.

\begin{figure}[t]
    \centering
    
    \begin{subfigure}{0.45\linewidth}
        \centering
        \includegraphics[width=\linewidth]{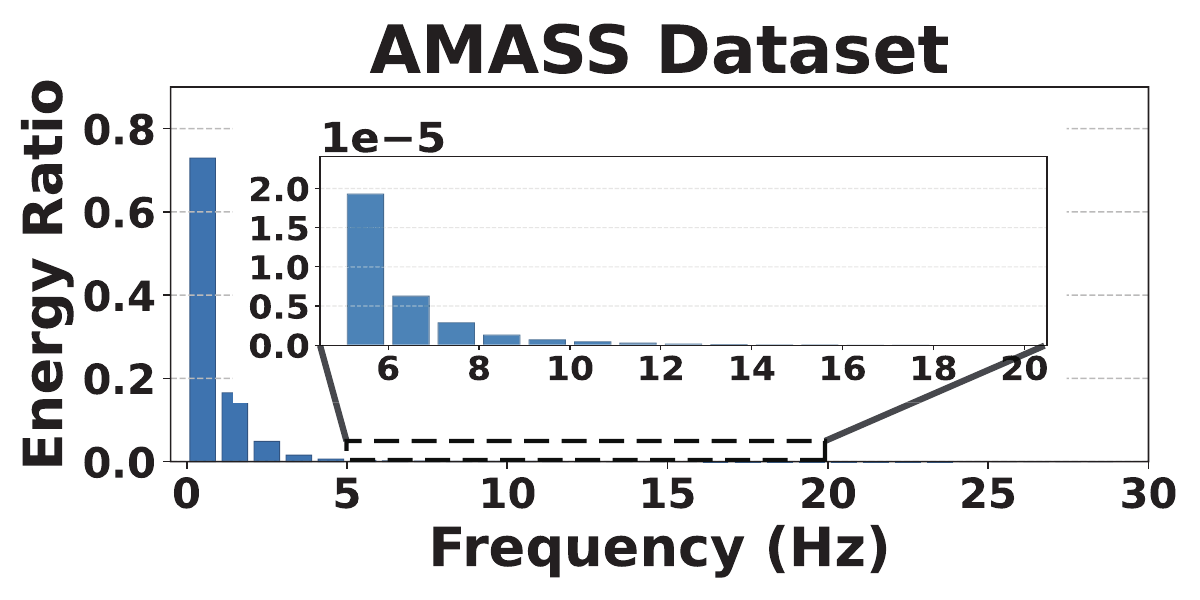}
    \end{subfigure}
    \hfill
    \begin{subfigure}{0.45\linewidth}
        \centering
        \includegraphics[width=\linewidth]{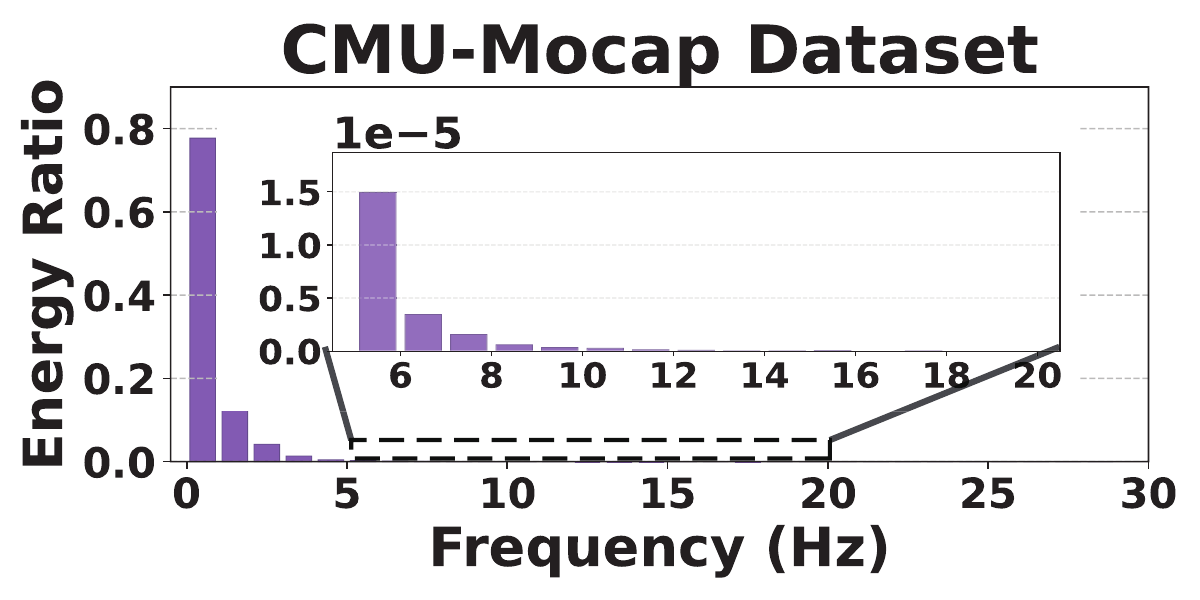}
    \end{subfigure}
    \begin{subfigure}{0.45\linewidth}
        \centering
        \includegraphics[width=\linewidth]{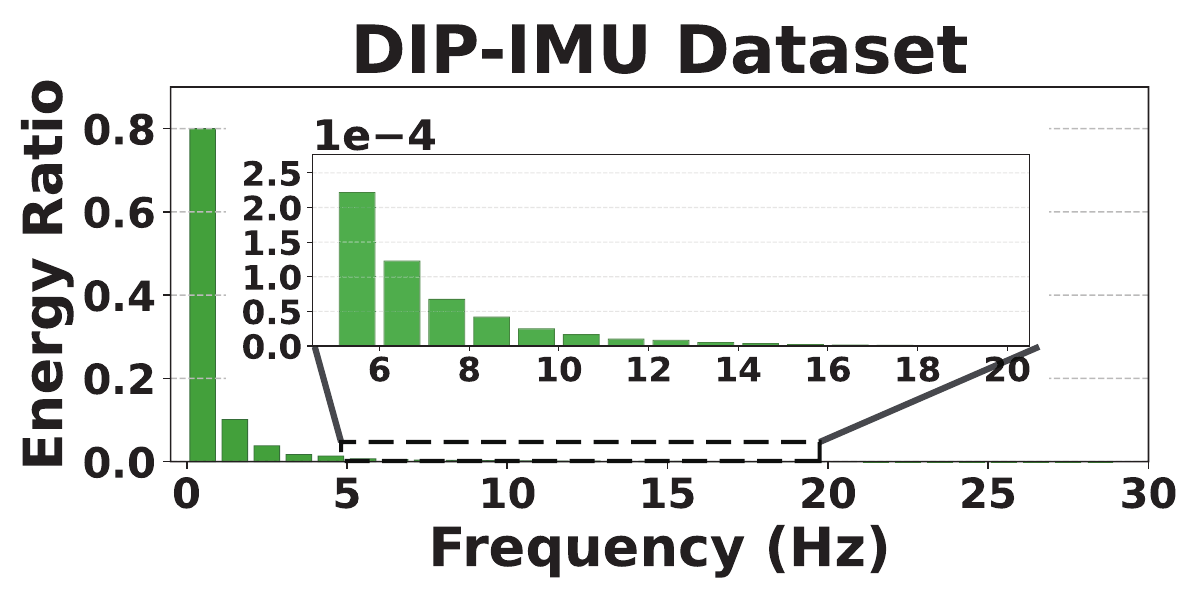}
    \end{subfigure}
    \hfill
    \begin{subfigure}{0.45\linewidth}
        \centering
        \includegraphics[width=\linewidth]{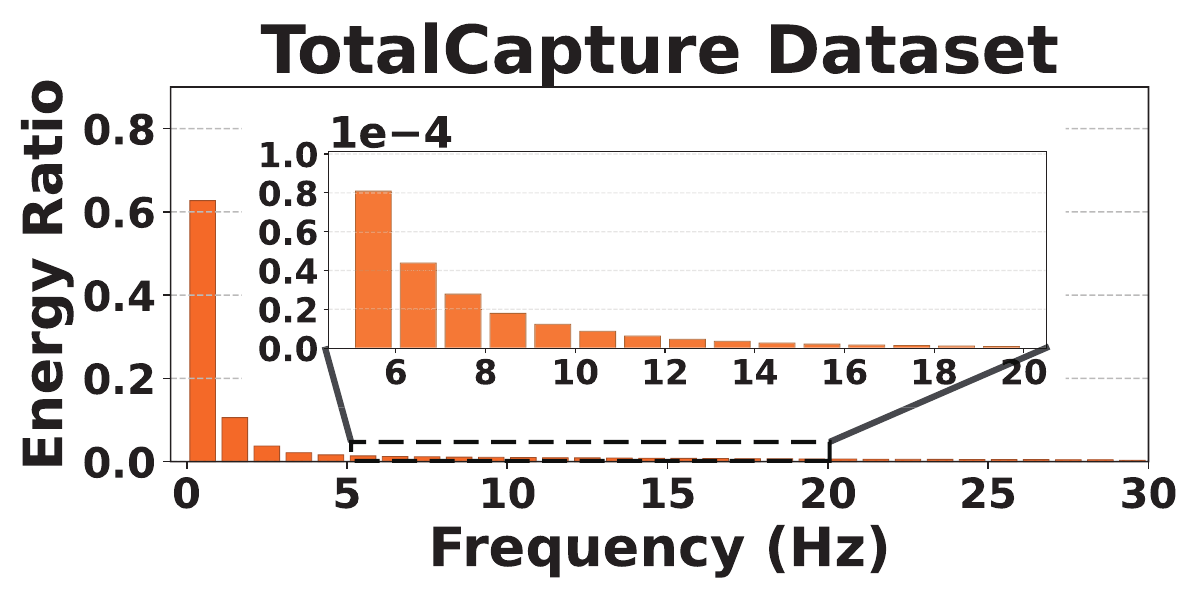}
    \end{subfigure}
    \caption{ Frequency energy distribution of human motion across four widely used public datasets.}
    \vspace{-0.3cm}
    \label{fig:freq-four-dataset}
\end{figure}
To address this challenge, we first structurally decompose the sensor’s relative motion. In typical scenarios, multiple sensors of known device types (e.g., smartphone, smartwatch, earbuds) operate jointly. Each device type is associated with a small set of plausible operating regions on the body. For example, a smartwatch may be worn on the left or right wrist, and a smartphone may be carried in the left/right pocket, held in the left/right hand, or placed in a backpack. We refer to the discrete choice over these regions as the \emph{sensor placement}.

Each candidate placement is associated with an anchor joint in the kinematic-tree model, denoted by $\phi_m$ for the $m$-th IMU. For instance, the left/right wrist joint for a smartwatch on the left/right wrist, and the left/right hip joint for a smartphone in the corresponding trouser pocket. Given a sensor placement with anchor joint $\phi_m$, we decompose the relative motion of the $m$-th IMU (denoted $\text{imu}_m$) into two components: (i) the initial relative kinematic state (position and orientation) with respect to $\phi_m$, $[p_0^{(\text{imu}_m \rightarrow \phi_m)}, R_0^{(\text{imu}_m\rightarrow \phi_m)}]$, and (ii) the time-varying motion relative to this initial state, $[\Delta p_k^{(\text{imu}_m\rightarrow \phi_m)}, \Delta R_k^{(\text{imu}_m\rightarrow \phi_m)}]$. 
Thus, the kinematic state of $\text{imu}_m$ in the global reference frame at time step $k$ can be written as follow, taking the positional part as an example:
\begin{equation}
    p_k^{(\text{imu}_m)} = p_0^{(\text{imu}_m \rightarrow \phi_m)} + \sum_{\tau=1}^t \Delta p_{\tau}^{(\text{imu}_m\rightarrow \phi_m)} + p_k^{(\phi_m)}\label{equ:imuGRF_compose}
\end{equation}

\noindent
where $p_k^{(\phi_m)}$ is obtained from the kinematic-tree forward-kinematics function $f_{\text{obj}}(\cdot)$.

However, the core challenge is to disentangle the sensor’s relative motion from the human body motion. Our key insight is that these two types of motion are more naturally separable in the joint frequency–spatial domain. On the one hand, human motion is typically band-limited, as established in biomechanics and kinesiology~\cite{bigland1981emg,king1984technical,nielsen2016human}. We further confirm this empirically by analyzing several open motion-capture datasets with diverse ground-truth human motion trajectories. As shown in Figure~\ref{fig:freq-four-dataset}, across all these datasets a 25 Hz cutoff frequency captures more than $99\%$ of the human motion energy.

On the other hand, the sensor’s relative motion is spatially bounded. For example, for a smartwatch worn on the wrist, rotation about the wrist joint is typically limited to about $30^\circ$; displacement away from the forearm surface is on the order of a finger’s width (e.g., $\sim$1 cm); and sliding motion along the forearm is confined to a small range (e.g., $\sim$3 cm). Similarly, for a smartphone in a trouser pocket, both its 3D rotation and translation are constrained within a modest angular and spatial range (e.g., $\sim 40^\circ$ and a few centimeters). 
\begin{figure}[t]
    \centering\includegraphics[width=1\linewidth]{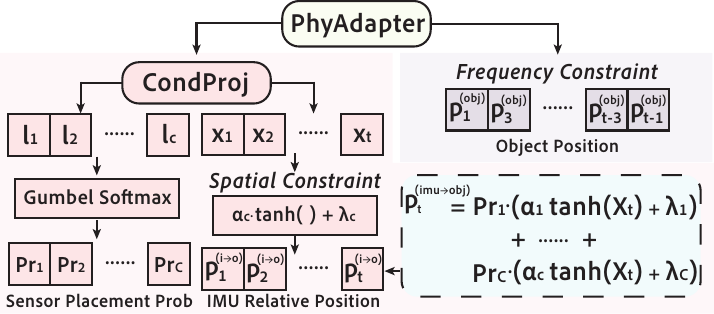}  
        \caption{Probabilistic frequency–spatial constraints for motion disentanglement.}
        \label{fig:DNNOutput}
\end{figure}

In summary, human body motion is frequency-band limited but not strongly spatially constrained (beyond the natural link limits of the SMPL kinematic tree), whereas the sensor’s relative motion is spatially limited but not necessarily band-limited. As shown in Figure~\ref{fig:DNNOutput}, by explicitly encoding these frequency–spatial priors in the output heads of the PhyAdapter and CondProj networks, we substantially reduce the learning complexity and allow physical self-Supervised training to more naturally disentangle object motion from sensor motion.

We enforce the frequency constraint by limiting the output rate over time according to the Nyquist criterion—producing outputs at 50 Hz for a 25 Hz cutoff frequency, while the raw IMU is sampled at 100 Hz. Spatial constraints are imposed via bounded activations, typically a scaled and shifted $\tanh$ of the form $\alpha \cdot \tanh(\cdot) + \lambda$, which restricts predicted translations and rotations to plausible ranges.

However, some of the aforementioned design choices (i.e., the parameters of the spatial constraints and the anchor joint) depend on the sensor placement, which is itself unknown a priori. We therefore treat sensor placement as a learnable variable that is automatically adapted by the corresponding CondProj head in the physics decoder. Consequently, the sensor’s relative motion $f_{\text{imu}\rightarrow \text{obj}}(\cdot)$ is modeled in a probabilistic, expectation-based formulation.
Given each input IMU with a known device type (e.g., watch, phone, or earbuds), CondProj outputs a set of logits over the corresponding candidate placements. To obtain a probability distribution over sensor placements $\{\text{Pr}_c\}$, we apply the Gumbel–Softmax reparameterization~\cite{jang2016categorical} rather than a standard softmax, which encourages approximately discrete placement selection/sampling instead of “averaging’’ over all candidates, and likewise improves the subsequent relative-motion estimation. Moreover, we make the spatial-constraint parameters $\{\alpha, \lambda\}$ in the bounded activation functions depend explicitly on the inferred placement, enabling more fine-grained constraints instead of having to accommodate the worst case (e.g., an IMU in a backpack). The default configurations are listed in Table~\ref{tab:sensor-motion} and are used for all experiments. 
These constraints are chosen based on empirical experience rather than tuned for optimality; deriving tighter, data-driven bounds from large-scale statistics is left for future work. 
\begin{table}[t]\scriptsize
\centering

\renewcommand{\arraystretch}{1.35}
\begin{tabular}{c|l|c|c}
\hline
\textbf{Sensor} & \textbf{Placement} & \textbf{Rotation} & \textbf{Translation} \\
\hline\hline

Earbud &Designated ear & Fixed & Fixed \\
\hline\hline

\multirow{2}{*}{Watch}
    & Left wrist  & \multirow{2}{*}{$\pm30^\circ$ }
                   & \multirow{2}{*}{$\pm1$ cm (away from forearm), } \\
\cline{2-2}
    & Right wrist & (wrist axis)& $\pm3$ cm (along forearm)\\
\hline\hline

\multirow{5}{*}{Phone}
    & Left hand  & \multirow{2}{*}{$0^\circ$ }
                  & \multirow{2}{*}{$\pm0.1$ cm} \\
\cline{2-2}
    & Right hand & & \\
\cline{2-4}

    & Left trouser pocket   & \multirow{2}{*}{$\pm40^\circ$}
                  & \multirow{2}{*}{$\pm3$ cm} \\
\cline{2-2}
    & Right trouser pocket  &  (3 axes)& \\
\cline{2-4}

    & Backpack   & Unlimited & $\pm10$ cm \\
\hline
\end{tabular}
\caption{Default spatial limitations and configurations for each sensor type and its candidate placements.}
\vspace{-1cm}
\label{tab:sensor-motion}
\end{table}
Putting this together, the IMU’s relative motion is estimated as follows, illustrated here for the translational component at time step $k$.
\begin{equation}
\begin{aligned}
    \Delta p_{k,c}^{\text{imu}_m\rightarrow\phi_{m,c}} &= \alpha_c\cdot \tanh(\Delta x_{k}^{\text{imu}_m}) + \lambda_c \\
    \Delta p_{k}^{\text{imu}_m\rightarrow\text{obj}} &=\sum_{c=1}^C \text{Pr}_c\cdot \Delta p_{k,c}^{\text{imu}\rightarrow\phi_{m,c}}
\end{aligned}
    \label{equ:imu2obj_expect}
\end{equation}

\noindent
where $\Delta x_{k}^{\text{imu}_m}$ is the CondProj output for $\Delta p_{k}^{\text{imu}\rightarrow\text{obj}}$ before the activation function, and $\phi_{m,c}$ denotes the anchor joint associated with $\text{imu}_m$ under the $c$-th placement.
In this way, the IMU–object relative-motion function $f_{\text{imu}\rightarrow\text{obj}}(\cdot)$ is constructed so that sensor motion is naturally disentangled from object motion through the imposed probabilistic frequency–spatial constraints.

\subsection{Multi-View Kinematic Tree}\label{sec:multiView_tree}

Even with known sensor deployments and sensors tightly attached to the object of interest, IMU-based tasks such as full-body motion capture with only 2–6 IMUs remain highly challenging due to their inherently ill-posed nature, i.e., the number of kinematic states of interest far exceeds the number of sensed states. This difficulty is further amplified by our goal of achieving such sensing without any manual labels.

Empirically, we find that directly using a standard kinematic tree as the object/human forward kinematics model $f_{\text{obj}}(\cdot)$ leads to poor performance, especially when fewer than 6 IMUs are available. The key reason is that physical self-supervised learning imposes a very different supervision pattern from conventional fully supervised training or fine-tuning.
In a standard kinematic tree, multiple kinematic chains are formed from the root joint (e.g., the pelvis) to the leaf joints. Under supervised learning, every joint is directly supervised, so each kinematic chain receives dense learning signals along its entire length. 
In contrast, under physical self-supervised learning, supervision is available only at anchor joints where IMUs are attached. As a result, many joints may receive little or no effective supervision—particularly when anchor joints are not leaves, or when some kinematic chains contain no anchor joints at all, which is common when using fewer than six IMUs for human pose estimation.

\begin{figure}[t]
\centering\includegraphics[width=1\linewidth]{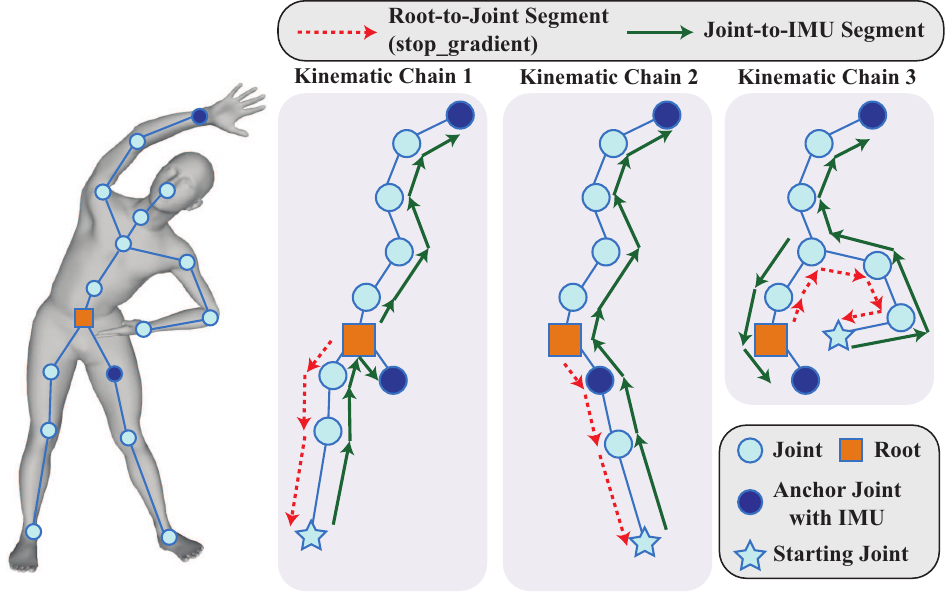}  
        \caption{Multi-view kinematic tree. Three “detour’’ kinematic chains are shown, originating from the right foot, left foot, and left wrist.}
        \vspace{-0.5cm}
        \label{fig:multiView_tree}
\end{figure}
To address this, we propose a \emph{multi-view kinematic tree} with an associated forward-kinematics function that more effectively exploits and propagates physics-based supervision to all joints. Unlike a conventional kinematic tree, which defines kinematic chains only from the root to the leaves, the multi-view kinematic tree constructs chains from every joint to every IMU-anchored joint. Thus, for a model with $N$ joints and $M$ attached IMUs, the multi-view kinematic tree yields $N \times M$ kinematic chains, each terminating at an IMU anchor joint.

We denote $f_{\text{obj}}^{(j,\phi_m)}(\cdot)$ as the forward-kinematics mapping along the chain from joint $j$ to anchor joint $\phi_m$ in the global reference frame (GFR). A issue is that the global kinematic state (i.e., the object’s global translation and orientation over time in the GFR) is typically represented only at the root joint. Naïvely replicating global states at every joint would introduce redundant variables and additional consistency constraints.
Instead, we introduce a “detour’’ kinematic chain that is split into two segments: the first segment runs from the root to joint $j$, denoted $f_{\text{obj}}^{(\text{root}\rightarrow j)}(\cdot)$, and the second segment from joint $j$ to $\phi_m$, denoted $f_{\text{obj}}^{(j\rightarrow\phi_m)}(\cdot)$. The first segment is used to express the global kinematic state from the perspective of joint $j$.

However, naively using this two-segment chain still poses challenges for supervision propagation. As illustrated in Figure~\ref{fig:multiView_tree}, the root-to-joint segment (shown in red) can partially overlap, with opposite direction, the joint-to-IMU segment. Since this forward mapping is used as part of a differentiable computation graph, gradients flowing along these overlapping segments can cancel out, weakening the effective supervision.
To address this, we insert a \texttt{stop\_gradient} operation on the first segment: the root-to-joint path is used only to propagate the global kinematic state in the forward pass, but is excluded from gradient backpropagation. Note that when the joint coincides with the root, neither the “detour’’ kinematic chain nor the \texttt{stop\_gradient} segment is needed; the global kinematic state can still be directly and correctly supervised.
Therefore, the multi-view forward-kinematics function $f_{\text{obj}}(\beta, p^{(b)}, R^{(b)}, \{R^{(j)}\})$ is defined as
\vspace{-0.2cm}
\begin{equation}
    f_{\text{obj}}^{(j,\phi_m)}(\cdot)= \texttt{stop\_grad}(f_{\text{obj}}^{(\text{root}\rightarrow j)}(\cdot)) + f_{\text{obj}}^{(j\rightarrow \phi_m)}(\cdot)
    \label{equ:multiView_tree}
\end{equation}
\vspace{-0.3cm}

\noindent
yielding a total of $N \times M$ kinematic chains for an $N$-joint, $M$-IMU system and enabling sparse physics-based supervision to be stably propagated to all parameters of the object model.

\vspace{-0.3cm}
\subsection{Uncertainty-Aware Formulation}\label{sec:uncertaitny}
Despite the above efforts, IMU sensing remains challenging due to the underdetermined nature of the problem when only a limited number of IMUs are available. A single IMU time series can correspond to multiple, distinct human motion trajectories, making the solution inherently ambiguous. This uncertainty is intrinsic and cannot be ignored. Moreover, it rarely follows a Gaussian distribution and is often multi-modal or highly irregular.
To address this, we explicitly model joint rotations in the object model as probability distributions and adopt an uncertainty-aware formulation in constructing the multi-view kinematic chains.

While Gaussian distributions are the most common choice in DNN-based formulations~\cite{kingma2013auto}, they are restricted to unimodal behavior. Instead, we propose to directly learn the underlying distribution $P(x)$ without imposing a strong prior. For a random variable $x$ with support $[x_0, x_n]$, its expectation is $\bar{x}=\int_{-\infty}^{+\infty}x\cdot P(x)dx = \int_{x_0}^{x_n}x\cdot P(x)dx$. 
To obtain a tractable formulation, we discretize this range into evenly spaced points ${x_0, x_1, \dots, x_n}$ with interval $\Delta$. Under the discrete constraint $\sum_{i=0}^n P(x_i) = 1$, the estimated regression value becomes $\bar{x} = \sum_{i=0}^n x_i\cdot P(x_i)$.
In other words, a regression task can be reformulated as a classification problem over discretized bins, enabling a flexible distributional representation without imposing restrictive parametric priors.

To enable uncertainty-aware multi-view kinematic chains, we replace all kinematic states $\{\beta, p^{(b)}, R^{(b)}, \{R^{(j)}\}\}$ in the forward-kinematics function $f_{\text{obj}}(\cdot)$ with the discrete distribution formulation described above. Consequently, all multi-view kinematic-chain equations in~\eqref{equ:multiView_tree} are reformulated in a probabilistic manner, yielding a stochastic process reminiscent of a Kalman filter that propagates distributions over the state space. Although the transition equations in the multi-view kinematic chains are linear, the resulting distributions are non-Gaussian. To handle this, we borrow the idea of particle filtering and employ Monte Carlo sampling to realize uncertainty-aware multi-view kinematic chain generation.

However, Monte Carlo sampling from discrete distributions is not differentiable. To overcome this, we again use the Gumbel–Softmax reparameterization trick~\cite{jang2016categorical}, which provides a differentiable approximation to one-hot categorical samples and thus preserves end-to-end trainability.

\section{Experiment Setup}~\label{sec:exp_setup}
\vspace{-0.6cm}
\subsection{Data Collection}
\vspace{-0.2cm}
\begin{figure}[h]
  \centering
  \vspace{-0.3cm}
  \includegraphics[width=1.5\linewidth]{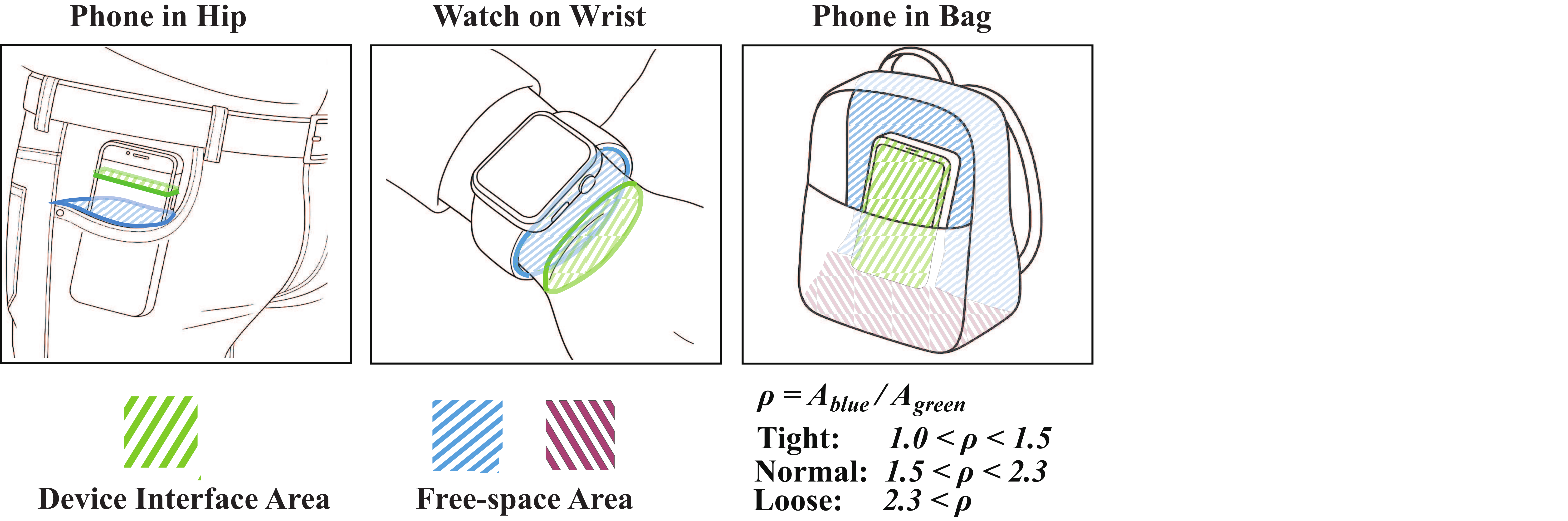}
  \vspace{-0.2cm}
  \caption{ Definition of wearing tightness. On a cross-sectional plane, free-space area ($A_{\text{blue}}$) and device–interface area ($A_{\text{green}}$) define $\rho = A_{\text{blue}}/A_{\text{green}}$, categorized as Tight ($1.0<\rho<1.5$), Normal ($1.5<\rho<2.3$), or Loose ($\rho>2.3$). 
 }
 \vspace{-0.2cm}

  \label{fig:loose-tight-normal}
\end{figure}

\begin{figure*}[htbp]
        \centering
\vspace{-0.2cm}
        \begin{subfigure}{0.24\textwidth}
            \centering
         \includegraphics[width=\textwidth]{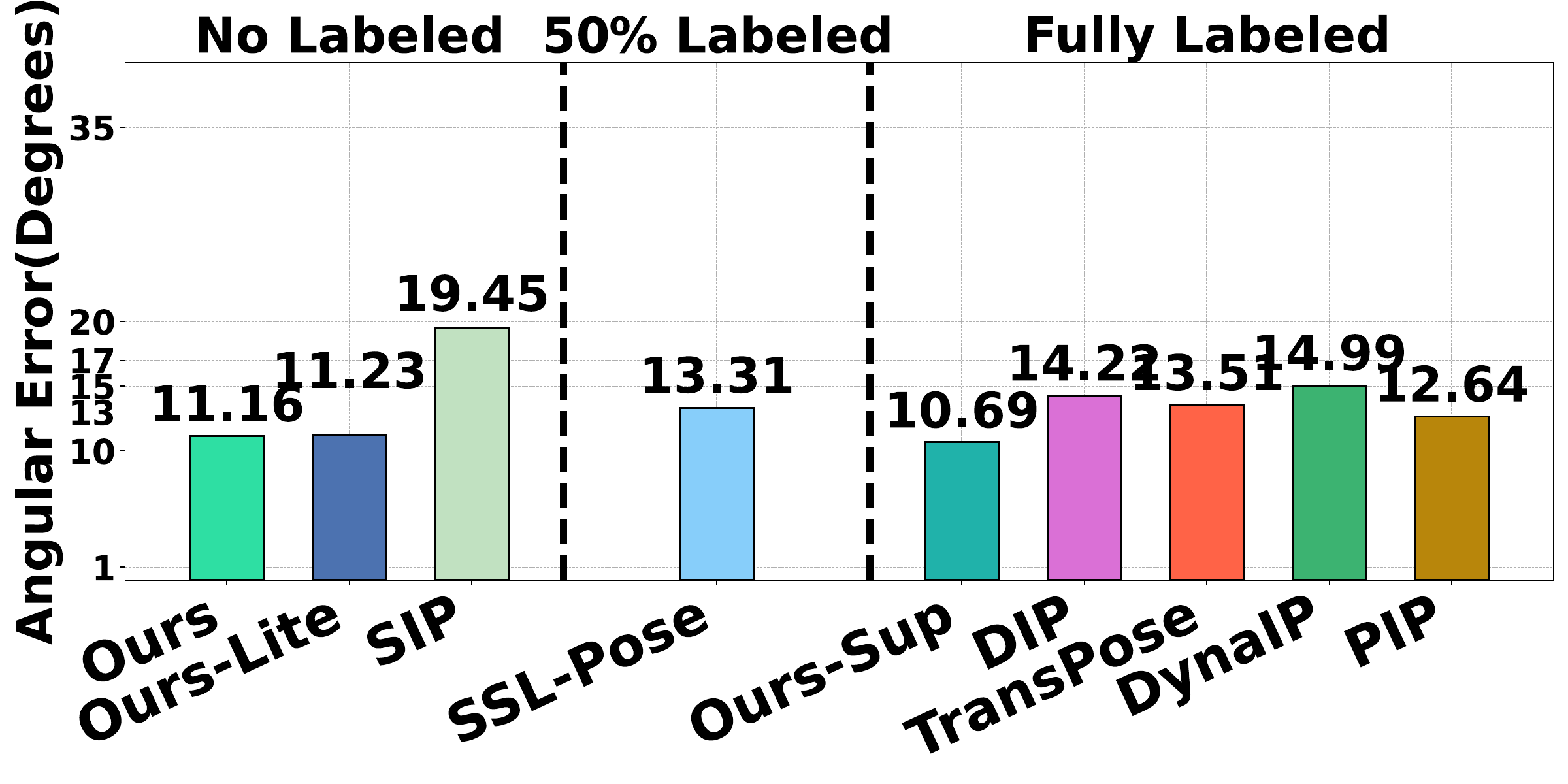}\vspace{-0.3cm}
        \caption{Tight}
        \end{subfigure}
        \begin{subfigure}{0.24\textwidth}
            \centering
         \includegraphics[width=\textwidth]{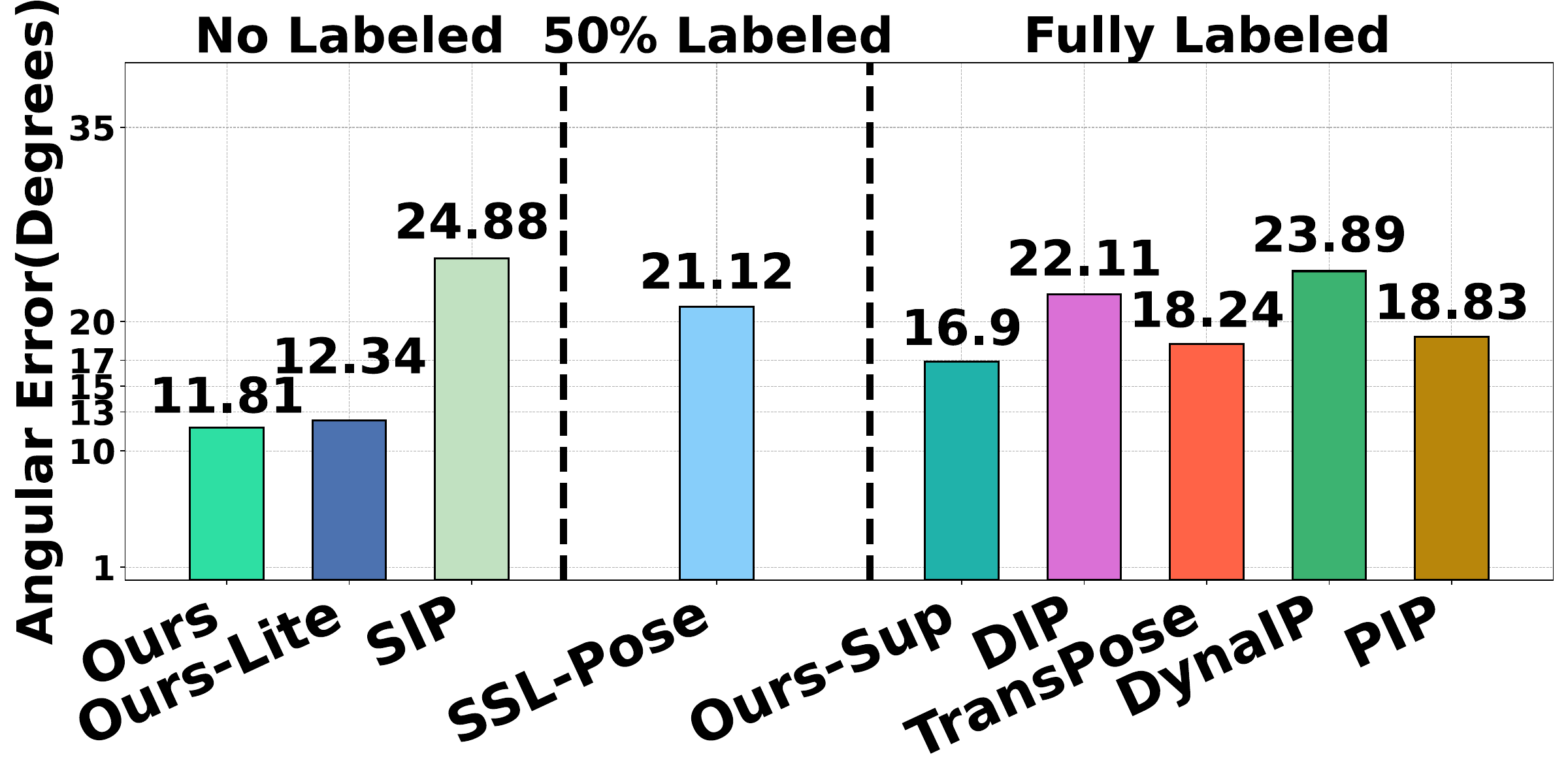}\vspace{-0.3cm}
        \caption{Normal}
        \end{subfigure}
        \begin{subfigure}{0.24\textwidth}
            \centering
         \includegraphics[width=\textwidth]{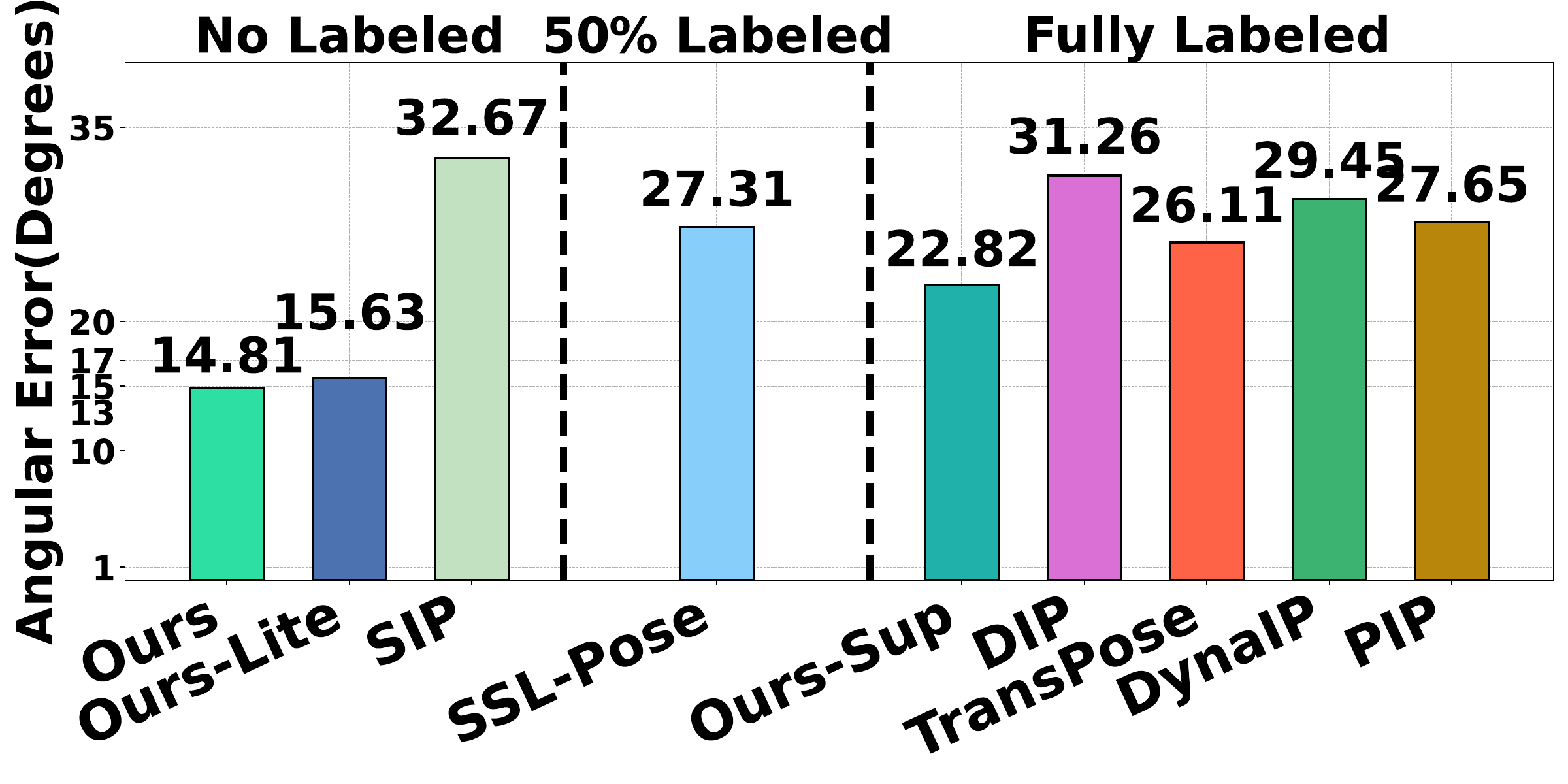}\vspace{-0.3cm}
        \caption{Loose}
        \end{subfigure}
                \begin{subfigure}{0.24\textwidth}
            \centering
         \includegraphics[width=\textwidth]{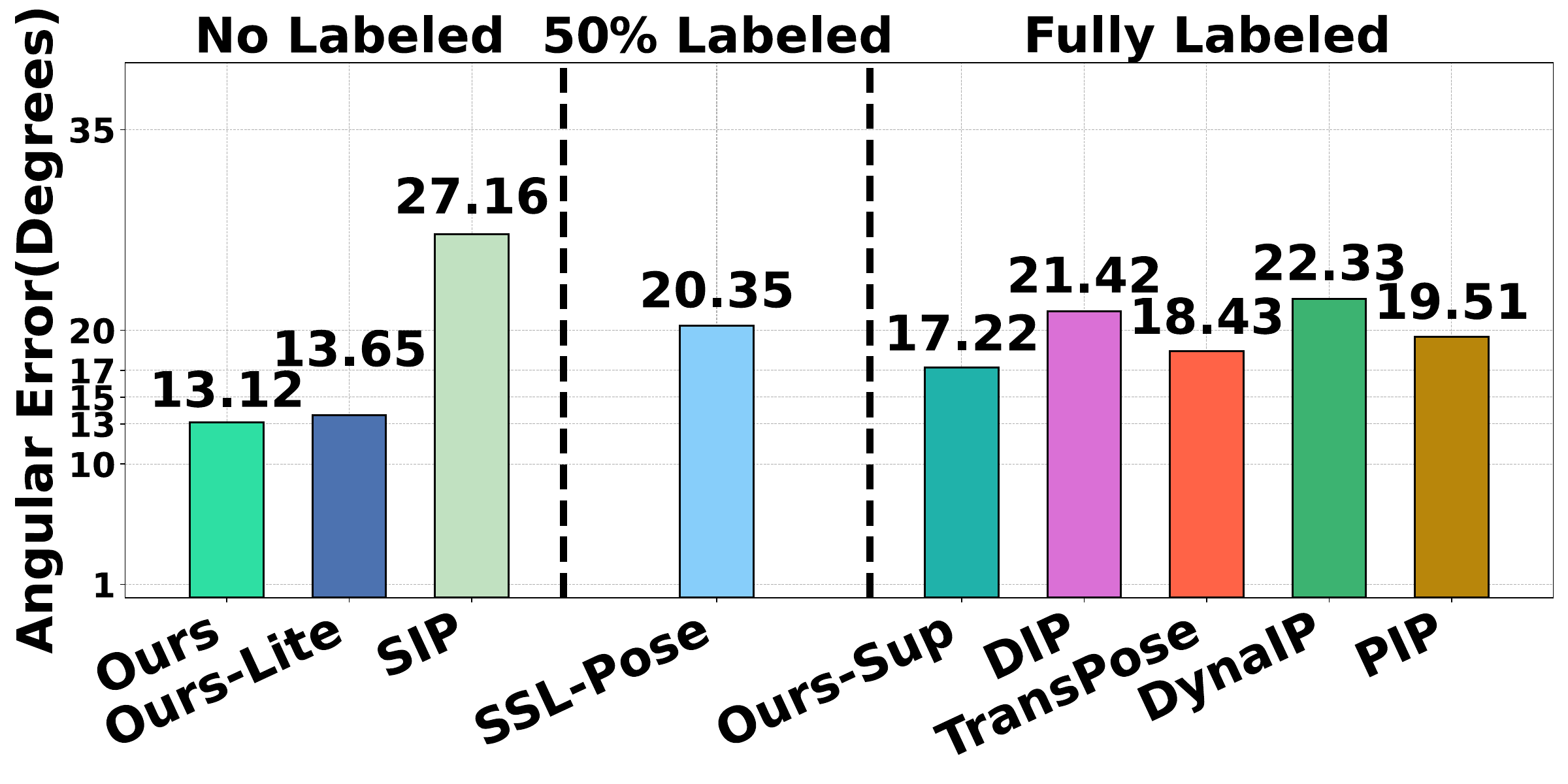}\vspace{-0.3cm}
        \caption{Mixed Data}
        \end{subfigure}
        \vspace{-0.45cm}
        \caption{Angular error across different tightness conditions. All other baselines assume ground-truth sensor placements and skeleton parameters.}

\vspace{-0.3cm}
\label{fig:overall_mocap}
    \end{figure*}

\begin{figure*}[htbp]
\begin{subfigure}{0.3\textwidth}
        \centering
        \includegraphics[width=\textwidth]{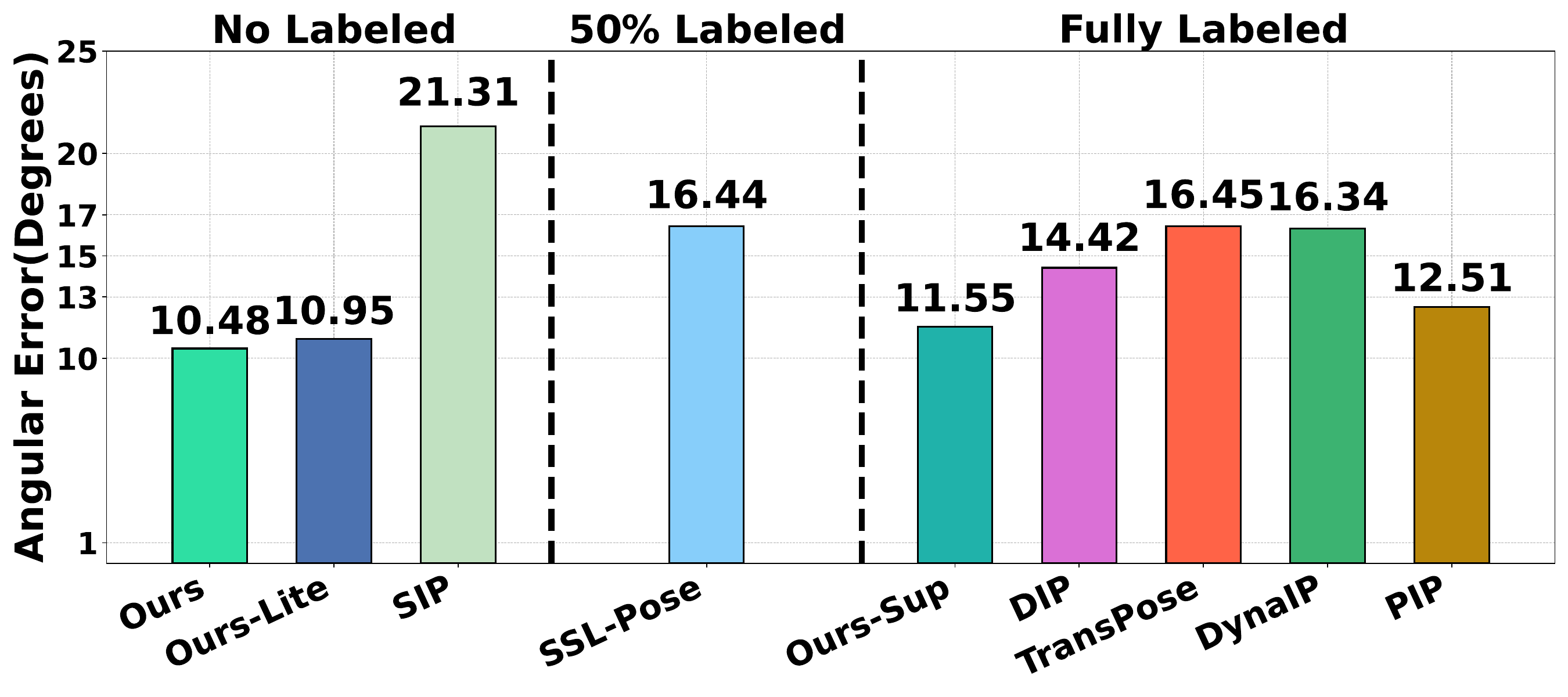}\vspace{-0.3cm}
        \caption{TotalCapture}\vspace{-0.2cm}

    \end{subfigure}\hfill
    \begin{subfigure}{0.3\textwidth}
        \centering
        \includegraphics[width=\textwidth]{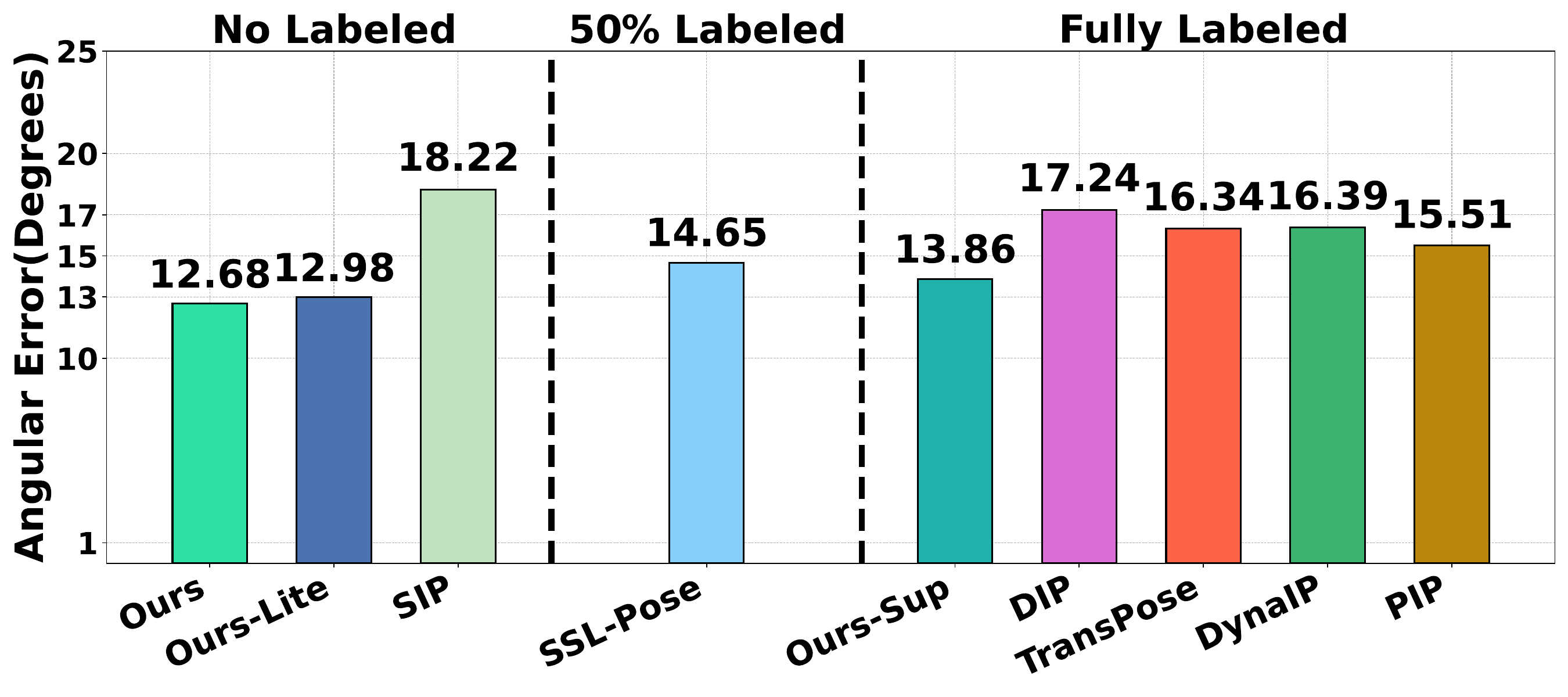}\vspace{-0.3cm}
        \caption{DIP-IMU}\vspace{-0.2cm}
    \end{subfigure}\hfill
    \begin{subfigure}{0.3\textwidth}
        \centering
        \includegraphics[width=\textwidth]{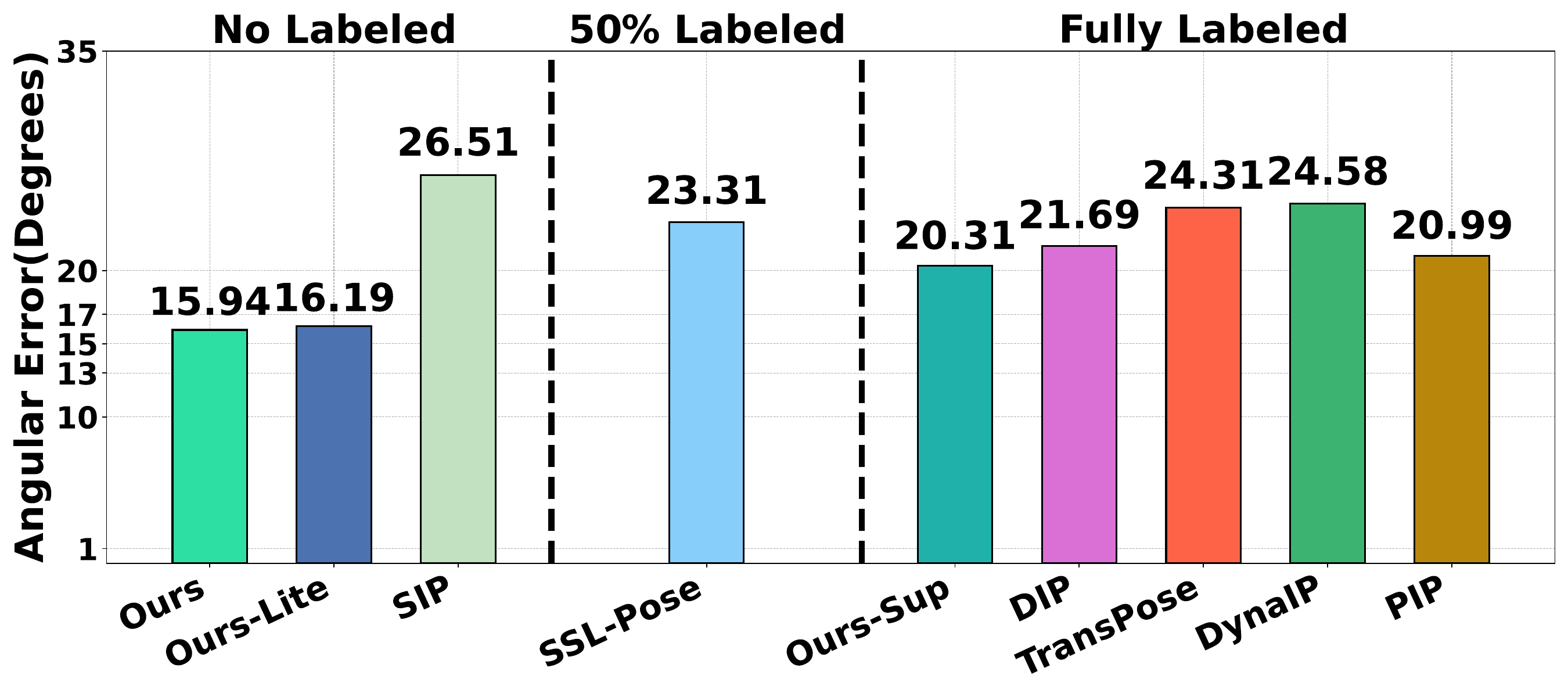}\vspace{-0.3cm}
        \caption{Nymeria}\vspace{-0.2cm}
    \end{subfigure}\hfill
    \vspace{-0.2cm}
    \caption{Angular error on datasets with 6 sensors.}
\vspace{-0.3cm}
    \label{fig:overall_mocap2}
\end{figure*}

\begin{figure}[htb]
\vspace{-0.1cm}
 \includegraphics[width=0.45\textwidth]{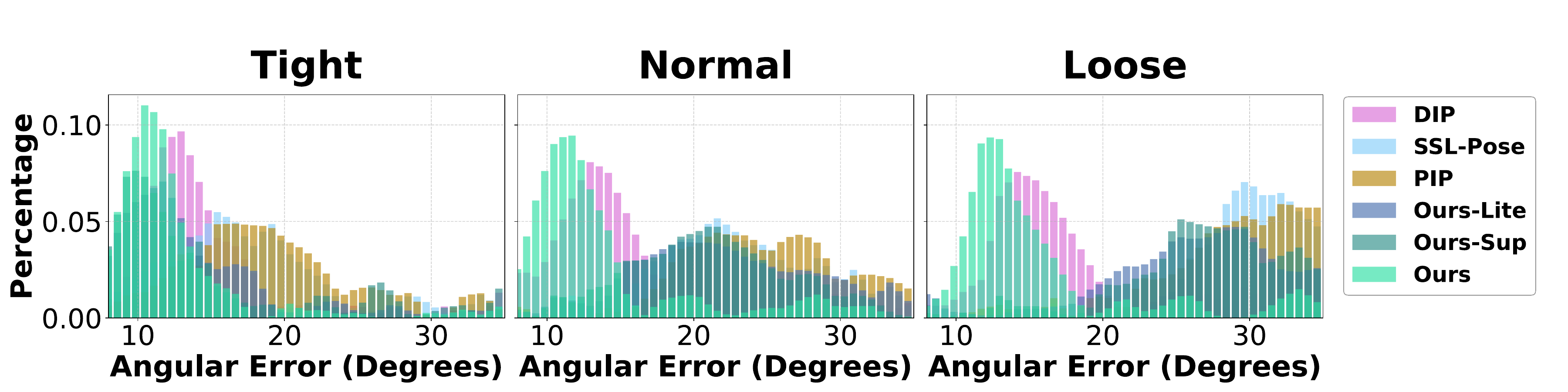}

\vspace{-0.4cm}\caption{Angular error distributions shown for tight, normal, and loose wearing conditions.}\vspace{-0.55cm}
\label{fig:combine_sensor_distribution}
\end{figure}

\begin{figure}[t]
    \centering
    \hspace{-0.1cm}
    \includegraphics[width=0.9\linewidth]{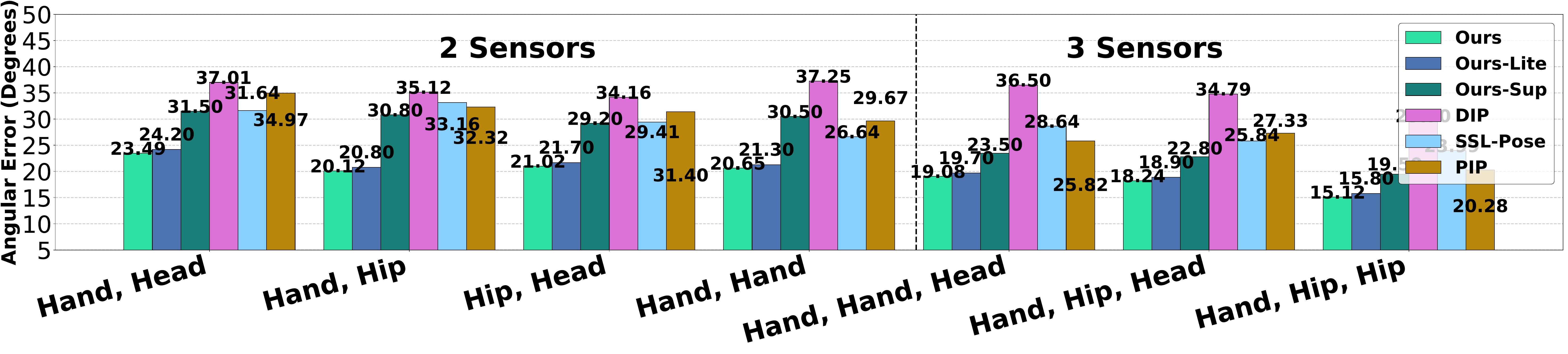}
    \vspace{-0.1cm}
    \caption{ Angular errors under popular sensor placement combinations.}\vspace{-0.8cm}

    \label{fig:popular_combine}
\end{figure}

\noindent
\textbf{Data Collection and Wearing Conditions.}
We recruited four participants.
In the Motion Capture (MoCap) setting, we instrumented six limb segments with Xsens IMUs. Participants also wore a smartwatch on the wrist, carried a smartphone in a hip pocket, and used an earbud on the head. Ground-truth motion was recorded with four Kinect v2 sensors positioned around the capture volume.

In the Inertial Tracking setting, the participants carried a smartphone and IMU streams were recorded as participants followed predefined paths. The waypoint timestamps provided reference positions for temporal and spatial alignment.

To capture how sensor motion affects downstream sensing performance, we also collect data under varying wearing tightness conditions. As illustrated in Fig.~\ref{fig:loose-tight-normal}, each placement is associated with a cross-sectional plane on which we define a free-space region $F$ (blue) and a device–interface region $D$ (green). This plane corresponds to the pocket opening for the phone-at-hip case, the wristband loop for the watch, or the supporting internal face for the backpack. The looseness ratio $\rho = A_F / A_D$ is computed on that plane and used to categorize wearing conditions as tight, normal, or loose.

\begin{table*}[htbp]\tiny
\centering
\setlength{\tabcolsep}{1.0mm}

\resizebox{\textwidth}{!}{
\begin{tabular}{l|c|c|c|c|c|c|c|c|c|c|c|c|c|c|c|c|c}
\toprule
\hline
\multirow{2}{*}{\textbf{Method}}
& \multicolumn{4}{c|}{\textbf{Our Data}}
& \multicolumn{4}{c|}{\textbf{TotalCapture}}
& \multicolumn{4}{c|}{\textbf{DIP-IMU}}
& \multicolumn{4}{c|}{\textbf{Nymeria}}
& \multirow{2}{*}{\textbf{Label}} \\
\cline{2-17}
 & SIP Err(°) & Ang Err(°) & Pos Err(cm)  & Mesh Err(cm)
 & SIP Err(°) & Ang Err(°) & Pos Err(cm) & Mesh Err(cm)
 & SIP Err(°) & Ang Err(°) & Pos Err(cm) & Mesh Err(cm)
 & SIP Err(°) & Ang Err(°) & Pos Err(cm) & Mesh Err(cm)
 & \\

\hline
\hline

SIP
& 33.84 & 27.16 & 21.43 & 23.91
& 23.29 & 21.31 & 12.62 & 14.85
& 23.45 & 18.22 & 13.62 & 14.37
& 31.45 & 26.51 & 23.59 & 20.46
& No \\

\hline
\hline
\multirow{3}{*}{SSL-Pose}
& 24.90 & 22.54 & 18.62 & 20.25
& 29.90 & 24.54 & 13.46 & 16.25
& 26.64 & 23.60 & 10.48 & 15.72
& 31.21 & 27.54 & 22.33 & 24.61
& 10\% \\\cline{2-18}

& 21.62 & 20.46 & 12.32 & 14.36
& 26.45 & 21.46 & 10.32 & 13.36
& 26.45 & 22.33 & 9.11 & 12.88
& 27.52 & 24.10 & 20.38 & 19.65
& 20\% \\

\cline{2-18}
& 21.32 & 20.35 & 10.11 &12.56
& 19.32 & 16.44 & 8.13 & 9.85
& 16.46 & 14.65 & 7.11 & 9.45
& 24.19 & 23.31 & 14.69 & 16.88
& 50\% \\

\hline
\hline
DIP
& 23.58 & 21.42 & 10.95 & 12.04
& 16.58 & 14.42 & 8.44 & 9.04
& 18.42 & 17.24 & 9.16 & 11.97
& 25.44 & 21.69 & 12.51 & 17.63
& Full \\

\hline

DynaIP
& 24.55 & 22.33 & 13.51 & 11.27
& 19.86 & 16.55 & 12.44 & 12.21
& 17.55 & 16.34 & 6.51 & 7.27
& 27.30 & 24.58 & 16.92 & 18.44
& Full \\

\hline

TransPose
& 19.25 & 18.43 & 9.64 & 10.91
& 14.31 & 16.45 & 8.64 & 7.91
& 15.84 & 16.39 & 8.44 & 9.39
& 30.98 & 24.31 & 18.46 & 19.45
& Full \\

\hline

PIP
& 22.44 & 19.51 & 11.68 & 9.45
& 14.44 & 13.51 & 7.68 & 8.45
& 16.03 & 15.51 & 9.86 & 10.82
& 26.79 & 20.99& 14.51 & 13.24
& Full \\
\hline
Ours-Sup
& 19.35&17.22 & 9.69 & 9.32
& 15.01& 11.55&7.89 &  8.31
& 8.59 & 13.86 & 7.99 & 9.09
& 23.04 & 20.31 & 16.32 & 15.96
& Full \\

\hline
\hline
Ours-Lite
& 15.69&13.65 & 7.58 & 8.90
& 12.69& 10.95&7.31 &  6.59
& 13.91 & 16.54 & 10.06 & 12.31
& 17.64 & 16.19 & 9.56 & 10.44
& No \\

\hline

\textbf{Ours}
& \cellcolor{red!20}15.15 & \cellcolor{red!20}13.12 & \cellcolor{red!20}6.84 & \cellcolor{red!20}7.45
& \cellcolor{red!20}12.66 & \cellcolor{red!20}10.48 & \cellcolor{red!20}5.91 & \cellcolor{red!20}5.99
& \cellcolor{red!20}11.13 & \cellcolor{red!20}12.68 & \cellcolor{red!20}4.69 & \cellcolor{red!20}5.62
& \cellcolor{red!20}16.91 & \cellcolor{red!20}15.94 & \cellcolor{red!20}8.95 & \cellcolor{red!20}9.63
& No \\

\hline
\bottomrule
\end{tabular}
}

\caption{MoCap results with 6 IMUs on three datasets.}
\vspace{-0.8cm}
\label{tab:ovarall_mocap}
\end{table*}

\subsection{Baseline Models and Datasets}
We compare our approach with three types of baselines: (1) physics-based optimization methods that use no labels, (2) self-supervised methods that use partial labels, and (3) fully supervised methods that rely on complete labels.

\subsubsection{Motion Capture}
\begin{itemize}[nosep,leftmargin=*]
    \item \textbf{Sparse IMU Pose  (SIP)~\cite{von2017sparse}: Label-free physics-based optimization that fits a statistical body model to IMU readings to recover motion.}
    \item \textbf{SSL-Pose}~\cite{SSL-Pose}: Self-supervised Transformer, pretrained with a masked autoencoder objective and then fine-tuned.
    \item \textbf{Physical Inertial Poser (PIP)}~\cite{yi2022physical}: Fully supervised RNN with an integrated physics-based motion optimizer.
    \item \textbf{TransPose}~\cite{transpose}: Fully supervised RNN with multi-stage inference.
    \item \textbf{Dynamic Inertial Poser (DynaIP)}~\cite{dynaip}: Fully supervised biRNN with part-based pose estimation.
    \item \textbf{Deep Inertial Poser (DIP)}~\cite{dip}: Fully supervised biRNN with synthetic IMU data augmentation.
\end{itemize}
Other than our own data collection, we also evaluate on three public MoCap datasets, DIP-IMU, TotalCapture, and Nymeria, a large-scale daily-activity dataset, which provide synchronized IMU streams and ground-truth poses over diverse motions. We report four standard metrics:
(1) SIP Error: mean orientation error of upper arms and legs in the global frame (°);
(2) Angular Error: mean joint rotation deviation (°);
(3) Position Error: mean Euclidean joint-position distance (cm);
(4) Mesh Error: mean vertex distance between reconstructed and ground-truth meshes (cm).

\vspace{-0.2cm}
\subsubsection{Inertial Tracking}
\begin{itemize}[nosep,leftmargin=*]
    \item \textbf{MUSE}~\cite{shen2018closing}: Label-free, sensor-fusion method that prioritizes magnetometer measurements over gravity for orientation tracking.

    \item \textbf{PDR}~\cite{jimenez2009comparison}: Label-free, step-based pedestrian dead reckoning approach.

    \item \textbf{EKF}: Label-free, classical kalman filtering-based method that recursively estimates position and orientation.

    \item \textbf{LIMU-BERT}~\cite{limu_loc}:Self-supervised BERT-like architecture with pre-training and fine-tuning.

    \item \textbf{SSLHAR-LOC}~\cite{rahimi2021self}: Self-supervised CNN, pretrained with a masked autoencoder objective and then fine-tuned.

    \item \textbf{RoNIN}~\cite{herath2020ronin}:
    Fully supervised hybrid LSTM/TCN model.

    \item \textbf{CTIN}~\cite{rao2022ctin}:
    Fully supervised hybrid ResNet/Transformer.

    \item \textbf{TLIO}~\cite{liu2020tlio}:
    Fully supervised hybrid LSTM/TCN integrated with an Extended Kalman Filter.

    \item \textbf{IONet}~\cite{chen2018ionet}:
    Fully supervised RNN model.

\end{itemize}
Other than our own data collection, we also evaluate on two public datasets, \textbf{SHL}~\cite{shl} and \textbf{OxIOD}~\cite{oxiod}, which cover diverse motion patterns and environments for evaluating generalization. We use \textit{Location Error} as evaluation metric.
\vspace{-0.3cm}

\subsubsection{Our Variants.}
We additionally include two variants of our method for comprehensive comparison:
\begin{itemize}[nosep,leftmargin=*]
    \item \textbf{Ours-Sup}: A supervised variant of our model trained fully with pose labels (without sensor relative motion modeling).
    \item \textbf{Ours-Lite}: A lightweight compressed variant for efficient inference on embedded devices, obtained by reducing Transformer blocks from 6 to 3, attention heads from 8 to 4, latent dimension from 512 to 256, and using FP16 instead of FP32.
\end{itemize}

\vspace{-0.3cm}
\section{Evaluation}~\label{sec:evaluation}
In this section, we evaluate our method on two tasks: motion capture and inertial tracking. We first present overall performance, then examine cross-domain generalization. We next provide an ablation study of our physical self-supervised learning, and finally report execution time and results on downstream task adaptation.
\vspace{-0.2cm}
\subsection{Overall Performance}
\begin{figure*}[!htb]
    \vspace{-0.5cm}
    \centering
    \begin{subfigure}{0.26\textwidth}
        \centering
        \includegraphics[width=\textwidth]{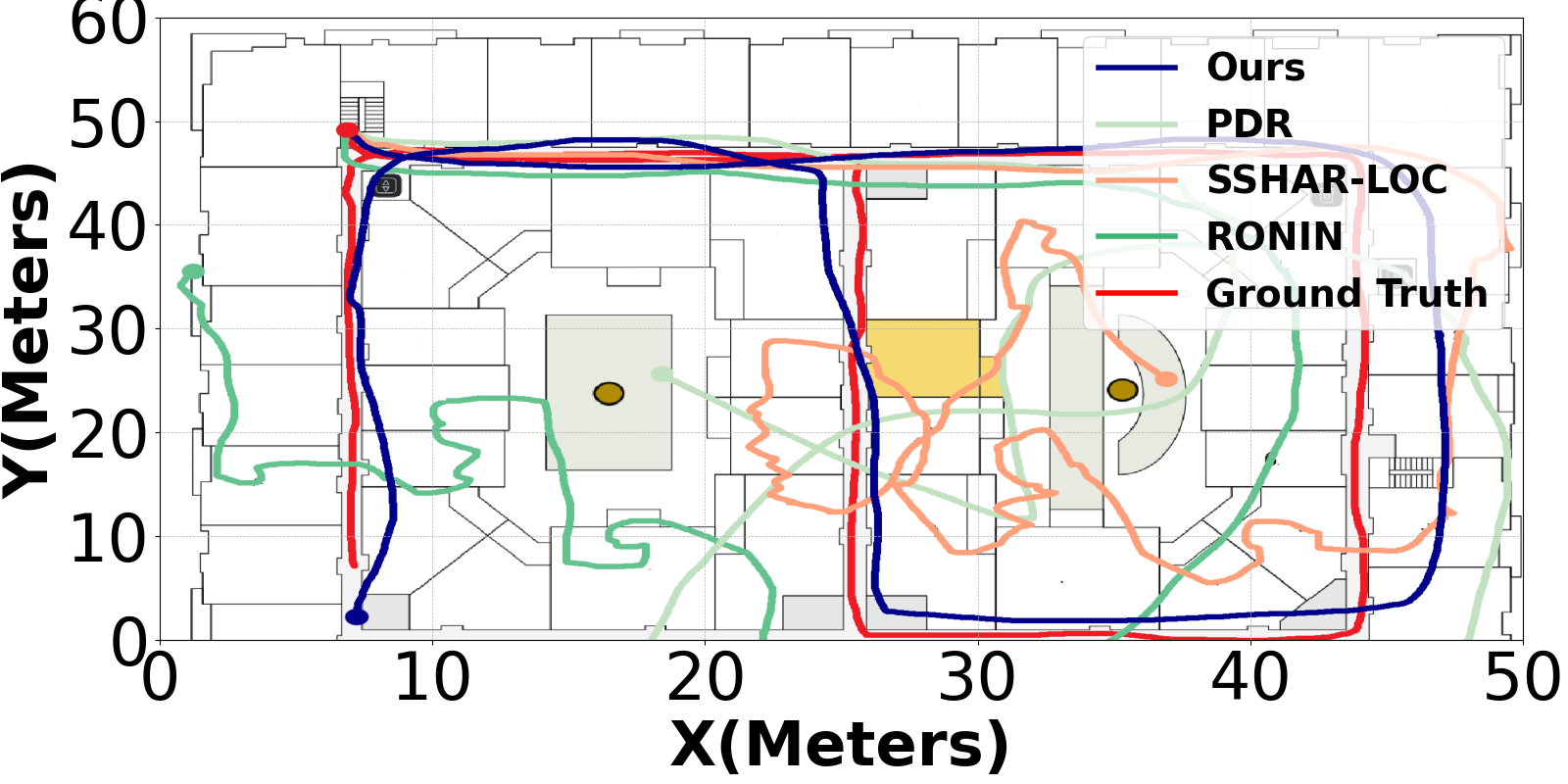}
        \caption{Walking Outdoor}
        \label{fig:walking_indoor}
    \end{subfigure}\hfill
    \begin{subfigure}{0.26\textwidth}
        \centering
        \includegraphics[width=\textwidth]{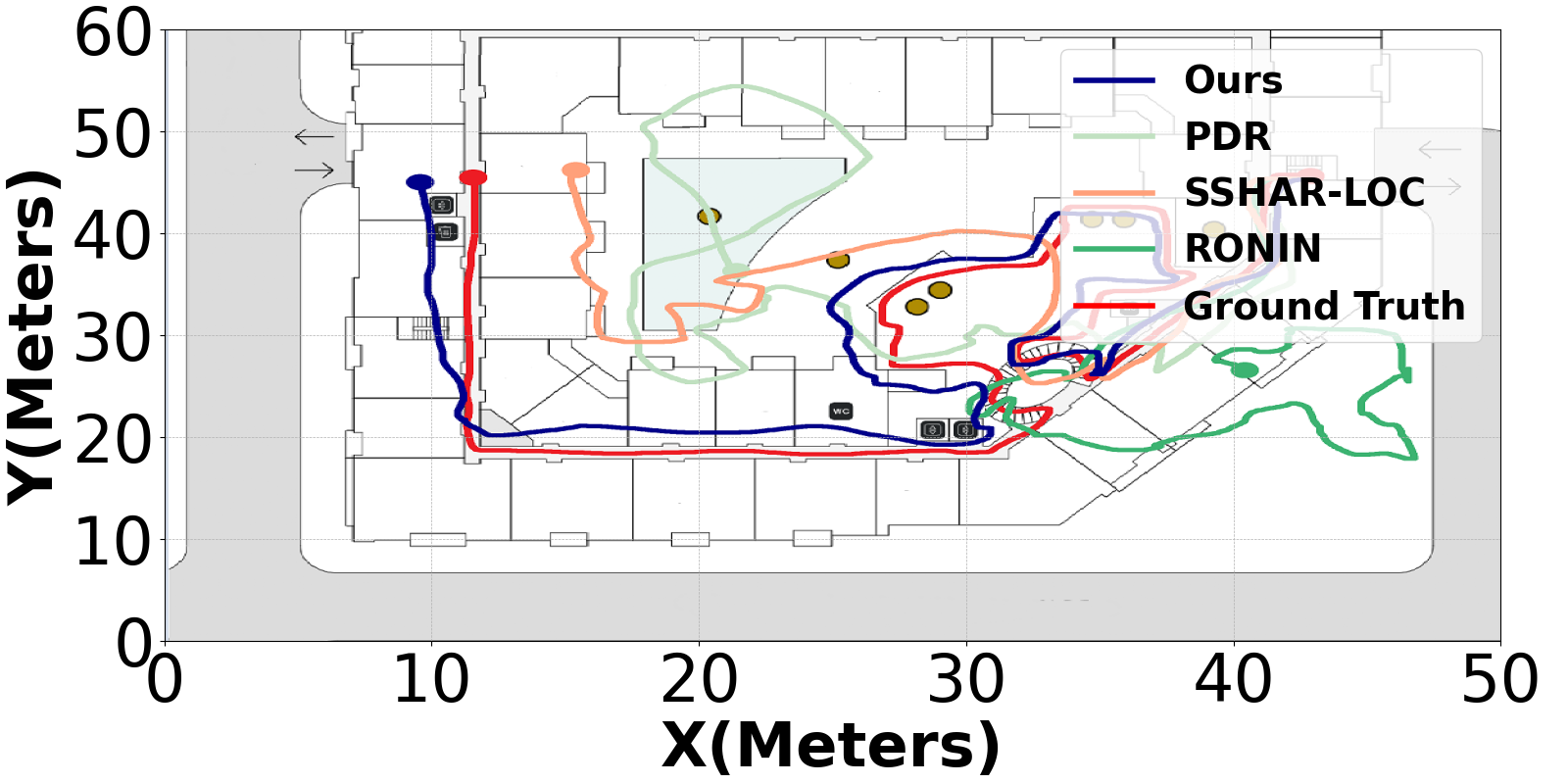}
        \caption{Running Indoor}
        \label{fig:running_indoor}
    \end{subfigure}\hfill
    \begin{subfigure}{0.26\textwidth}
        \centering
        \includegraphics[width=\textwidth]{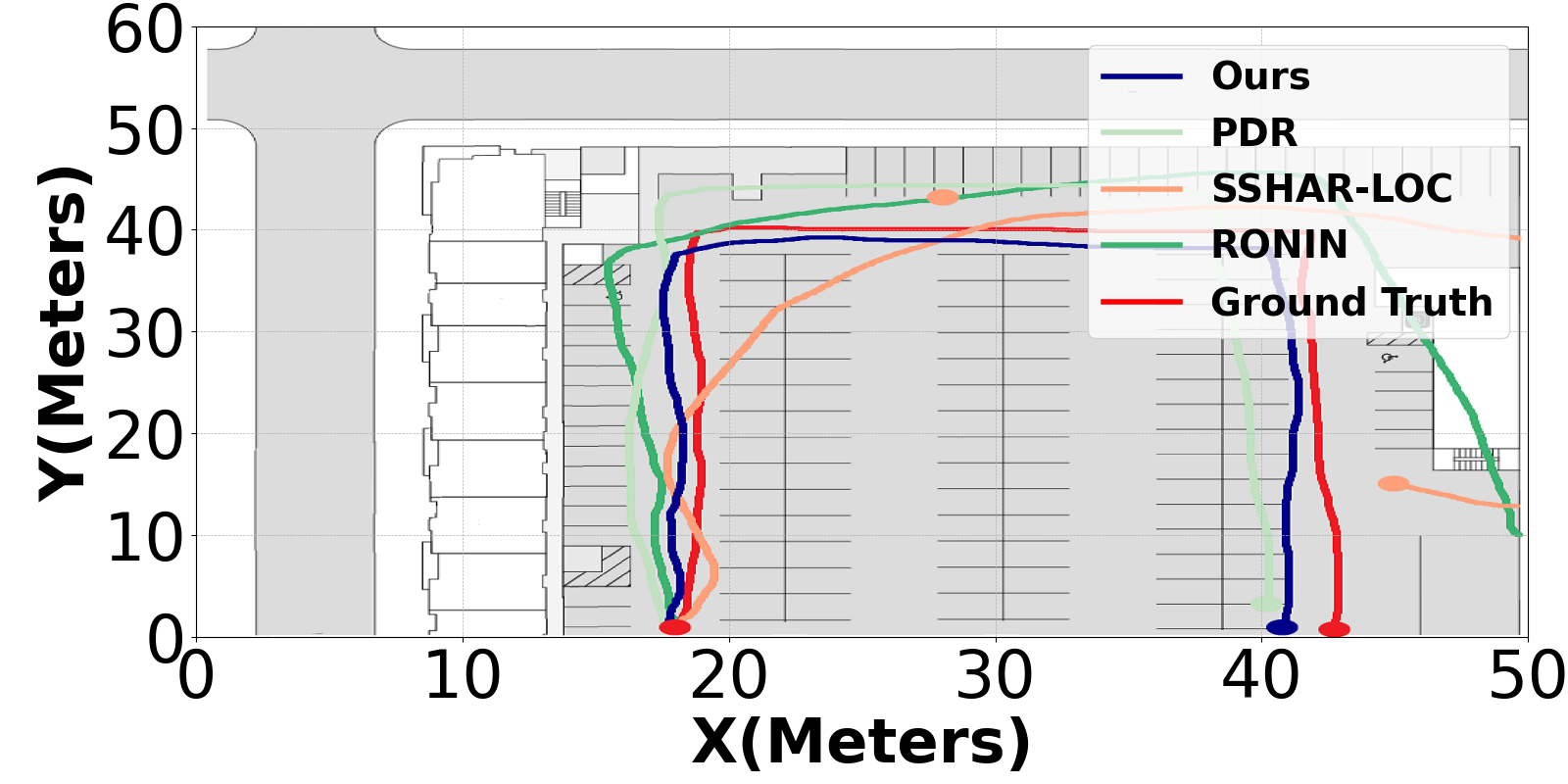}
        \caption{Running Outdoor}
        \label{fig:running_outdoor}
    \end{subfigure}

        \vspace{-0.3cm}
    \caption{Inertial tracking trajectory estimation. Baselines diverge within minutes due to accumulated error.}
            \vspace{-0.3cm}
    \label{fig:maps}

\end{figure*}
\begin{figure*}[htbp]
\begin{subfigure}{0.23\textwidth}
        \centering
        \includegraphics[width=\textwidth]{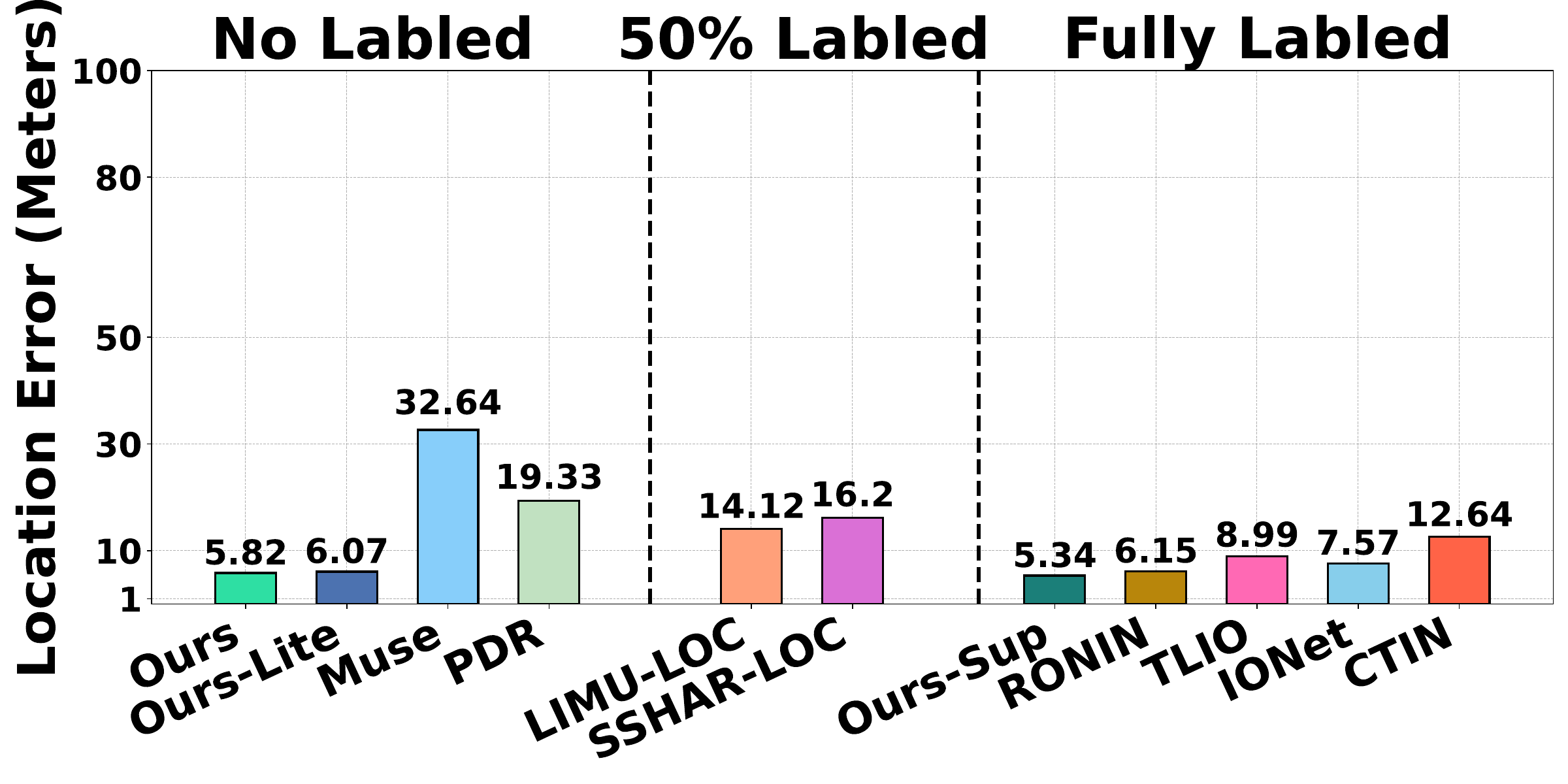}
        \vspace{-0.6cm}
        \caption{Tight}
        \label{}
    \end{subfigure}\hfill
\begin{subfigure}{0.23\textwidth}
        \centering
        \includegraphics[width=\textwidth]{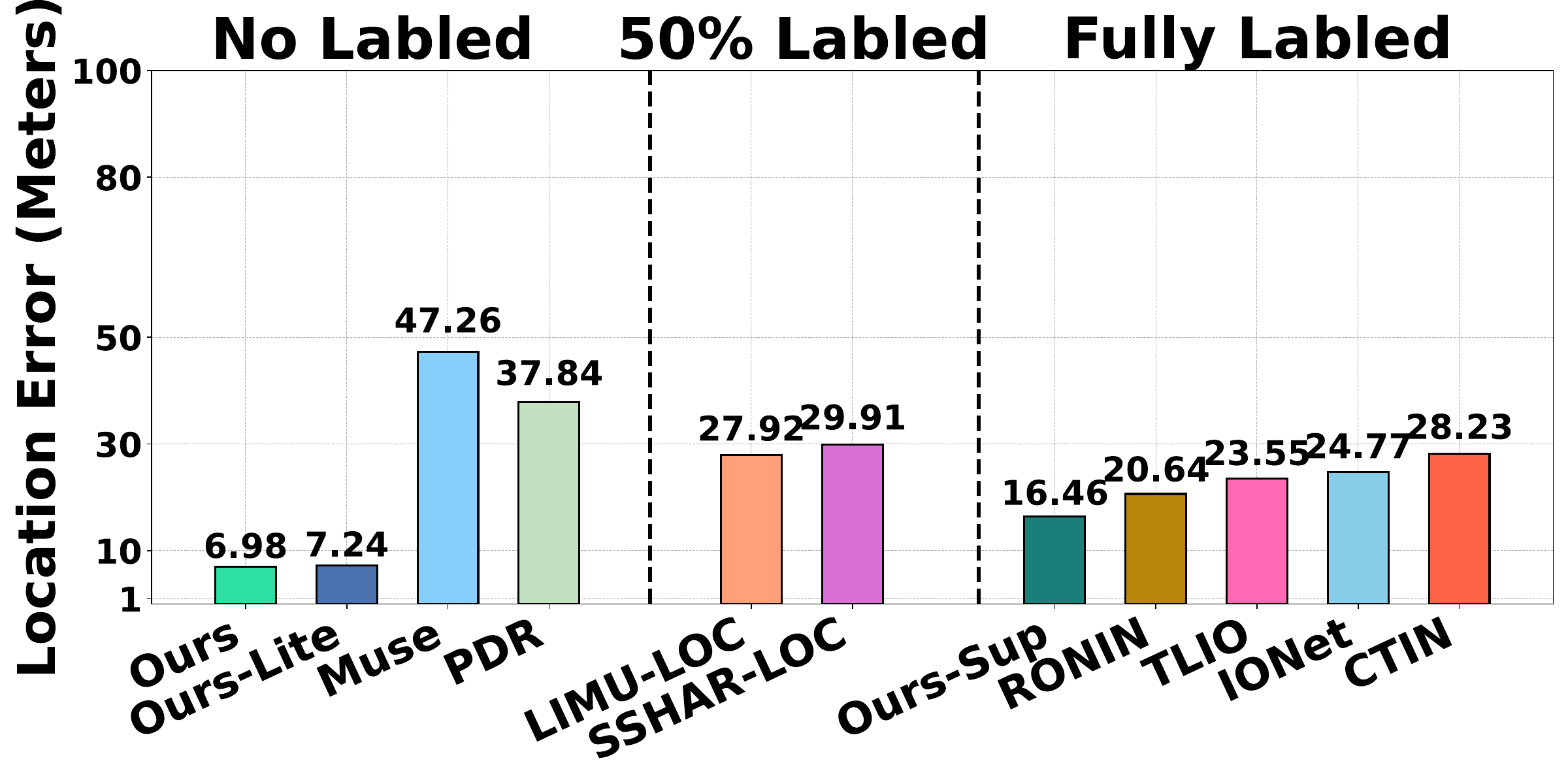}
        \vspace{-0.6cm}
        \caption{Normal}
        \label{}
    \end{subfigure}\hfill
\begin{subfigure}{0.23\textwidth}
        \centering
        \includegraphics[width=\textwidth]{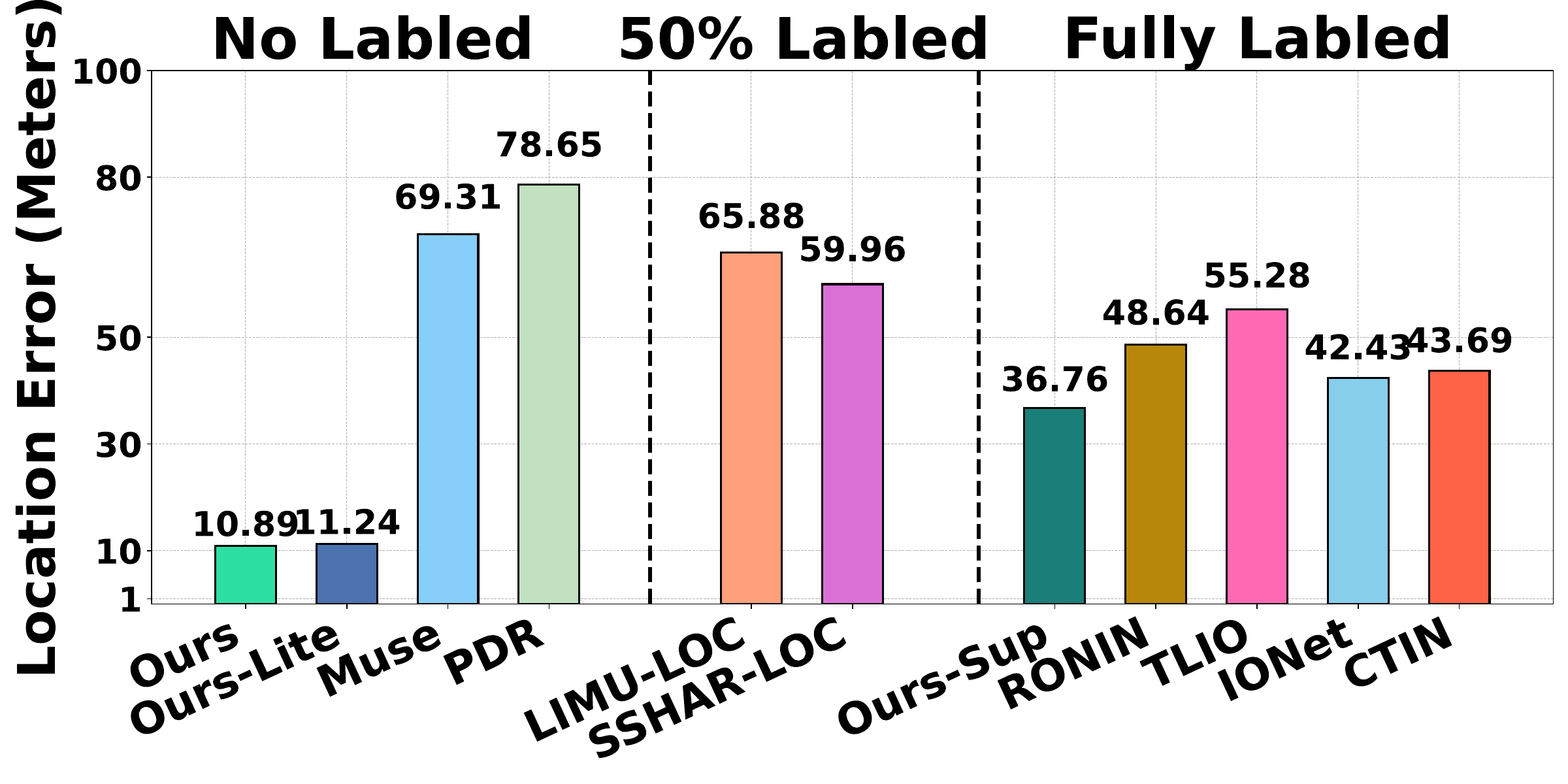}
        \vspace{-0.6cm}
        \caption{Loose}
        \label{}
    \end{subfigure}\hfill
\begin{subfigure}{0.23\textwidth}
        \centering
        \includegraphics[width=\textwidth]{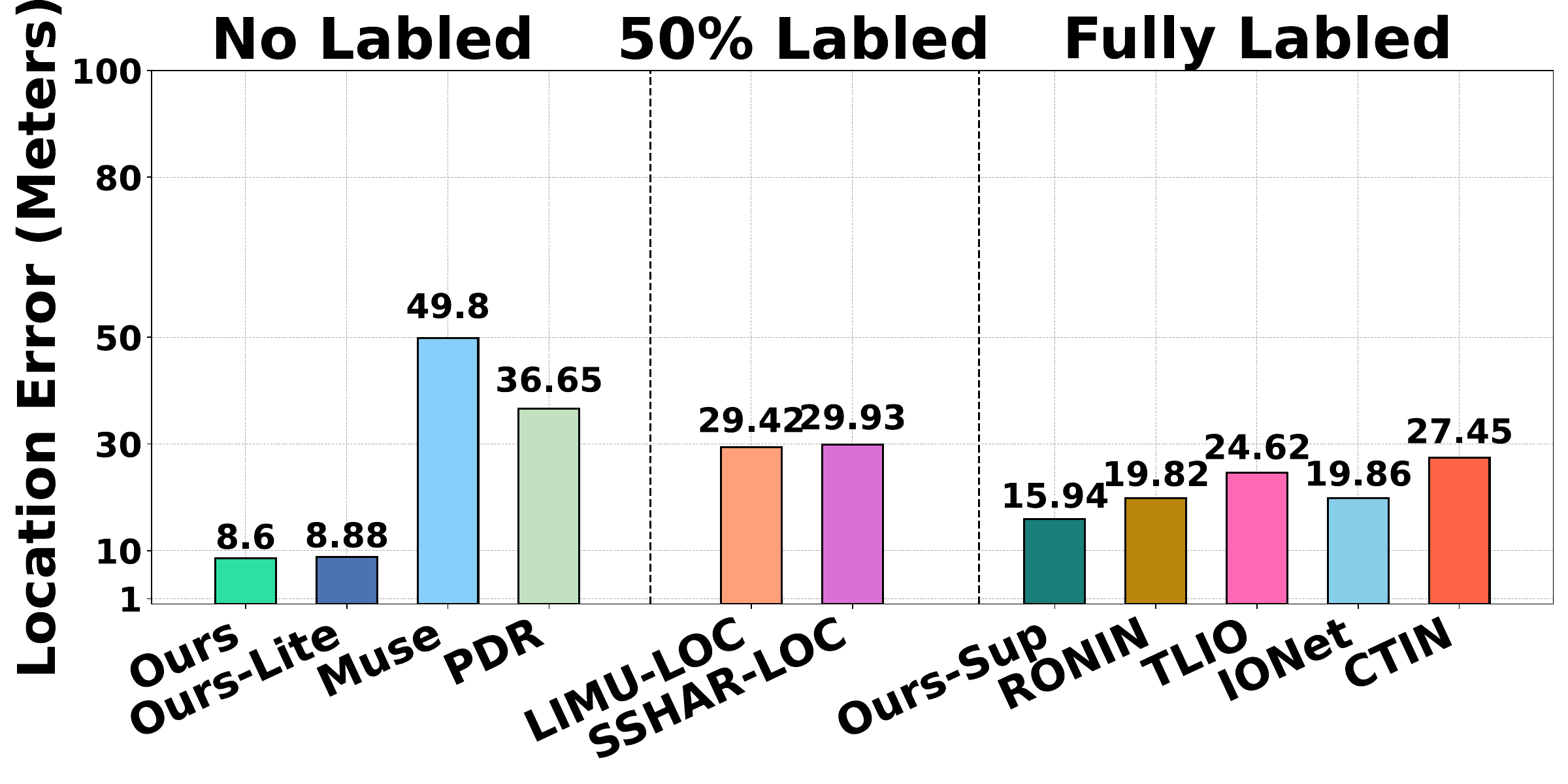}
        \vspace{-0.6cm}
        \caption{Mixed Data}
               \end{subfigure} \vspace{-0.4cm}
    \caption{Location error across different tightness conditions.}
    \vspace{-0.1cm}
\label{fig:overall_tracking}
\end{figure*}
\begin{table*}[!h]\scriptsize
\centering
\setlength{\tabcolsep}{1.65mm}
\newlength{\thickLineWidth}
\setlength{\thickLineWidth}{1.5pt}
\newcommand{\thickhline}{\noalign{\hrule height \thickLineWidth}}
\vspace{-0.1cm}
\begin{tabular}{l|c|c|c|c|c|c|c|c|c|c|c|c|c|c|c|c}
\toprule
\hline
\multirow{2}{*}{\textbf{Method}} & \multicolumn{5}{c|}{\textbf{Our Data}} & \multicolumn{5}{c|}{\textbf{SHL }} & \multicolumn{5}{c|}{\textbf{Oxiod }} & \multirow{2}{*}{\textbf{Label}} \\
\cline{2-16}
 & 2mins & 6mins & 9mins & 12mins& 15mins & 2mins & 6mins & 9mins & 12mins & 15mins & 2mins & 6mins & 9mins& 12mins & 15mins & \\

\hline\hline
Muse  & 22.42 & 49.80 &66.72 & 80.04 & 215.30 & 15.40 & 34.81 &65.40& 151.60 & 350.70 & 16.40 & 33.51 &78.81& 97.20 & 254.20 &No\\
\hline
PDR & 19.25 & 36.65 &36.91& 45.00 & 57.66 & 11.32 & 22.65&38.24 & 46.31 & 82.26 & 8.39 & 19.44 & 28.89 & 35.74 & 77.64 & No\\

\hline
\hline
\multirow{3}{*}{LIMU-LOC}  & 36.11 & 59.98 &75.84& 98.53 & 212.00 & 32.54 & 64.45 &93.42& 124.31 & 251.51 & 10.76 & 27.34& 65.29 & 139.28 & 172.17 & 10\%\\
\cline{2-17}
 & 29.15 & 43.41 &78.54& 89.32 & 162.3 & 23.51 & 34.87 &48.32& 66.94 & 97.54 & 9.44 & 26.61 &31.58& 37.66 & 61.22 & 20\%\\
\cline{2-17}
  & 18.44 & 29.42 &69.37& 88.01 & 165.9 & 6.31 & 12.50 & 18.55 & 22.60 & 46.55 & 7.54 & 19.32 & 22.74&38.70 & 62.61 & 50\%\\
\hline

\multirow{3}{*}{SSHAR-LOC}   & 39.62 & 62.47 &85.77& 144.68 & 237.54 & 35.56 & 75.67 &135.24& 182.31 & 348.89 & 13.57 & 27.82&39.21 & 58.94 & 112.18 & 10\%\\
\cline{2-17}
  & 32.24 & 58.87&76.45 & 121.55 & 189.54 & 27.57 & 33.24&49.26 & 87.53 & 189.4 & 12.53 & 27.16&39.78 & 58.54 & 98.68& 20\%\\\cline{2-17}
  & 11.21 & 29.93 & 25.26 & 35.04 & 49.24 & 6.56 & 10.98 &16.52& 21.64 & 29.74 & 6.64 & 16.50 &28.35& 35.39 & 64.94 & 50\%\\
\hline
\hline

RONIN  &13.64 & 19.82 &24.94 & 39.40 & 51.22 & 5.12 & 8.05&11.38 & 14.95 & 25.11 & 4.15 & 7.58 & 12.14 &19.84& 29.96 & Full\\
\hline

TLIO & 17.91 & 24.62 &33.45& 65.54 & 91.95 & 6.00 & 7.48 & 9.46&14.34 & 19.56 & 6.01 & 8.25 & 12.44&16.25 & 27.61 & Full \\
\hline

IONet  & 11.45 & 19.86 & 12.89 &14.88 & 22.34 & 6.22 & 9.64 &12.15& 15.65 & 28.81 & 5.32 & 13.32 &21.46& 33.64 & 39.85 &Full\\
\hline

CTIN  & 12.11 & 27.45 &32.54& 41.49 & 76.84 & 5.97 & 8.51 &17.64& 28.53 & 33.07 & 8.32 &10.45& 23.5 & 32.0 & 45.3 & Full\\

\hline
Ours-Sup  & 11.37 &15.94& 33.24 & 38.36 & 49.19  &4.98 & 9.34 & 18.61 & 27.45 &42.30& 5.45 & 8.91 & 16.49 & 27.51 &44.72& Full\\

\hline\hline
Ours-Lite  & 5.12 &8.88& 10.31 & 17.86 & 21.65  &4.69 & 6.62 & 10.23 & 13.44 &15.51& 4.08 & 5.29 & 7.06 & 9.68 &12.23& No\\
\hline

{\textbf{Ours}} & \cellcolor{red!20} 4.45 & \cellcolor{red!20}8.60 & \cellcolor{red!20}9.31&\cellcolor{red!20}14.02 & \cellcolor{red!20}18.64 & \cellcolor{red!20}4.01 & \cellcolor{red!20}6.35 &\cellcolor{red!20}8.52& \cellcolor{red!20}9.83 & \cellcolor{red!20}12.55 & \cellcolor{red!20}3.96 & \cellcolor{red!20}5.05 &\cellcolor{red!20}6.32& \cellcolor{red!20}7.92 & \cellcolor{red!20}9.60 & {No} \\\cline{2-14}
\hline
\bottomrule
\end{tabular}
\caption{
Inertial tracking results on three datasets}
\vspace{-0.8cm}
\label{tab:merged_trajectory_errors}
\end{table*}
\subsubsection{Motion Capture Performance}
We evaluate our framework on both our self-collected dataset and several public MoCap benchmarks. On our dataset, Figure~\ref{fig:overall_mocap} shows that the performance gap is small under tight attachment, but becomes much larger under normal and loose conditions. While our supervised variant is slightly better in the tight setting, supervised methods deteriorate sharply as attachment becomes less controlled, whereas our full framework remains stable. This indicates that the main advantage of our method comes from stronger robustness to sensor motion. On public datasets (Figure~\ref{fig:overall_mocap2}), sensors are worn more consistently and the performance gap narrows but remains clear, with an even larger gap on Nymeria, where the daily-activity setting is less controlled. Even controlled environments cannot fully eliminate sensor motion, and our label-free approach still outperforms both supervised and self-supervised baselines. Table~\ref{tab:ovarall_mocap} summarizes results across three benchmarks, where our method achieves the best performance on all metrics, following the same overall trend. Our lightweight variant also shows similar performance, supporting efficient deployment on embedded devices.

Figure~\ref{fig:combine_sensor_distribution} further shows angular error distributions across tight, normal, and loose conditions. When errors exceed $25^\circ$, motion becomes visually unstable; under loose attachment, SOTA baselines stay below this threshold in fewer than 5\% of cases, whereas our framework keeps over 90\% of poses within $25^\circ$ across all regimes. As looseness increases, baselines develop heavy tails, while ours  and our lite variant remain compact, indicating much stronger robustness.

We also evaluate sparse-sensor configurations—common in practice when only a phone, smartwatch, or earbud is available. Such minimal setups are ill-posed and highly sensitive to sensor relative motion. As shown in Figure~\ref{fig:popular_combine}, the performance gap between our method and the baselines becomes even larger than in the six-sensor setting: our full model and its Lite variant remain the most robust, while the supervised variant no longer does, further confirming the effectiveness of our label-free formulation under sparse and unstable sensing. This robustness comes from explicitly modeling sensor–object relative motion, which limits degradation, and from multi-view propagation, which recovers limb trajectories without direct sensor coverage, providing reliable supervision under sparse and noisy placements.

\subsubsection{Inertial Tracking Performance}

\begin{figure}
\centering
    \begin{subfigure}{0.23\textwidth}
        \centering
        \includegraphics[width=\textwidth]{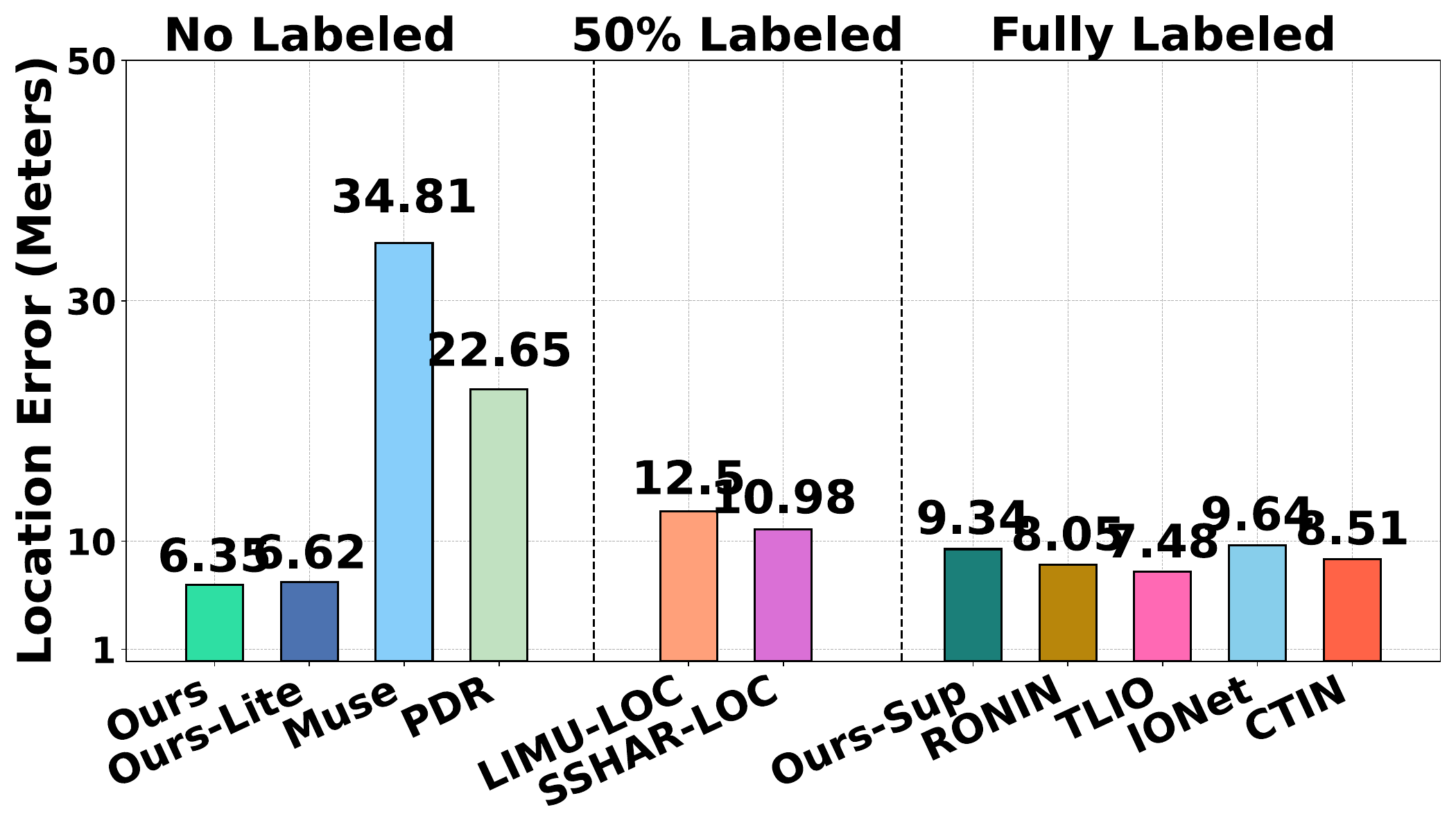} \vspace{-0.6cm}
              \caption{SHL}

        \label{}
    \end{subfigure}
    \begin{subfigure}{0.23\textwidth}
        \centering
        \includegraphics[width=\textwidth]{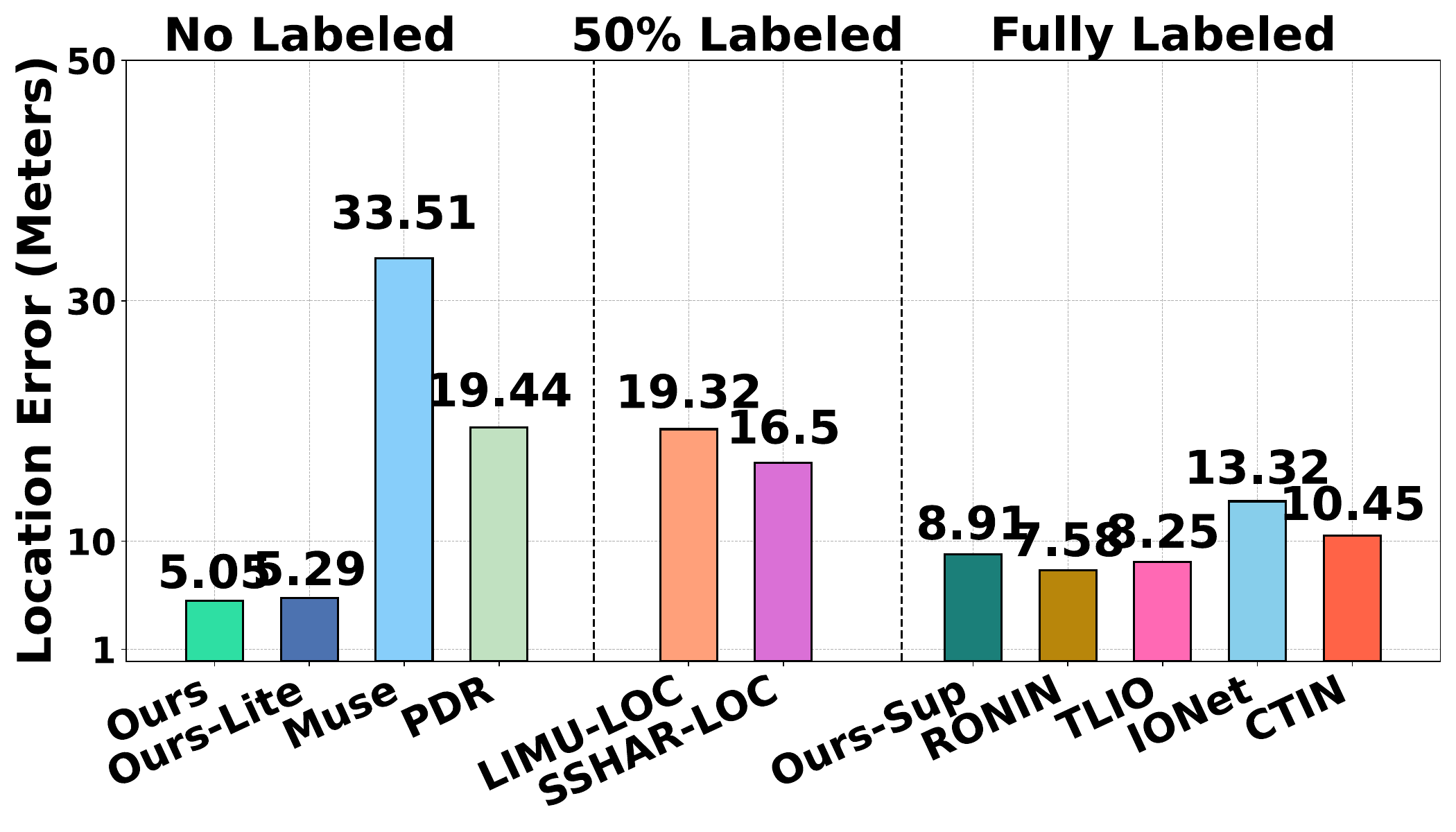}
        \vspace{-0.6cm}
        \caption{Oxiod}
        \label{}
    \end{subfigure}\hfill
    \vspace{-0.4cm}
    \caption{Inertial Tracking comparison on public datasets with 6 minutes trajectory slice.}
        \label{fig:crossdataset}
\end{figure}
\begin{figure}
\vspace{-0.5cm}
        \centering
        \hspace{-0.3cm}\includegraphics[width=0.5\textwidth]{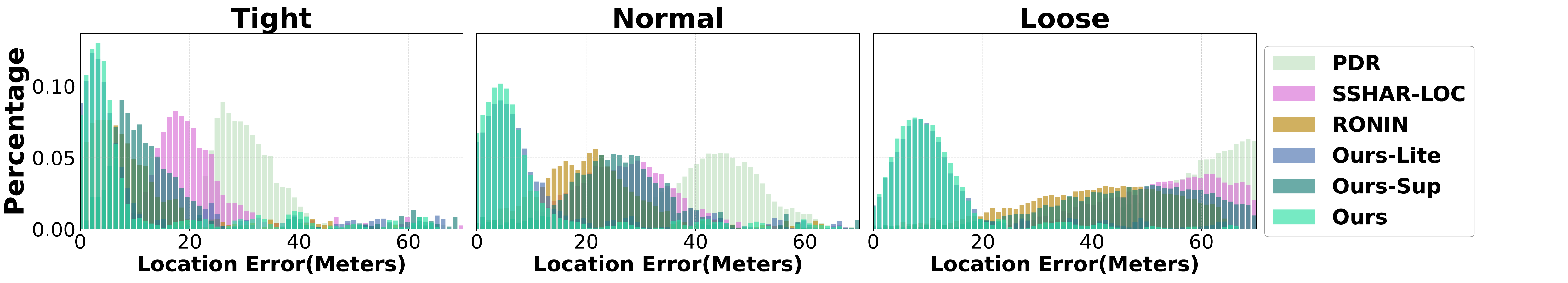}
                \vspace{-0.5cm}
    \caption{Location error distribution with three tightness conditions.}
    \vspace{-0.5cm}
        \label{fig:Distrubute_navigation}
\end{figure}

\begin{figure*}
    \centering
    \hspace{-0.2cm}
    \begin{subfigure}{0.45\textwidth}
        \centering
        \includegraphics[width=\textwidth]{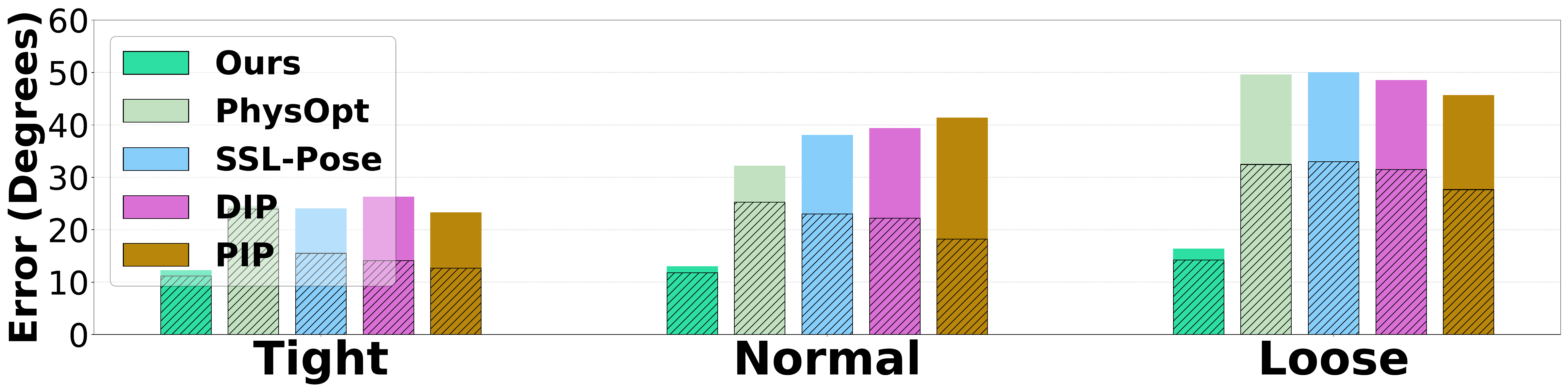}
                \vspace{-0.2cm}
        \caption{Motion Capture}
        \label{}
    \end{subfigure}
    \hfill
    \hspace{+0.2cm}
    \begin{subfigure}{0.45\textwidth}
        \centering
        \includegraphics[width=\textwidth]{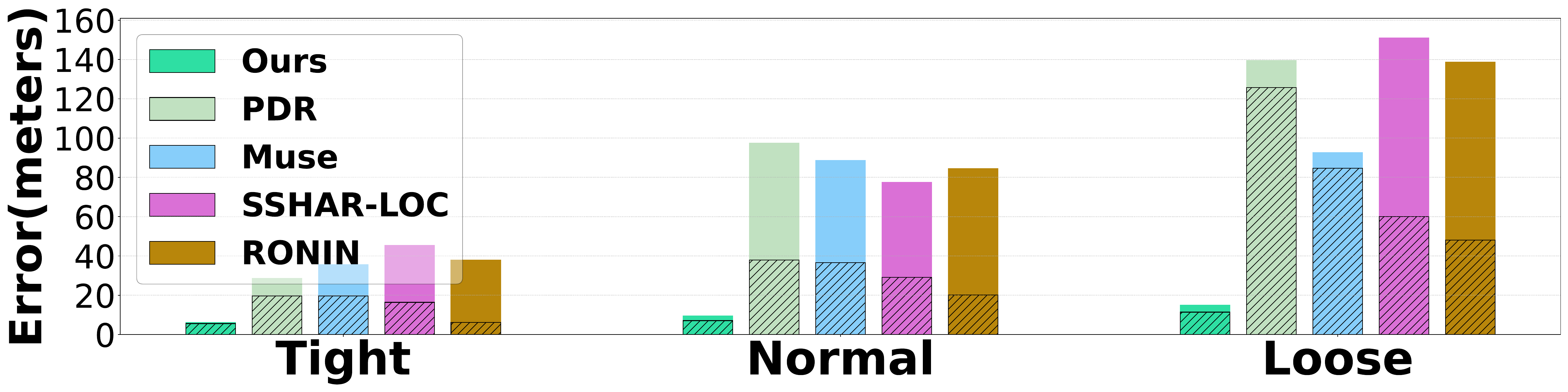} 
                \vspace{-0.2cm}
        \caption{Inertial Tracking}
        \label{}
    \end{subfigure}\hfill
    \vspace{-0.45cm}
    \caption{
    Leave-one-condition-out evaluation by wearing tightness. Non-shaded bars show out-of-domain performance when a tightness condition is unseen during training; shaded bars show in-domain performance after fine-tuning with data from that condition.
    }
\label{fig:Leave-one-tightness}

\end{figure*}

\begin{figure*}[]
\vspace{-0.1cm}
    \centering
    \begin{subfigure}{0.24\textwidth}
        \centering
        \includegraphics[width=\textwidth]{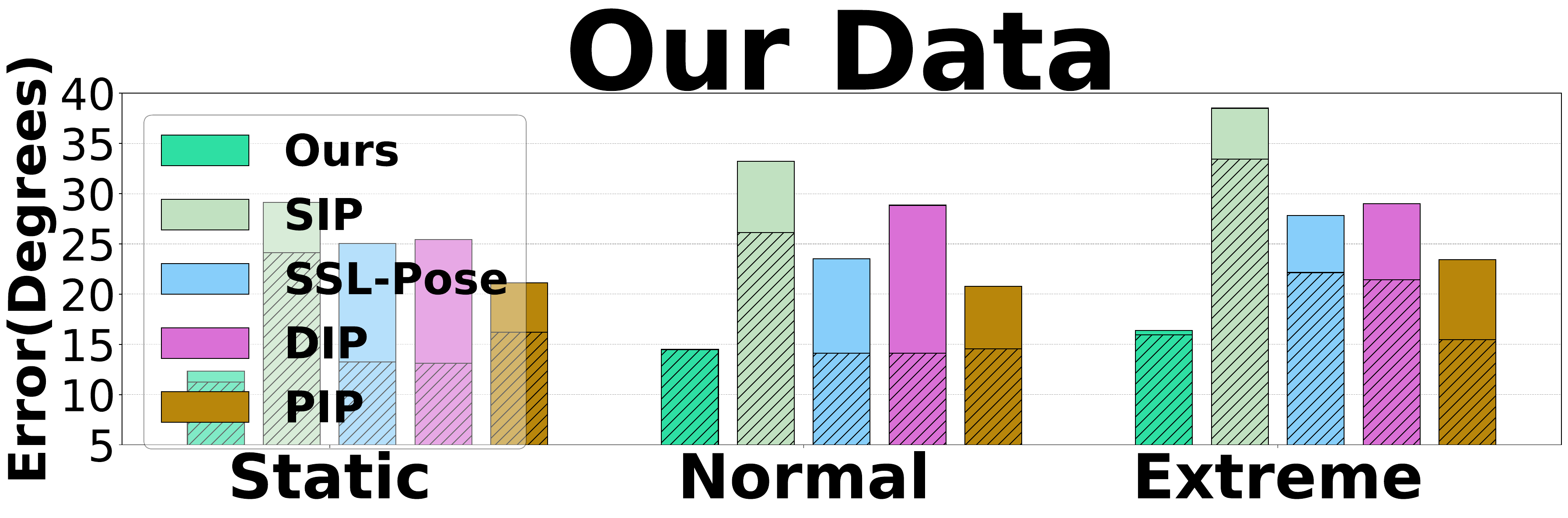}
    \end{subfigure}\hfill
    \begin{subfigure}{0.24\textwidth}
        \centering
        \includegraphics[width=\textwidth]{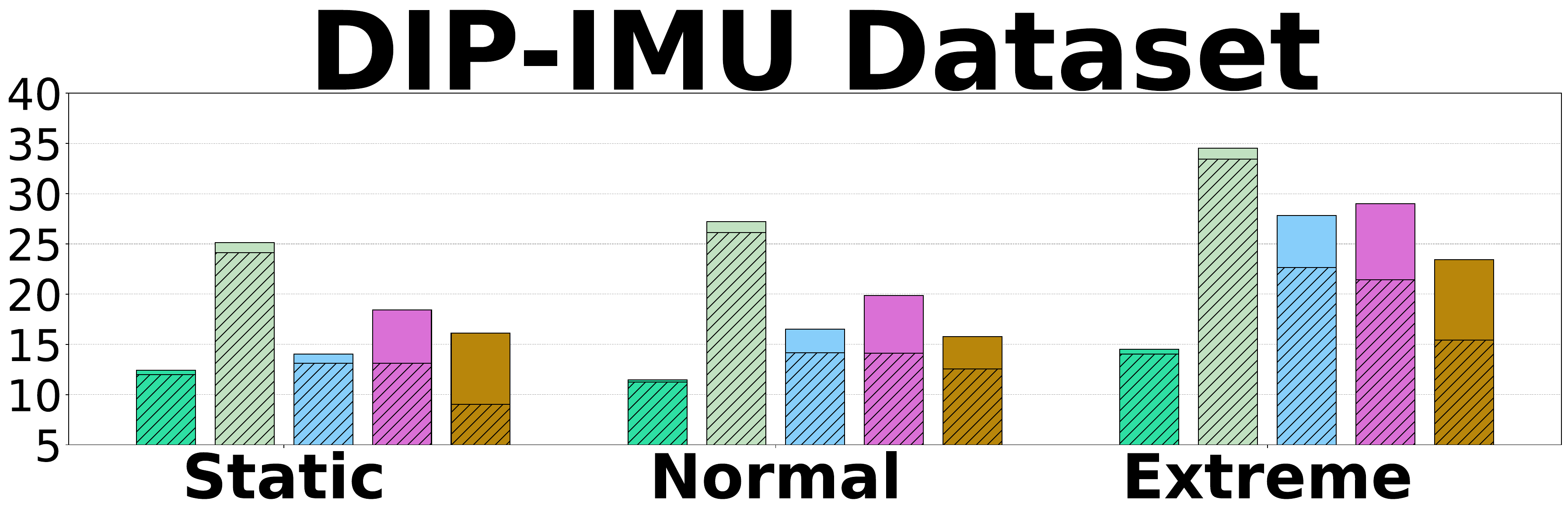}
    \end{subfigure}\hfill
    \begin{subfigure}{0.24\textwidth}
        \centering
        \includegraphics[width=\textwidth]{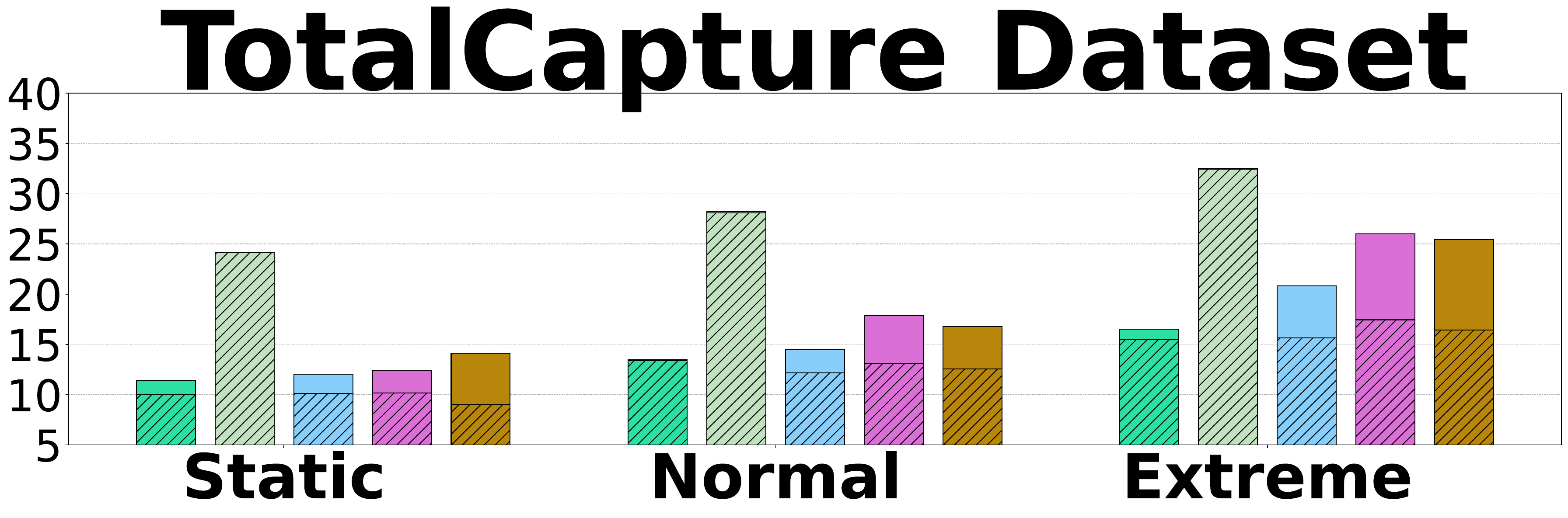}
    \end{subfigure}\hfill
    \begin{subfigure}{0.24\textwidth}
        \centering
        \includegraphics[width=\textwidth]{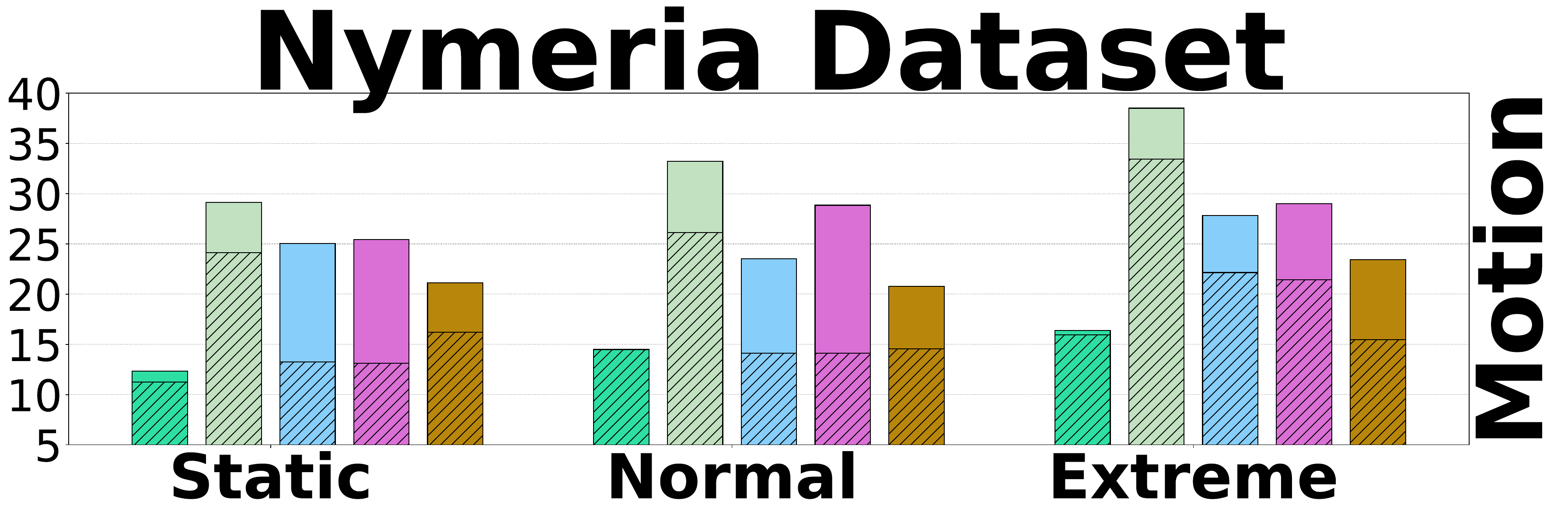}
    \end{subfigure}\hfill

       \begin{subfigure}{0.24\textwidth}
        \centering
        \includegraphics[width=\textwidth]{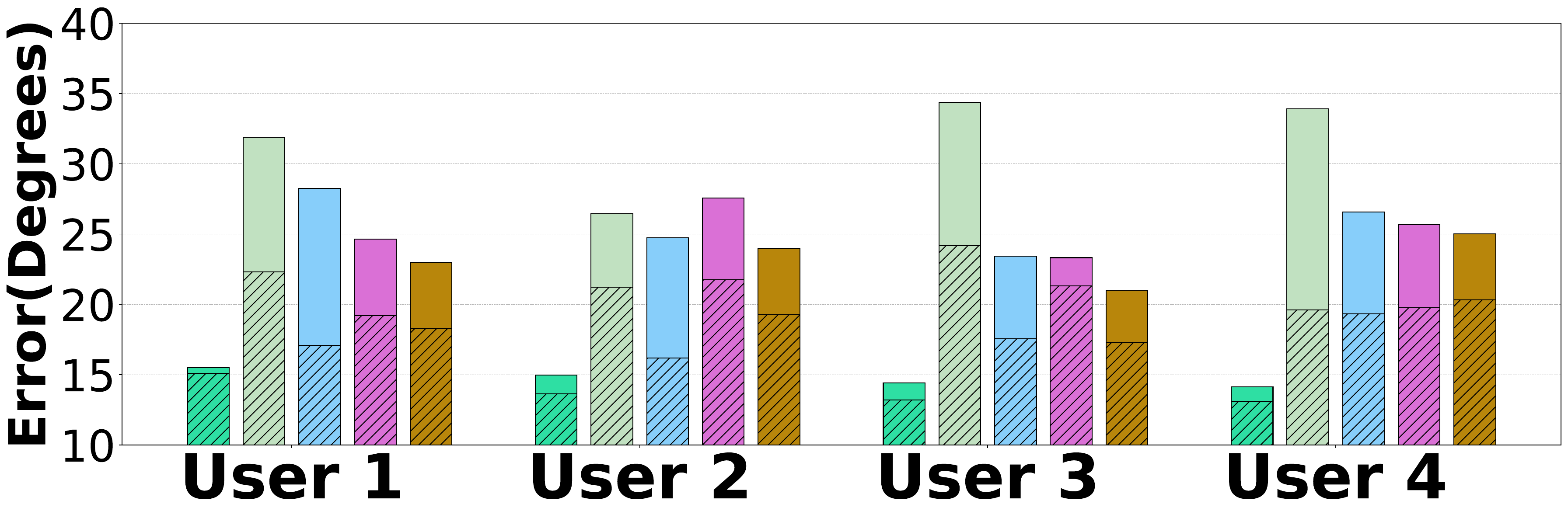}
    \end{subfigure}\hfill
    \begin{subfigure}{0.24\textwidth}
        \centering
        \includegraphics[width=\textwidth]{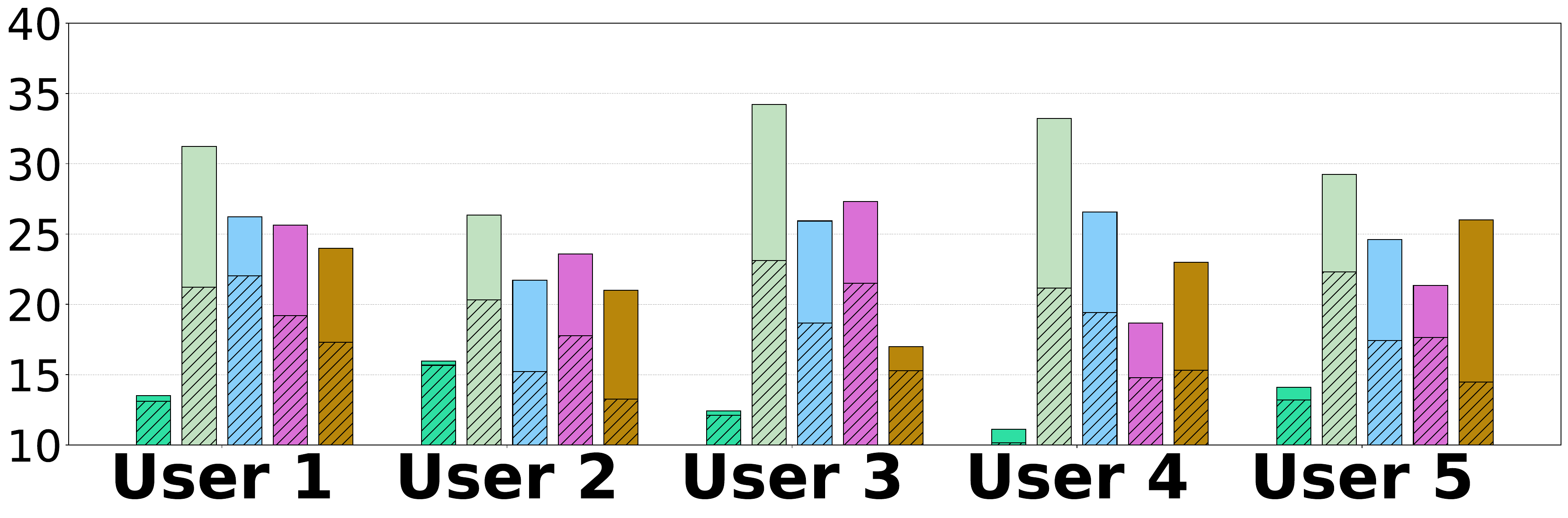}
    \end{subfigure}\hfill
    \begin{subfigure}{0.24\textwidth}
        \centering
        \includegraphics[width=\textwidth]{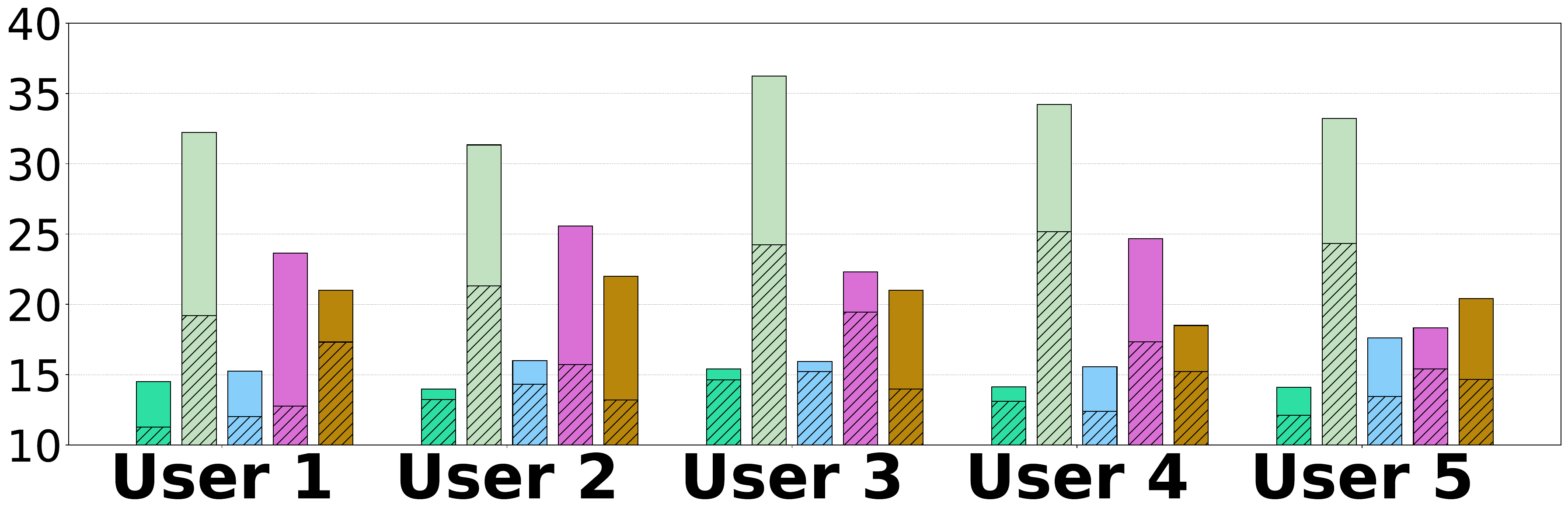}
    \end{subfigure}\hfill
        \begin{subfigure}{0.24\textwidth}
        \centering
        \includegraphics[width=\textwidth]{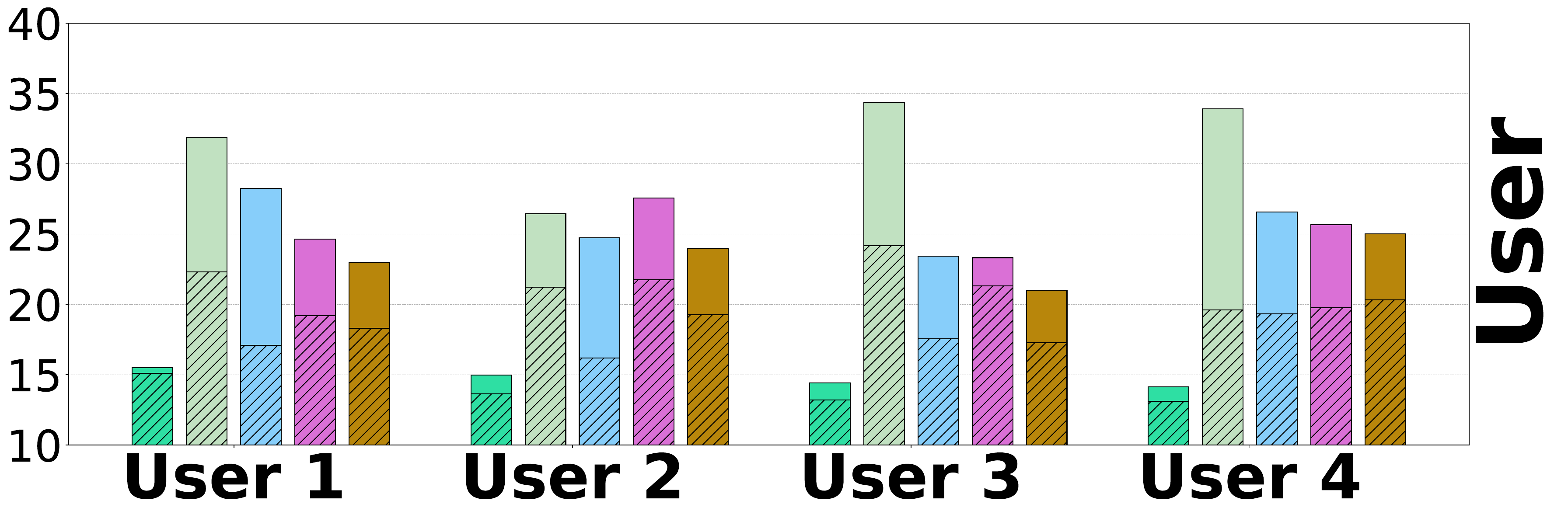}
    \end{subfigure}
    \vspace{-0.2cm}
    \caption{ Motion capture results under additional leave-one-scenario-out evaluations, covering both motion and user scenarios. Non-shaded bars show out-of-domain performance on the held-out scenario; shaded bars show in-domain performance after fine-tuning on that scenario.}
\label{fig:generalization_mocap}

\end{figure*}
\begin{figure*}[ht]
    \begin{subfigure}{0.33\textwidth}
        \centering
        \includegraphics[width=\textwidth]{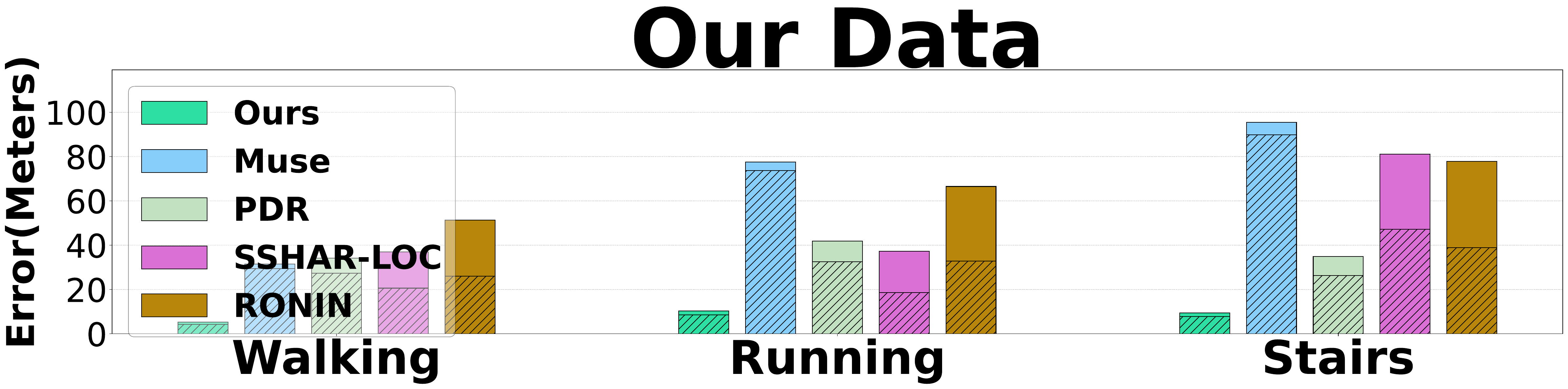}
        \label{fig:placement_generalization2}
    \end{subfigure}\hfill
    \begin{subfigure}{0.33\textwidth}
        \centering
        \includegraphics[width=\textwidth]{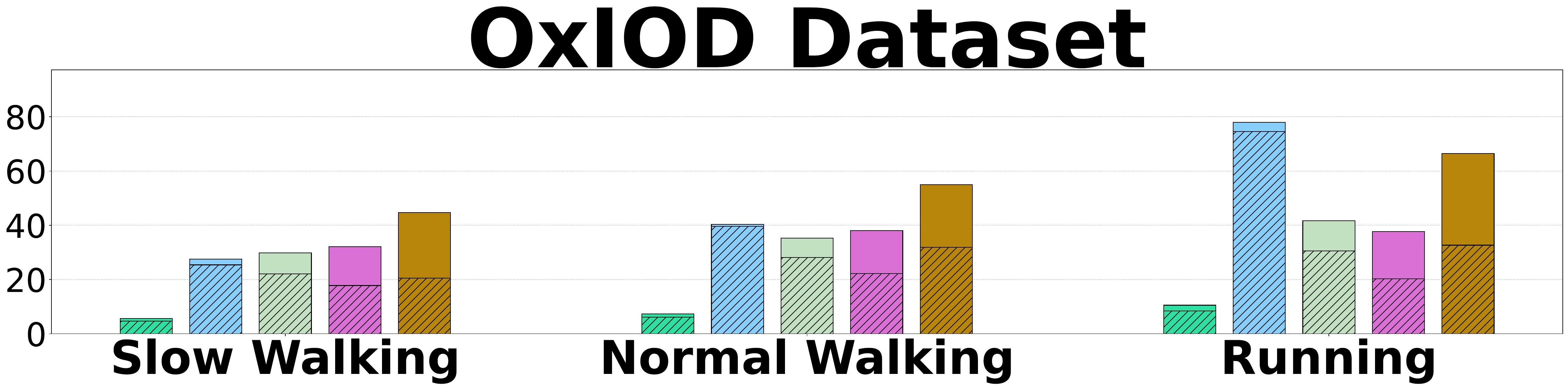}
        \label{fig:motion_generalization2}
    \end{subfigure}\hfill
    \vspace{-0.1cm}
    \begin{subfigure}{0.33\textwidth}
        \centering
        \includegraphics[width=\textwidth]{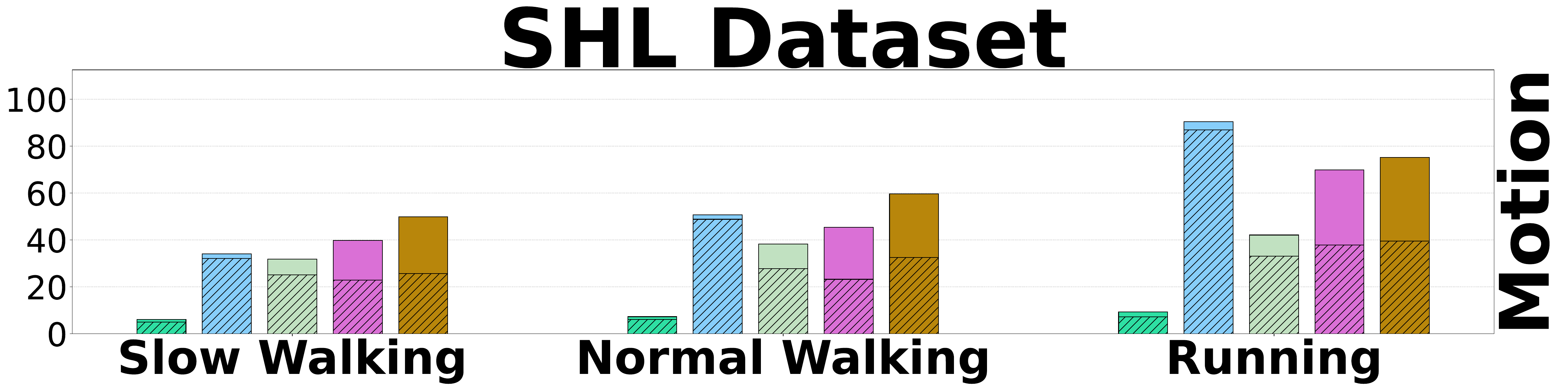}
        \label{fig:user_generalization2}
    \end{subfigure}

    \begin{subfigure}{0.33\textwidth}
        \centering
        \includegraphics[width=\textwidth]{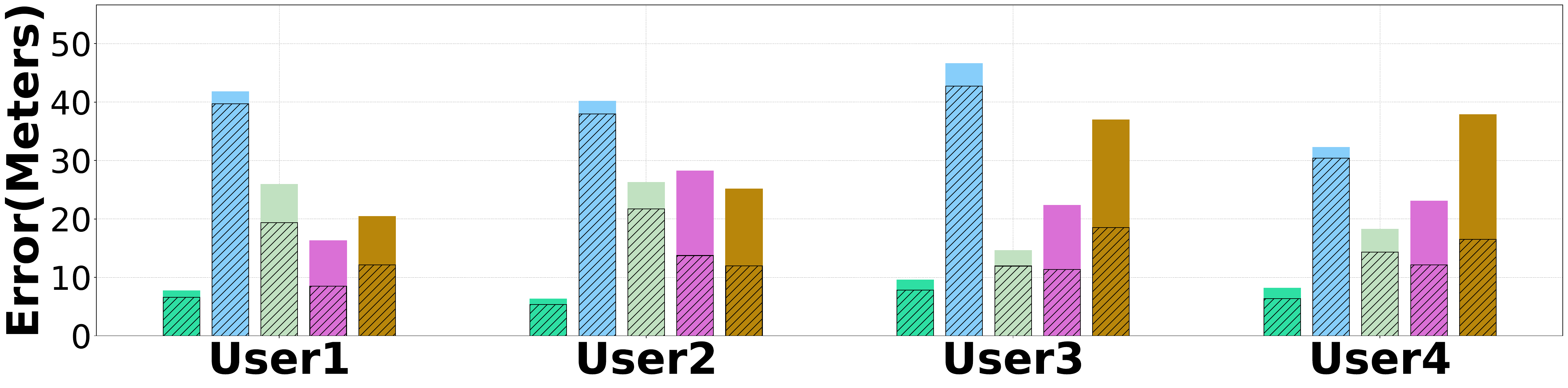}
        \label{fig:placement_generalization}
    \end{subfigure}\hfill
    \begin{subfigure}{0.33\textwidth}
        \centering
        \includegraphics[width=\textwidth]{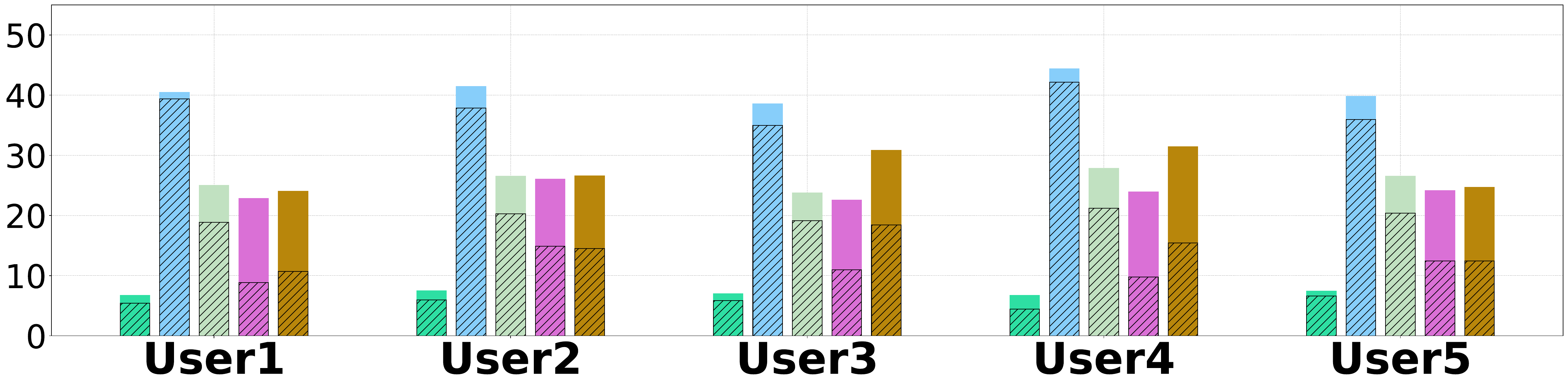}
        \label{fig:motion_generalization}
    \end{subfigure}\hfill
    \begin{subfigure}{0.33\textwidth}
        \centering
        \includegraphics[width=\textwidth]{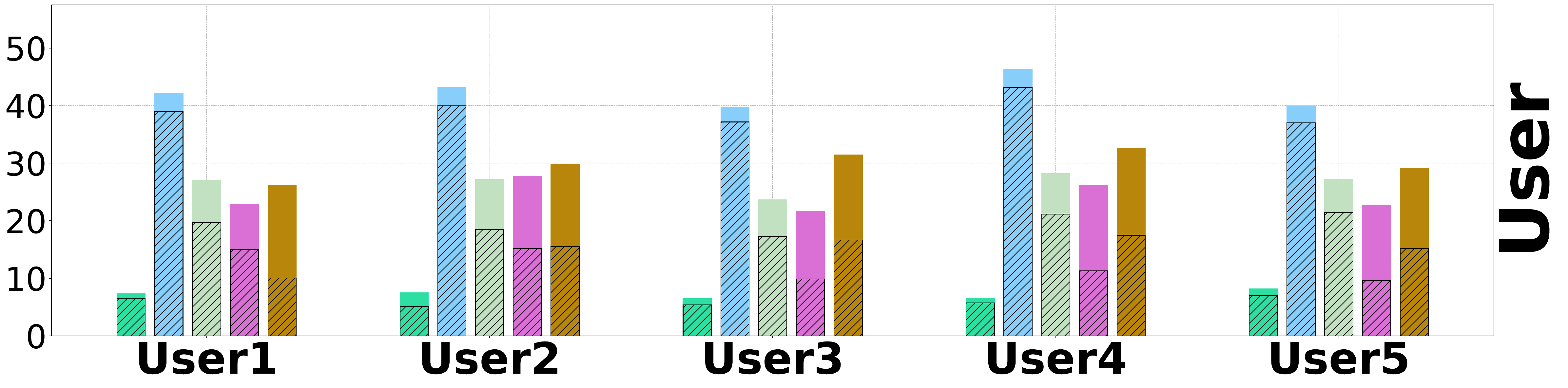}
        \label{fig:user_generalization}
    \end{subfigure}

  \vspace{-0.3cm}
    \caption{Inertial tracking results under additional leave-one-scenario-out evaluations, covering both motion and user scenarios. Non-shaded bars show out-of-domain performance on the held-out scenario; shaded bars show in-domain performance after fine-tuning on that scenario.}
\vspace{-0.2cm}
  \label{fig:dataset_gene_track}
\end{figure*}
\begin{figure}[!h]
    \centering
    \begin{subfigure}{0.22\textwidth}
       \includegraphics[width=\textwidth]{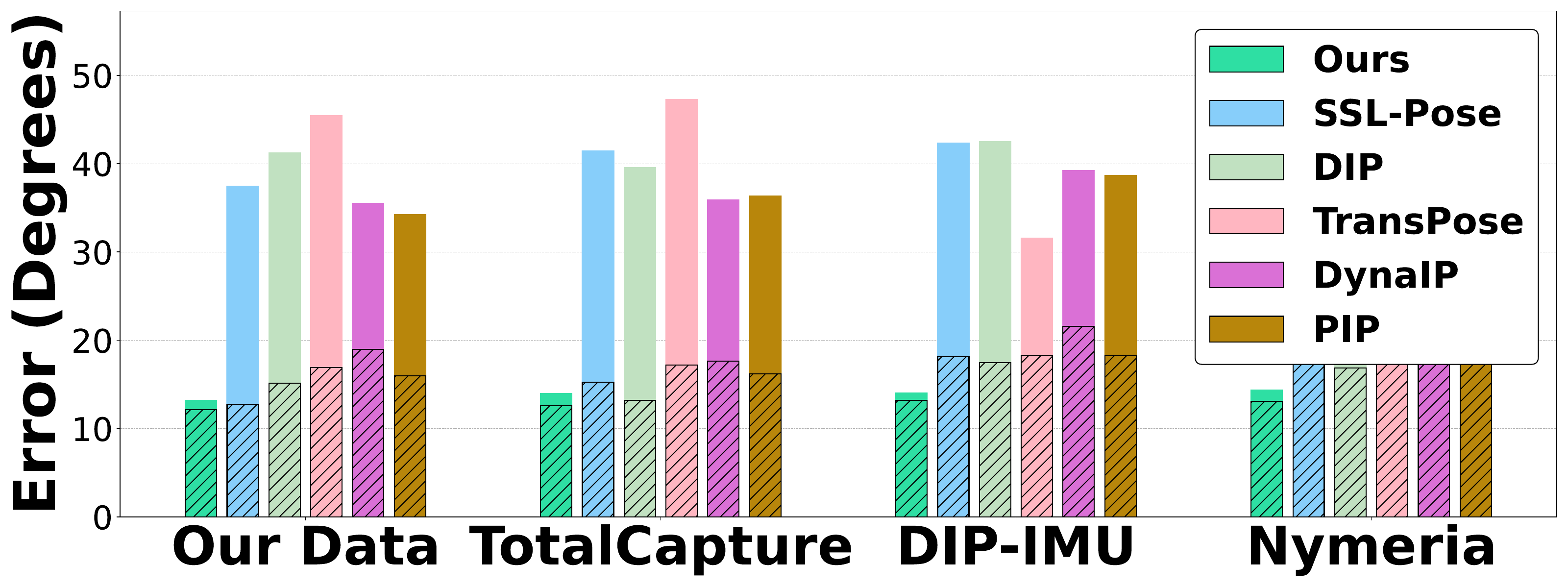}
        \label{}
    \end{subfigure}
    \begin{subfigure}{0.22\textwidth}
        \centering
        \includegraphics[width=\textwidth]{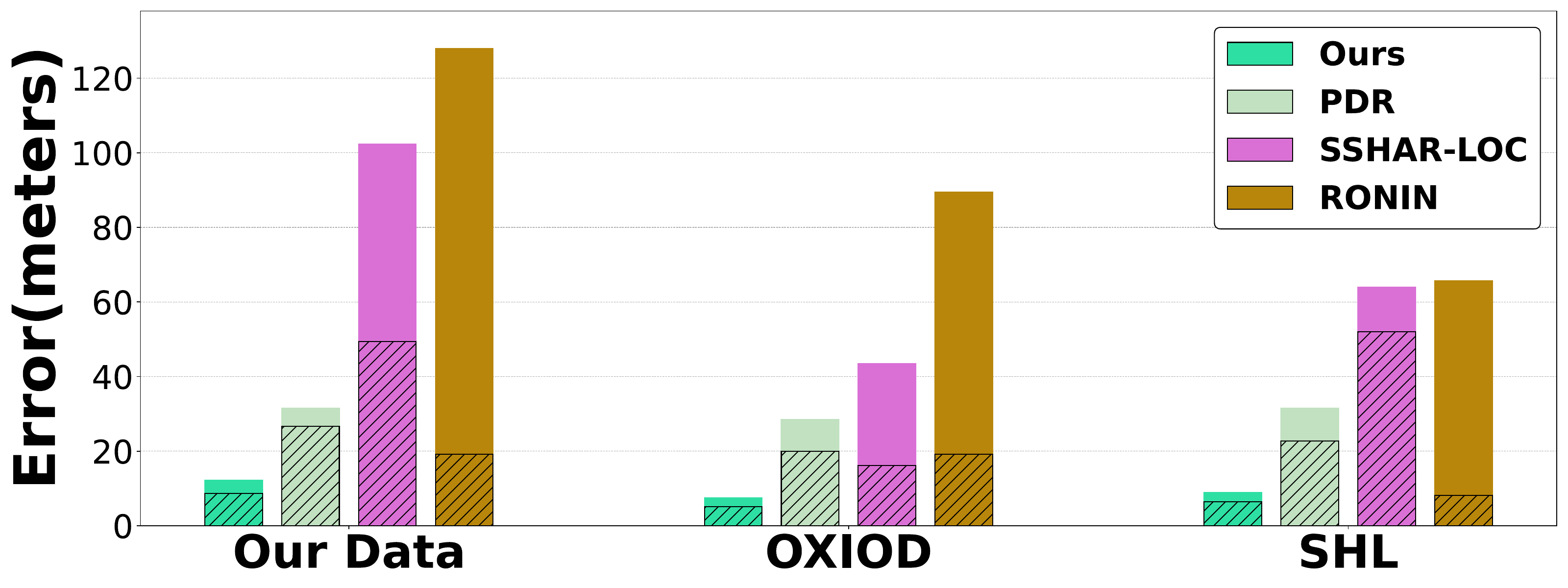}
        \label{}
    \end{subfigure}
    \vspace{-0.3cm}
    \caption{Leave-One-Dataset-Out Training to Evaluate Model Performance in Unseen Environments.}\vspace{-0.4cm}
    \label{fig:cross_dataset}
\end{figure}
We evaluate inertial tracking on our dataset and two public benchmarks (Figures~\ref{fig:maps}--\ref{fig:overall_tracking}). Our method achieves the lowest trajectory error across all tightness conditions, reducing drift by roughly 1.5--2$\times$ on six-minute trajectories, despite using no labels. Consistent with the MoCap results, the key gap again appears as attachment becomes less controlled: methods that implicitly rely on rigid attachment degrade rapidly, while our framework remains stable by explicitly predicting sensor--object motion. Classical PDR accumulates substantial drift, and self-supervised baselines improve with labels, yet even with $50\%$ supervision their errors are still nearly 2$\times$ ours. Similar trends hold on public datasets. Error distributions (Figure~\ref{fig:Distrubute_navigation}) show that fewer than 10\% of our trajectories exceed 15 m after six minutes, whereas competing methods have much heavier tails.

\begin{figure*}[t]

    \centering

    \begin{subfigure}{0.48\linewidth}
        \centering
        \includegraphics[width=\linewidth]{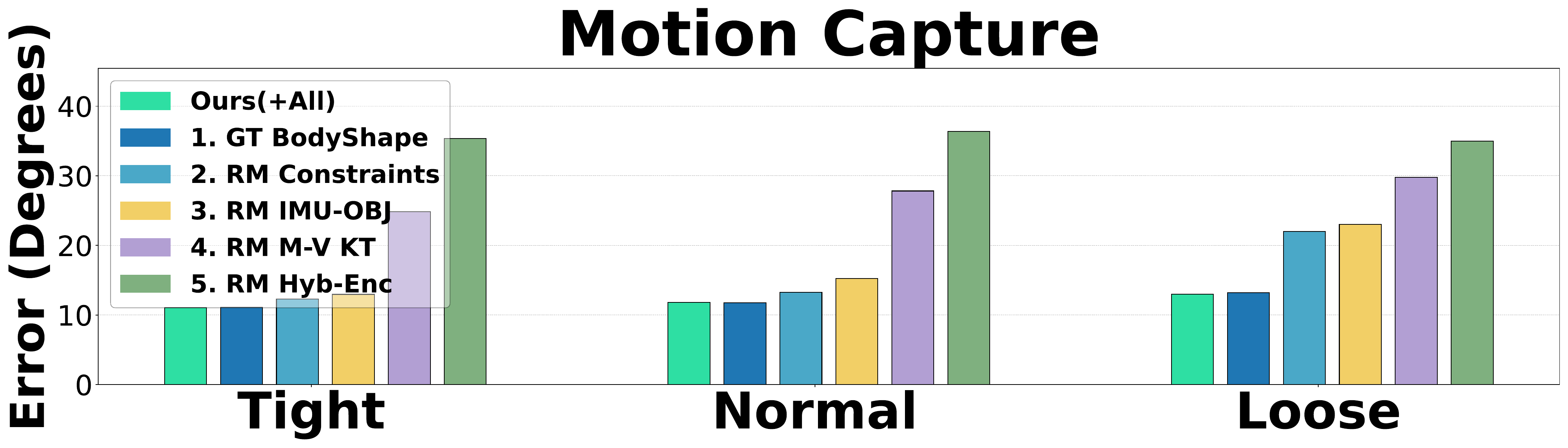}
    \end{subfigure}
    \begin{subfigure}{0.48\linewidth}
        \centering
        \includegraphics[width=\linewidth]{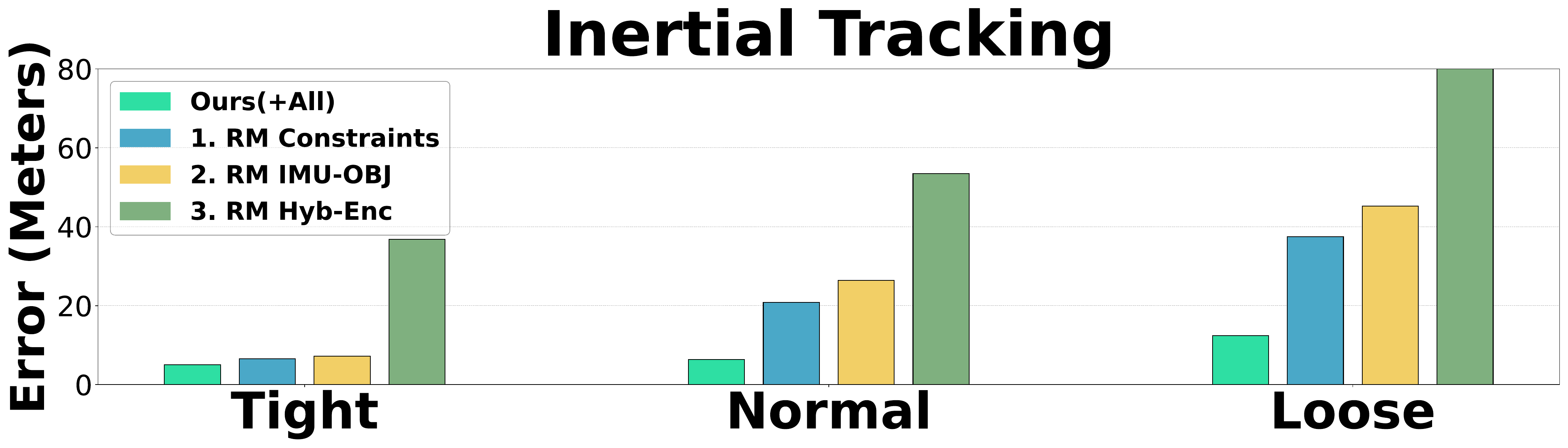}
    \end{subfigure}
\vspace{-0.3cm}
\caption{Stepwise ablation of our model under different sensor-attachment conditions.
Starting from the full model (Ours(+All)), we (1) replace the learned body-shape module with ground-truth body shape (GT BodyShape), (2) further remove frequency-spatial constraints (RM Constrain), (3) additionally remove IMU–object relative-motion prediction (RM IMU–OBJ), (4) drop the multi-view kinematic tree (RM M-V KT), and (5) finally remove the hybrid encoder (RM Hyb-Enc).}
    \label{fig:ablation}
\end{figure*}

\begin{figure*}[!h]

    \centering
    \begin{subfigure}{0.24\textwidth}
        \centering
        \includegraphics[width=\textwidth]{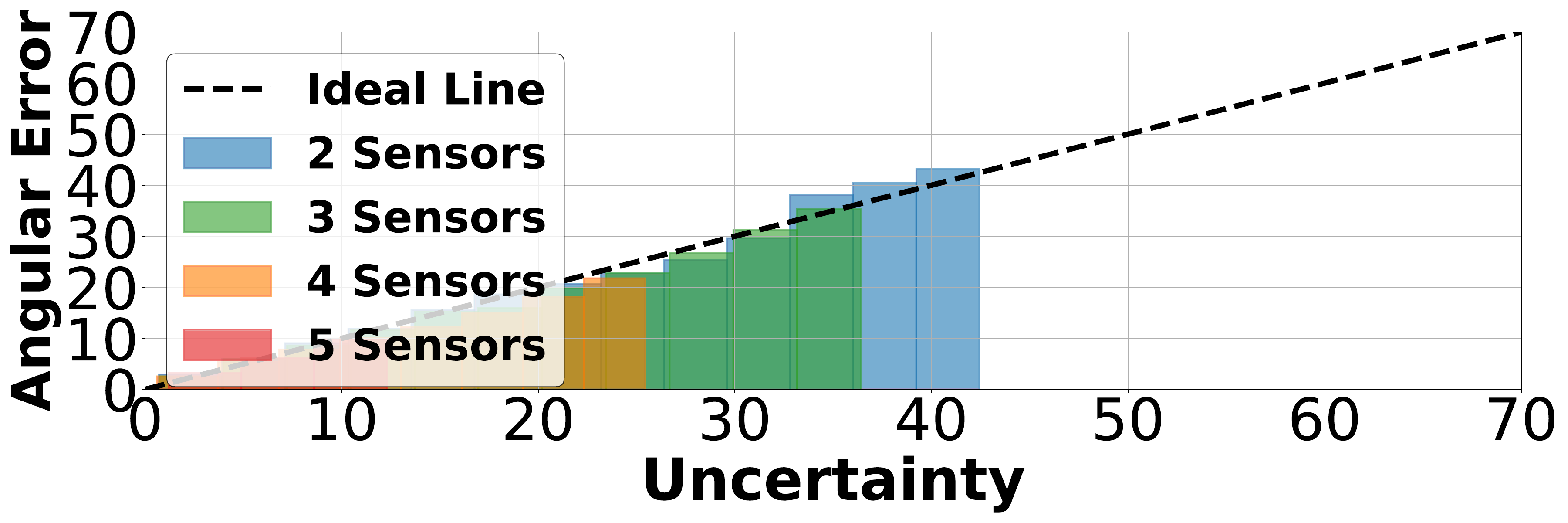} 
        \caption{\centering16 Samples}
        \label{}
    \end{subfigure}
    \begin{subfigure}{0.24\textwidth}
        \centering
        \includegraphics[width=\textwidth]{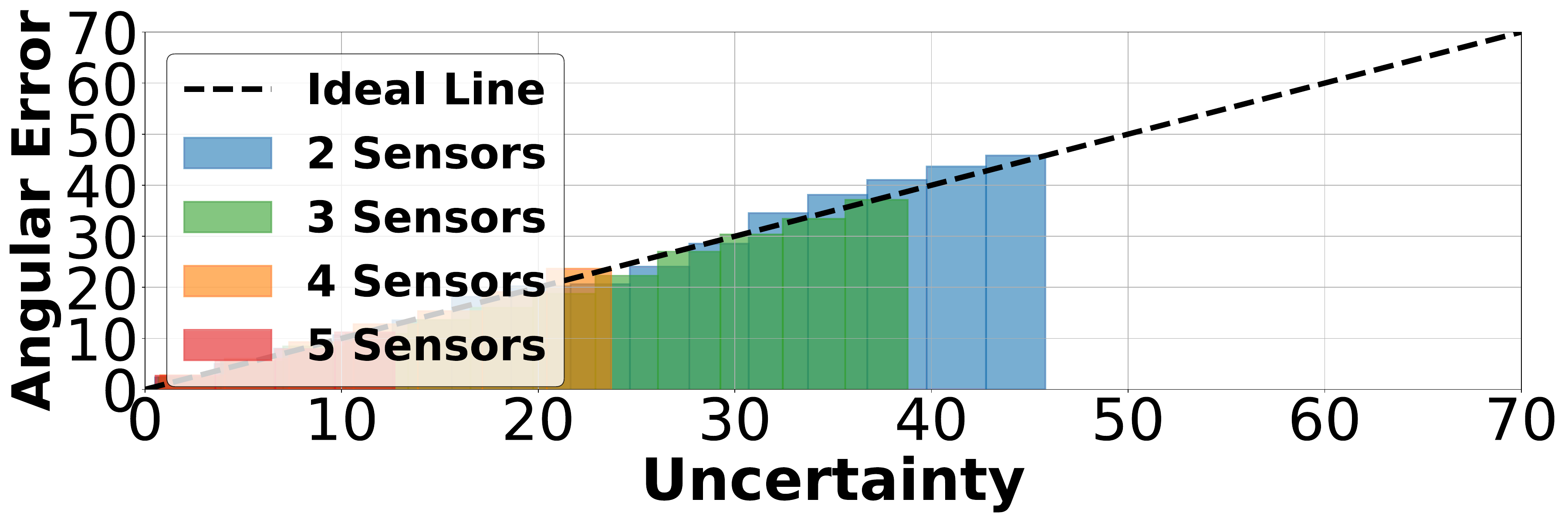} 
        \caption{\centering8 Samples}
        \label{}
    \end{subfigure}
    \begin{subfigure}{0.24\textwidth}
        \centering
        \includegraphics[width=\textwidth]{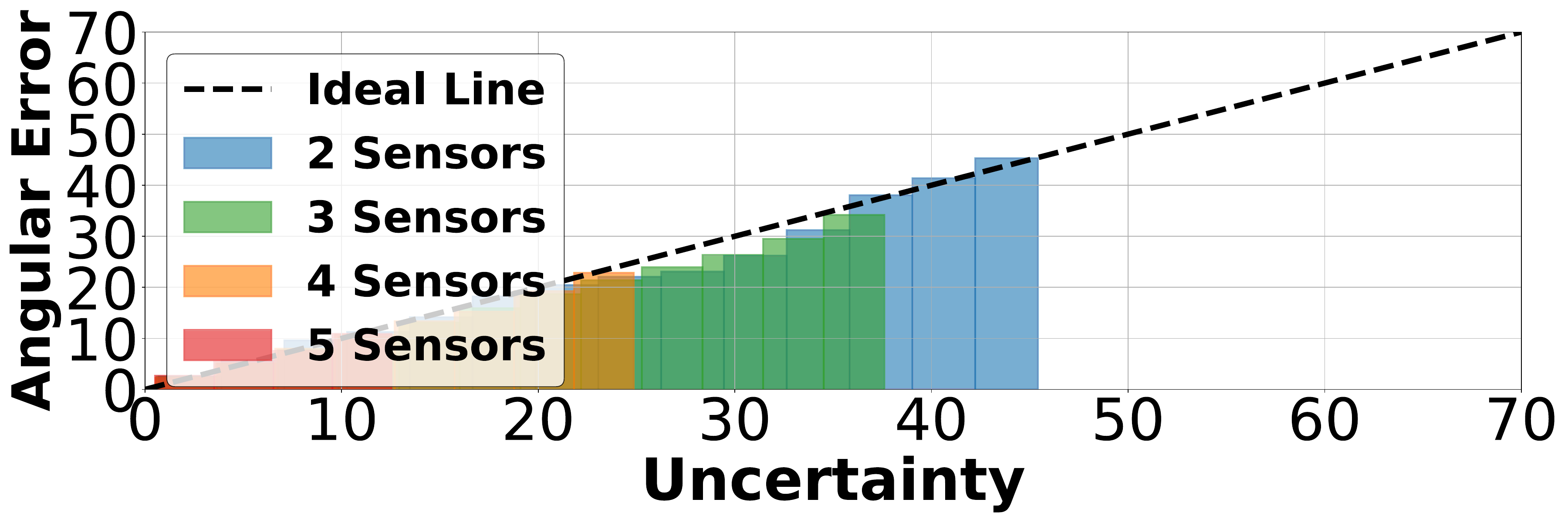}
        \caption{\centering4 Samples}
        \label{}
    \end{subfigure}
    \begin{subfigure}{0.24\textwidth}
        \centering
        \includegraphics[width=\textwidth]{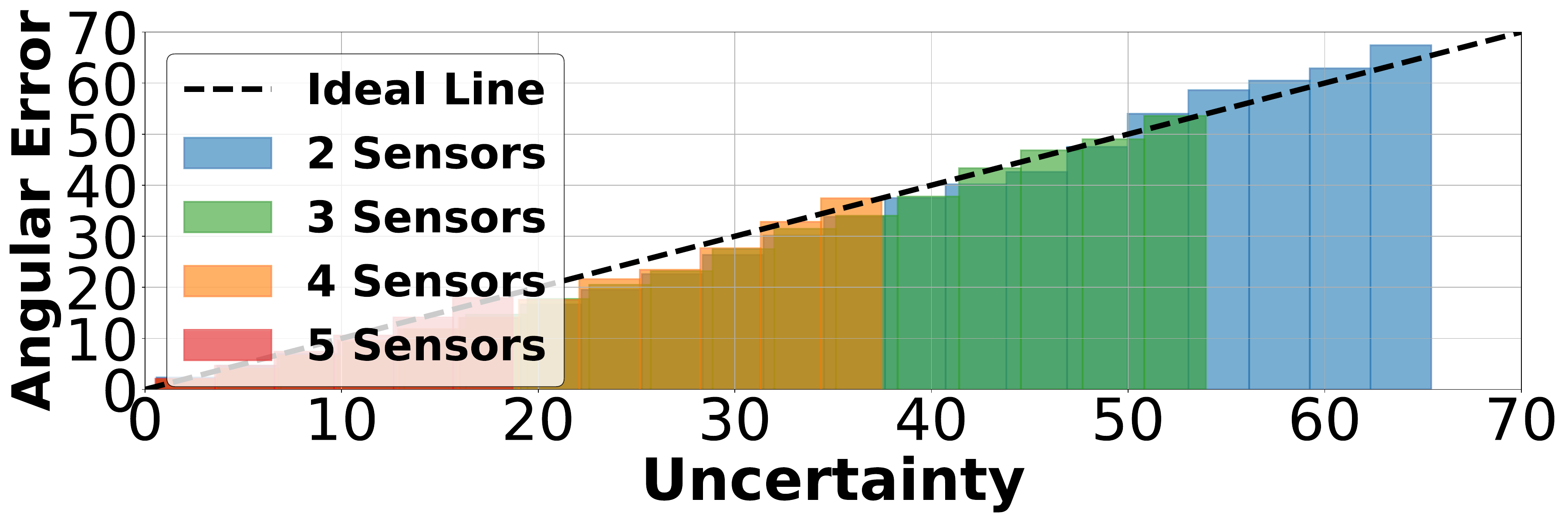}
\caption{\centering2 Samples}
        \label{}
    \end{subfigure}    \vspace{-0.2cm}\caption{
    Output uncertainty vs. predictive error for different sample counts. All cases remain close to the ideal, indicating well-calibrated uncertainty.
    }\vspace{-0.2cm}
        \label{fig:reliability_uncertainty}
\end{figure*}
\subsection{Cross-Domain Generalization Analysis}
Real-world deployments involve broad variability in wearing tightness, motion intensity, sensor placement, and user behavior. To evaluate robustness under unseen conditions, we adopt a leave-one-scenario-out protocol in which an entire condition is excluded during training and used only for testing. This provides a direct measure of generalization without additional labels or fine-tuning.

To assess robustness to attachment variability, we first hold out each wearing condition (tight, normal, loose). As shown in Figure~\ref{fig:Leave-one-tightness}, \textit{non-shaded bars show out-of-domain performance when a tightness condition is unseen during training; shaded bars show in-domain performance after fine-tuning with data from that condition.} Baseline methods degrade notably, especially under loose attachment, while our method remains stable across all three cases.

For motion capture, we further hold out (i) users and (ii) motion categories. These two settings stress different aspects of generalization. When motion type changes, the underlying physical equations remain valid, but physical-based methods are highly sensitive to noise; our noise-reduced latent space retains the generality of the physics while improving robustness to disturbances. When user identity changes, personalized physical parameters shift; our data-driven modeling infers these parameters directly from the input, enabling adaptation to new subjects. As shown in Figure~\ref{fig:generalization_mocap}, our method shows minimal degradation across both cases, whereas supervised baselines degrade substantially and still fail to predict unseen users or motions accurately even on the large-scale Nymeria dataset.
For inertial tracking, we follow analogous splits over users and motion categories. According to Figure~\ref{fig:dataset_gene_track}, baselines struggle under unseen motions or new users, while our model exhibits consistent performance across all held-out conditions, mirroring the trends observed in motion capture.

Finally, in a leave-one-dataset-out evaluation (Figure~\ref{fig:cross_dataset}), prior methods suffer large drops on unseen datasets, whereas our label-free framework maintains significantly higher accuracy, demonstrating robust cross-dataset generalization.

\begin{figure*}[htbp]
\vspace{-0.2cm}
\begin{subfigure}{0.3\textwidth}
        \centering
        \includegraphics[width=\textwidth]{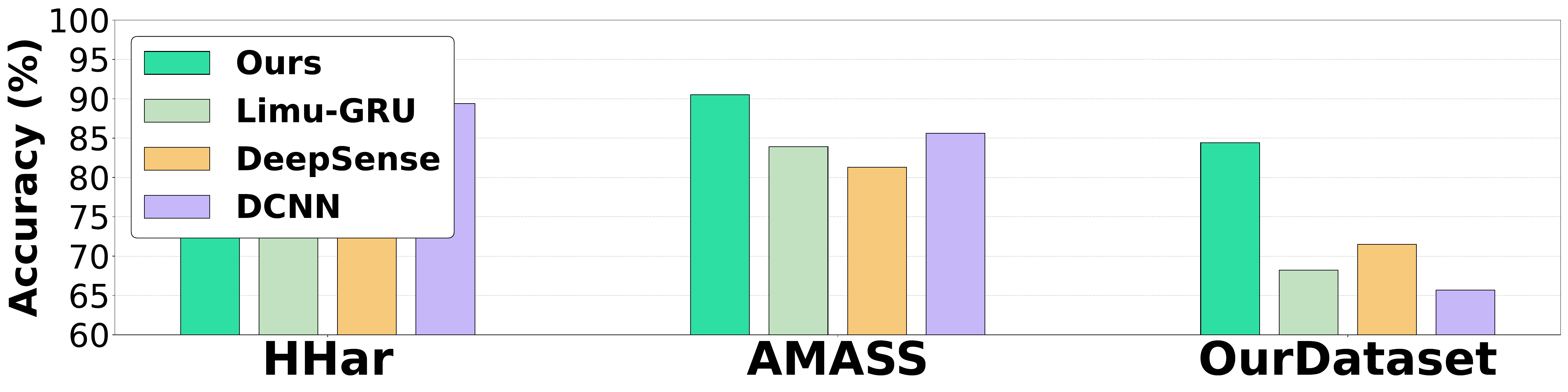}
        \caption{HAR Task}

    \end{subfigure}\hfill
    \begin{subfigure}{0.28\textwidth}
        \centering
        \includegraphics[width=\textwidth]{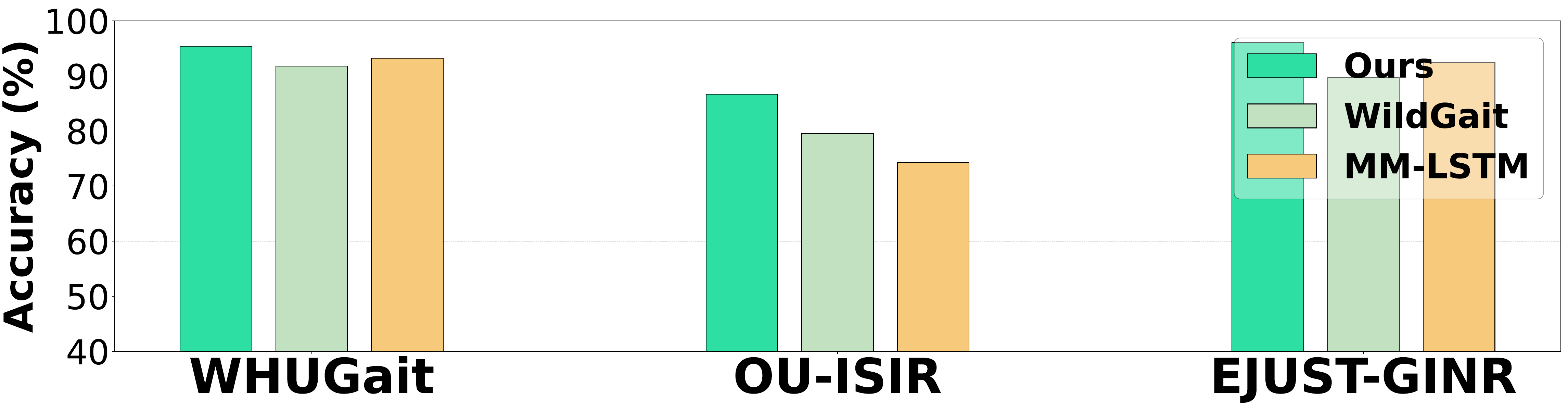}
        \caption{Gait Recognition Task}
    \end{subfigure}\hfill
    \begin{subfigure}{0.25\textwidth}
        \centering
        \includegraphics[width=\textwidth]{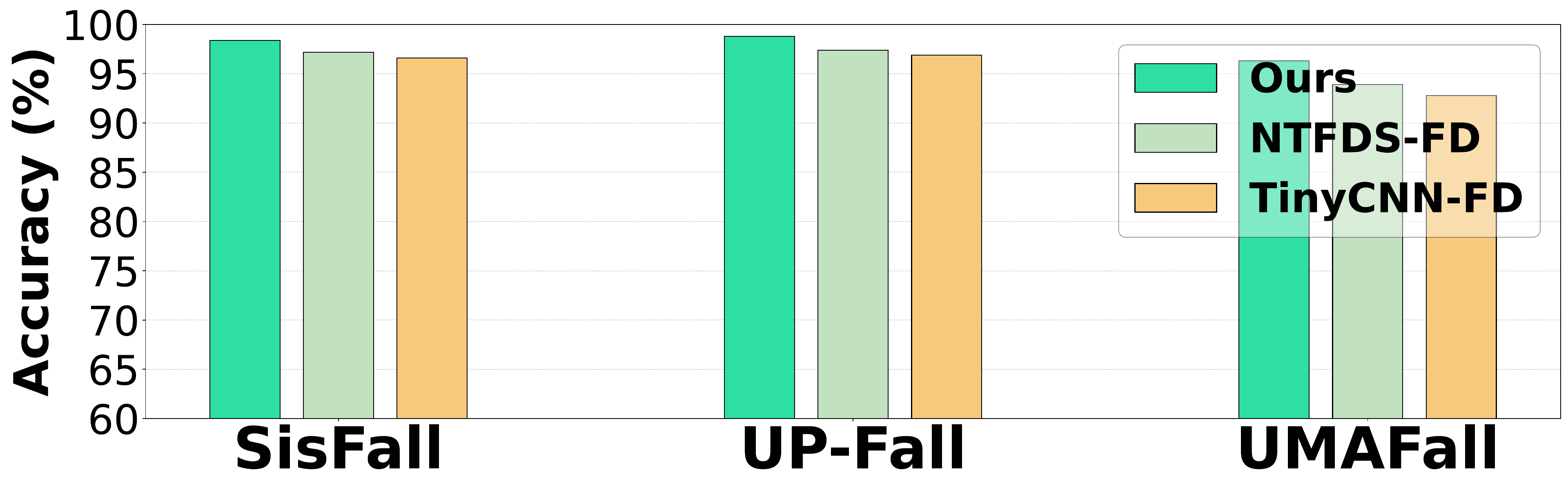}
        \caption{Fall Detection Task}
    \end{subfigure}\hfill
    \vspace{-0.2cm}
\caption{Performance on three downstream tasks: HAR, gait recognition, and fall detection. These results show that the underlying motion predicted by our framework can serve as a generic skeleton representation, allowing simple integration with existing skeleton-based models and straightforward extension to diverse downstream}\vspace{-0.1cm}
    \label{fig:Downstream_Task}
\end{figure*}
\subsection{Ablation Study}
\subsubsection{Ablation Study of Model Components}
\label{sec: moudle}In this section, we will perform a ablation study by step-wisely removing each model componments.Starting from the full model (Ours(+All)), we (1) replace the learned body-shape module with ground-truth body shape (GT BodyShape), (2) further remove frequency-spatial constraints (RM Constrain), (3) additionally remove IMU–object relative-motion prediction (RM IMU–OBJ), (4) drop the multi-view kinematic tree (RM M-V KT), and (5) finally remove the hybrid encoder (RM Hyb-Enc).
As shown in Figure~\ref{fig:ablation}, for the motion capture task, replacing ground-truth body-shape parameters with our predicted ones yields almost identical performance, indicating that the model has successfully captured the underlying complex kinematic structure. Under the tight-attachment setting, removing either the frequency-spatial constraint or the IMU–object relative-motion inference causes only marginal degradation. In contrast, removing the multi-view kinematic tree leaves several joints with little or no supervision, leading to a clear increase in error. Eliminating the hybrid encoder further amplifies this error by preventing effective noise suppression. Overall, the trend indicates that as sensor attachment becomes looser, accurate IMU–object relative-motion inference becomes increasingly critical.
For inertial tracking, which lacks a skeletal or kinematic prior, we ablate only the frequency-spatial constraint, the relative-motion modeling, and the hybrid encoder. Under the loose-attachment condition, removing any of these modules again leads to substantial performance degradation.

\subsubsection{Effect of Uncertainty Sampling Density}

To implement our uncertainty-aware multi-view kinematic chain, we use Monte Carlo sampling with particle-filter–style propagation. This raises a practical trade-off: high sampling densities increase computation, while too few samples can destabilize uncertainty estimates. To examine this, we study how sampling density affects reliability.

Figure~\ref{fig:reliability_uncertainty} plots output uncertainty versus angular error under different sampling densities. As the number of samples decreases, the error distribution shifts upward, especially in sparse-sensor settings. However, the calibration curves remain close to the ideal diagonal, indicating that our uncertainty estimates stay reliable over a broad range of sampling densities. Mean error increases substantially only when the sample count drops to two.

\vspace{-0.3cm}

\subsection{Downstream Task Extension}

\noindent
Once accurate MoCap poses are available, a wide range of downstream tasks become substantially easier. We further evaluate three representative downstream tasks: human activity recognition (HAR), gait recognition, and fall detection. Following common practice in skeleton-based learning, we feed our reconstructed joint sequences into standard downstream backbones, including ST-GCN~\cite{yan2018spatialtemporalgraphconvolutional} for HAR, GaitPT~\cite{catruna2023gaitpt} for gait recognition, and a lightweight skeleton-based 3D-CNN~\cite{noor2023lightweight} for fall detection. This setup allows us to directly assess whether our label-free pose estimation provides stable and informative motion representations beyond pose reconstruction itself.

\noindent\textbf{Baselines.}
We compare against representative task-specific baselines that operate directly on raw IMU signals. For HAR, we use LIMU-GRU~\cite{limu_loc}, DCNN~\cite{dcnn}, and DeepSense~\cite{yao2017deepsense}. For gait recognition, we use Deep Learning-Based Gait Recognition Using Smartphones in the Wild~\cite{zou2020gait} and Multi-Model Long Short-Term Memory Network~\cite{tran2021multi}. For fall detection, we use NT-FDS~\cite{waheed2021ntfds} and TinyCNN-FD~\cite{yu2023practical}.

\noindent\textbf{Results and Comparison.}
As shown in \ref{fig:Downstream_Task}, the overall trend is consistent across all three tasks. On cleaner and more controlled data, direct IMU-based baselines remain competitive. However, under less constrained real-world conditions, their performance degrades much more noticeably, while our method remains robust. This suggests that \textit{if a model cannot disentangle sensor-induced disturbances, even high-level downstream tasks become difficult for neural networks}.
\begin{table}[t]\small
\centering
\setlength{\tabcolsep}{3.5pt}
\begin{tabular}{c|c||c|c}
\hline
\textbf{Cutoff Freq. (Hz)} & \textbf{Ang Err ($^\circ$)} & \textbf{Spatial Scale ($k$)} & \textbf{Ang Err ($^\circ$)} \\
\hline
20 & 13.42 & 0.50$\times$ & 13.29 \\
25 (default) & 13.12 & 0.75$\times$ & 13.08 \\
30 & 12.98 & 1.00$\times$  & 13.12 \\
35 & 13.51 & 1.50$\times$ & 13.31 \\
40 & 13.46 & 2.00$\times$ & 13.79 \\
\hline
\hline
\end{tabular}

\caption{Sensitivity analysis of frequency and spatial priors. The spatial scale $k$ is applied to the default spatial motion bounds specified in Table~\ref{tab:sensor-motion}.}
\vspace{-0.8cm}
\label{tab:sensitivity_prior}
\end{table}

\subsection{Sensitivity Analysis of Frequency and Spatial Priors}

We further evaluate the sensitivity of the frequency and spatial priors in Table~\ref{tab:sensitivity_prior}. The results show a broad performance plateau, indicating that our framework is not highly sensitive to either hyperparameter. Varying the cutoff frequency from 20--40\,Hz changes the angular error only slightly, from 12.98$^\circ$ to 13.51$^\circ$. Similarly, scaling the spatial bound from 0.5$\times$ to 2.0$\times$ keeps the error largely stable, ranging from 13.08$^\circ$ to 13.79$^\circ$, with noticeable degradation appearing only under an overly loose bound of 3.0$\times$.

These results suggest that the proposed priors serve as effective structural regularizers rather than fragile hand-tuned constraints. They provide useful physical guidance while preserving sufficient flexibility across different motions and sensing conditions, further supporting the robustness and practicality of our framework.

\begin{table}[t]
    \centering
    \small
    \setlength{\tabcolsep}{5pt}

    \begin{minipage}[t]{0.48\textwidth}
        \centering
        \begin{tabular}{l|c|c|c}
            \hline
            \multicolumn{4}{c}{\textbf{Mocap}} \\
            \hline
            \textbf{Model} & \textbf{iPhone 16 Pro Max} & \textbf{ARM Cortex-M7 MCU} & \textbf{Error} \\
            \hline
            Ours-Lite & 0.065s & 3.462s & \cellcolor{red!20}13.65$^\circ$ \\
            PIP & 0.120s & 4.641s & 19.50$^\circ$ \\
            SSL-Pose & 0.266s & 15.329s & 27.31$^\circ$ \\
            \hline
        \end{tabular}
    \end{minipage}
    \hfill
    \begin{minipage}[t]{0.48\textwidth}
        \centering
        \begin{tabular}{l|c|c|c}
            \hline
            \multicolumn{4}{c}{\textbf{Tracking}} \\
            \hline
            \textbf{Model} & \textbf{iPhone 16 Pro Max} & \textbf{ARM Cortex-M7 MCU} & \textbf{Error} \\
            \hline
            Ours-Lite & 0.032s & 2.065s & \cellcolor{red!20}8.88m \\
            RoNIN & 0.192s & 9.645s & 19.82m \\
            LIMU & 0.044s & 5.36s & 45.31m \\
            \hline
        \end{tabular}
    \end{minipage}

    \caption{Runtime comparison on edge devices.}
    \vspace{-0.7cm}
    \label{tab:runtime_comparison}
\end{table}

\noindent
\subsection{Time Efficiency Evaluation}

Table~\ref{tab:runtime_comparison} reports the runtime of all methods on an iPhone 16 Pro Max and an ARM Cortex-M7 MCU for a 6-second input window. Ours-Lite runs in 0.032--0.065s on the iPhone and 2.065--3.462s on the MCU for both tracking and mocap, while still achieving the lowest error. These results show that our framework can support real-time deployment on resource-constrained embedded devices.

\section{Conclusion}~\label{sec:conclusion}
\vspace{-0.3cm}

This paper presents a physics self-supervised learning framework for IMU sensing, addressing both inertial tracking and motion capture without requiring manual labels. By combining a learnable physics decoder with a noise-aware neural encoder, our method preserves the structural advantages of physical modeling while improving robustness to sensing noise, placement variation, and user diversity. The framework further incorporates probabilistic frequency--spatial constraints, multi-view kinematic propagation, and uncertainty-aware modeling to better handle the ambiguity and heterogeneity inherent in real-world IMU sensing. Extensive experiments on both self-collected and public datasets show that our method consistently outperforms supervised and self-supervised baselines, especially under realistic and less controlled conditions. In addition, the results on edge devices demonstrate that our lightweight variant can support practical deployment with low runtime overhead. Together, these findings suggest that physics self-supervised learning provides a scalable and practical solution for robust IMU sensing in real-world deployments.

\section{Disclaimer}
This paper was prepared for informational
purposes with contributions from the Global
Technology Applied Research center of
JPMorgan Chase \& Co. (JPMC) and is not
a product of its, or its affiliates’, Research
Departments. JPMC and its affiliates make
no representations or warranties, express
or implied, regarding the completeness,
accuracy, or reliability of the information
herein, and accept no liability for its use or
any related outcomes. This document does
not constitute investment advice, financial
research, or a recommendation or offer to
buy or sell any security, financial instrument,
product, or service. 

\section{Acknowledgements}

This work is partially supported by the National Science Foundation under grants IIS-2107200, and CNS-2038923.

}

\bibliographystyle{ACM-Reference-Format}
\bibliography{references}
\clearpage

\end{document}